\documentclass[11pt]{article}

\usepackage[preprint]{acl}

\usepackage{times}
\usepackage{latexsym}
\usepackage[T1]{fontenc}
\usepackage[utf8]{inputenc}
\usepackage{microtype}
\usepackage{inconsolata}
\usepackage{graphicx}
\usepackage{subcaption}
\usepackage{float}   
\usepackage{pgfplots}
\pgfplotsset{compat=1.18}
\usetikzlibrary{shapes.geometric}
\usepackage{booktabs}
\usepackage{array}
\usepackage{amsmath}
\usepackage{amssymb}
\usepackage{xcolor}
\usepackage{colortbl}
\usepackage{tcolorbox}
\tcbuselibrary{breakable}

\definecolor{sansi}{HTML}{0B5CAD}
\definecolor{sansidark}{HTML}{05295C}
\definecolor{sansimid}{HTML}{4A98E3}
\definecolor{sansilight}{HTML}{9FCBF7}
\definecolor{sansibg}{HTML}{E9F2FC}   
\definecolor{qwen}{HTML}{1B8A4C}
\definecolor{smol}{HTML}{D9480F}
\definecolor{ouro}{HTML}{8C8C8C}
\definecolor{jev}{HTML}{4D4D4D}
\definecolor{groupbg}{HTML}{F2F2F2}   
\definecolor{better}{HTML}{0F7B6C}    
\definecolor{worse}{HTML}{C0392B}     
\newcommand{\mkfill}[1]{\tikz[baseline=-0.05em]\fill[#1] (0,0) rectangle (0.62em,0.62em);\hspace{0.35em}}
\newcommand{\mkopen}[1]{\tikz[baseline=-0.05em]\draw[#1,line width=0.8pt] (0.04em,0.04em) rectangle (0.58em,0.58em);\hspace{0.35em}}
\newcommand{\mkSansi}{\mkfill{sansi}}
\newcommand{\mkSansiL}{\mkfill{sansidark}}
\newcommand{\mkSansiV}{\mkopen{sansi}}
\newcommand{\mkQwen}{\mkfill{qwen}}
\newcommand{\mkQwenS}{\mkopen{qwen}}
\newcommand{\mkSmol}{\mkfill{smol}}
\newcommand{\mkOuro}{\mkfill{ouro}}
\newcommand{\mkJev}{\mkopen{jev}}
\definecolor{kev}{HTML}{A8327E}        
\newcommand{\mkKev}{\mkfill{kev}}
\newcommand{\mkKevR}{\mkopen{kev}}
\newcommand{\ciint}[2]{\,{\scriptsize\textcolor{black!50}{[$#1$, $#2$]}}}
\newcommand{\dbetter}[3]{\textcolor{better}{{\boldmath$#1$}}\ciint{#2}{#3}}
\newcommand{\dworse}[3]{\textcolor{worse}{{\boldmath$#1$}}\ciint{#2}{#3}}
\newcommand{\dns}[3]{\textcolor{black!55}{$#1$}\ciint{#2}{#3}}
\newcommand{\untrained}[1]{\textcolor{black!45}{#1}}
\newcommand{\rqlabel}[1]{\texorpdfstring{\textcolor{sansi}{RQ#1}}{RQ#1}}
\newtcolorbox{takeaway}{breakable, colback=sansibg, colframe=sansi, boxrule=0pt, leftrule=2.5pt, arc=0pt, outer arc=0pt,
  left=5pt, right=5pt, top=3pt, bottom=3pt, before skip=6pt, after skip=8pt}
\newcommand{\takeawaylabel}{\textbf{\textcolor{sansidark}{Takeaway.}}~}
\usepackage{iftex}
\ifPDFTeX
  \usepackage{CJKutf8}
  \newcommand{\zh}[1]{\begin{CJK*}{UTF8}{gbsn}#1\end{CJK*}}
\else
  ="Songti SC" at 8pt
  \newcommand{\zh}[1]{{\sansizhfont #1}}
\fi

\newcommand{\sansi}{\texorpdfstring{\textcolor{sansidark}{SanSi}}{SanSi}}

\newcommand{\sd}[1]{{\scriptsize\textcolor{black!50}{$\pm$#1}}}

\title{\sansi{}: A Looped Typed Decision Model for System~1.5 Thinking}

\author{Shuyu Gan \quad Young-Jun Lee \quad Dongyeop Kang \\
University of Minnesota \\
\texttt{\{gan00067,lee05727,dongyeop\}@umn.edu} \\[0.45em]
\href{https://minnesotanlp.github.io/Sansi/}{\raisebox{-0.2em}{\includegraphics[height=1.1em]{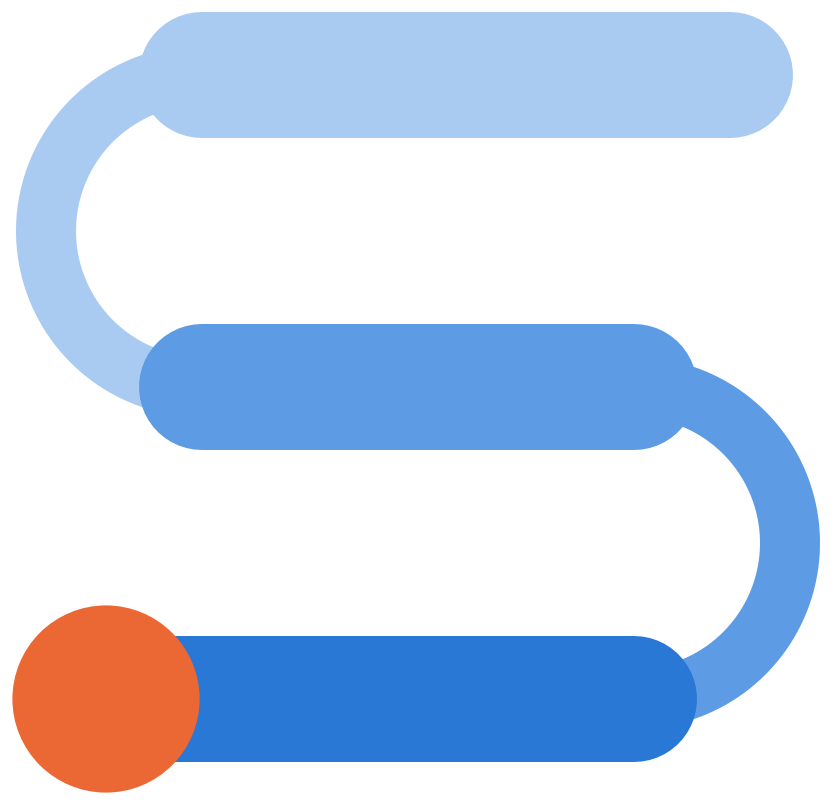}}\hspace{0.35em}\textcolor{sansi}{\texttt{SanSi Project Page}}}}

\begin{document}
\raggedbottom   
\maketitle

\begin{abstract}
Typed decision models answer a declared question without generating text: a decision head returns a probability for each of the declared options in a single forward pass. A single pass is fast, intuitive System~1 thinking.
We study what lies between one pass and generated reasoning: \emph{looping}, in which the same layers are recursively applied several times before one typed readout. Each loop lets the model revise its hidden state before it commits to an answer, without generating a token; we call this System~1.5 thinking.
We propose \textbf{\sansi{}}, which turns a pre-trained looped language model into a typed decision model. 
The option probabilities are read after every loop, and every loop is trained with a proper scoring rule, so that one model serves every budget from one loop to eight in a single run. 
On 10,027 test decisions from 59 sources, \sansi{} reaches 72.0\% accuracy: 13.5 points above a non-looped model of the same shape trained with the same recipe, 5.3 points above a newer non-looped model of its size, and 1.8 points below one with three times the parameters. 
On two depth-controlled tasks, loops extend the solvable depth beyond the depths seen in training, where the larger single-pass model fails. 
Used as the judge for policy optimization with reinforcement learning, without gold answers, \sansi{} raises the share of the generator's correct answers from 37\% to 52\%.
\end{abstract}

\section{Introduction}

A growing share of language model calls do not ask for text, but a \textit{decision}: is this message spam, does this case satisfy the policy, which of these four answers follows from the document. \emph{Typed decision models} \citep{typesafe2026jev,tang2026typed} serve these calls directly. The caller declares the options, for example \emph{spam} and \emph{not spam}, and the model returns one probability per option in a single forward pass, with no generated text to parse. The commercial Jev API popularised the format, open models such as Kev \citep{palmer2026kev} follow the same contract, and the probabilities are increasingly used to accept, escalate or route \citep{li2026judge,deusser2026evaluating}.

\begin{figure}[!t]
\centering
\includegraphics[width=\columnwidth]{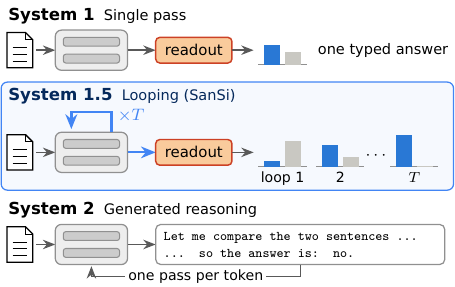}\\[5pt]
\includegraphics[width=\columnwidth]{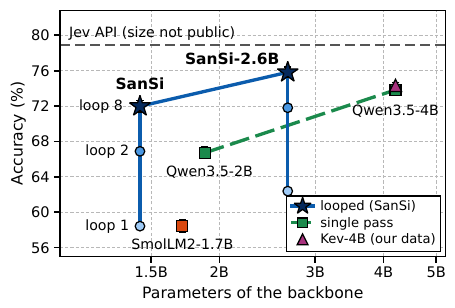}
\caption{\textbf{Top:} three ways to spend computation on a decision. \textbf{Bottom:} Accuracy on the 10,027 test decisions against backbone size, for the models trained on our data. Single-pass models gain accuracy by adding parameters; \sansi{} gains it by looping the same parameters. Kev-4B (our data) is Qwen3.5-4B trained with Kev's recipe. Dashed line is the Jev API, which was not trained on our data. 
}\vspace{-5mm}
\label{fig:pareto}
\end{figure}

A single pass gives a fixed amount of computation to every decision. This is enough to tell whether a message is spam, but many decisions require several dependent steps: following a chain of relations, composing rules, or noticing that the decisive evidence is missing. 
Take a chain of statements: \emph{Omar is honest, Bert says that Omar tells the truth, and Ben says that Bert lies}. Whether Ben tells the truth can only be found by following the chain, one statement at a time. On such chains, the accuracy of the Jev API falls from 100\% with one step to 62.5\% with three and is at chance from six steps on (Appendix~\ref{app:wall}). When one pass is not enough for complex reasoning, a larger model adds parameters and memory, or generated reasoning adds tokens and latency before the decision is made.
In this work, we explore \emph{looping}: to apply the same layers recursively to their own output, which adds computation but no parameters and keeps the typed contract. If a single pass is System~1 and generated reasoning is System~2, looping is a ``System~1.5'' \citep{wang2025system15}.


Looped transformers learn iterative algorithms and generalise to longer inputs \citep{giannou2023looped,yang2024looped,fan2024looped,bansal2022endtoend,saunshi2025reasoning}, building on Universal Transformers \citep{dehghani2019universal}, and language models are now pre-trained to loop as a form of latent reasoning \citep{geiping2025recurrent,zhu2025ouro}.
Jev-LCT \citep{cao2026jevlct} applies looping to a typed decision model, but loops only the top two layers of a non-looped model and exits early.  
We study in depth how much accuracy looping adds, what it costs, and what it does to the probabilities that callers rely on.

\begin{figure*}[t]
\centering
\begin{subfigure}[t]{0.39375\textwidth}
\includegraphics[width=\linewidth]{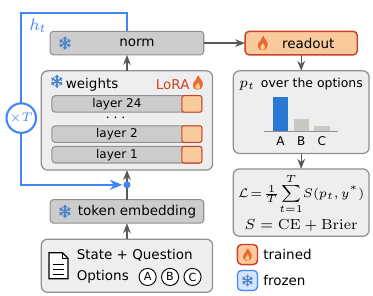}
\caption{Training}
\label{fig:model-train}
\end{subfigure}%
\begin{subfigure}[t]{0.60625\textwidth}
\includegraphics[width=\linewidth]{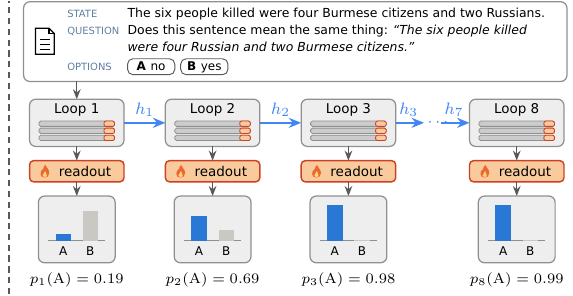}
\caption{Test time}
\label{fig:example}
\end{subfigure}
\caption{\sansi{}. (a) Training: one stack of 24 layers is applied $T$ times; after every loop the option probabilities are read and trained towards the target. (b) Test time: a test item from PAWS \citep{zhang2019paws} followed through the loops; the first loop prefers the wrong option, the later loops the correct one.}
\label{fig:model}
\end{figure*}

We present \textbf{\sansi{}}\footnote{\sansi{} is the pinyin of \zh{三思}, ``think thrice'', from the Analects: ``Ji Wenzi thought thrice before acting'' (\zh{季文子三思而后行}). Confucius is said to have replied that twice would do; in our data the second loop brings 62\% of the gain from loop 1 to loop 8, and the third brings it to 88\%.}, a new family of looped typed decision models built on Ouro \citep{zhu2025ouro}, a language model pre-trained to loop. We train two sizes, \sansi{} on Ouro-1.4B and \sansi{}-2.6B on Ouro-2.6B; most of our analyses use the first. The backbone stays frozen: we train only LoRA adapters and a small readout for every loop, 61M parameters in total for \sansi{}.
\sansi{} reads the option probabilities after \emph{every} loop and trains each with a proper scoring rule, so one model serves every budget from one loop to eight (Figure~\ref{fig:model}). Training every loop is what makes the early loops usable: read after one loop, \sansi{} reaches 58.4\%, against 35.6\% when only the last loop is trained (Table~\ref{tab:ablation}).

We test \sansi{} on 10,027 test decisions from 59 sources.
With the same data, recipe and seeds, looping adds 13.5 points over SmolLM2-1.7B, a non-looped typed decision model of the same shape, and brings \sansi{} within 1.8 points of Qwen3.5-4B, a single-pass typed decision model with three times the parameters (Figure~\ref{fig:pareto}).
The gain is smallest on classification (+3.1 points) and largest on multi-step reasoning (+15.1), long documents (+16.7) and knowledge questions (+17.6) (Figure~\ref{fig:types-gain}), and it costs 7.7 times the computation of a single pass.
Reading \sansi{} after three loops already gives 88\% of its gain from loop 1 to loop 8; most answers settle by the fourth loop, and harder items settle later. Ouro's own exit gate, which our training leaves frozen, gives a stopping rule without tuning: it keeps the accuracy of loop 8 with 4.5 loops on average. The loops make the model better at noticing that the evidence for a decision is missing, but they improve calibration only up to loop 3: after the answers settle, confidence keeps rising. On two depth-controlled tasks, loops reach depths never seen in training, where the larger single-pass model fails: on liar chains of 9 to 16 steps (training goes up to 8), \sansi{} is right on 72.6\% of the items and Qwen3.5-4B on 50.0\%, the chance level.
As the only reward for training a generator with reinforcement learning, \sansi{} raises the share of the generator's correct answers from 37.0\% to 52.1\%, which shows its potential as a judge for policy learning.


\section{Background and Related Work}

\paragraph{Typed decision models.} Jev \citep{typesafe2026jev} and its open reimplementation Kev \citep{palmer2026kev} take a state, a question and a declared set of options and return a probability for each option in one pass. The format covers familiar uses of language models: classification and multiple-choice questions read the probabilities of a fixed set of answers \citep{robinson2023leveraging}, verifiers and judges score candidate outputs \citep{cobbe2021training,zheng2023judging}, and selective prediction answers only when the model is confident enough \citep{kamath2020selective}. In Kev, one request may carry several questions about the same state, and a single small head answers each of them independently; we consider one question per call. Kev is trained with cross-entropy; Jev is reported to be trained with reinforcement learning for calibrated decisions \citep{typesafe2026jev}. Recent audits examine the accuracy of Jev and the reliability of its probabilities \citep{porcedda2026sys1cal,deusser2026evaluating,li2026beyond,sun2026typesafe,tang2026typed}. Unlike Jev-LCT \citep{cao2026jevlct}, we start from a backbone whose whole stack was pre-trained to loop, read every loop, and measure what the loops add against non-looped models under one recipe.

\paragraph{Looped models and anytime prediction.} Recent work on looped language models studies how the loops are used \citep{dau2026beyond,kohli2026loop,guo2026right,blayney2026mechanistic}, their stability and halting \citep{yang2026stabilizing,popescu2026adaptive}, architectural variants \citep{jeddi2026loopformer,yu2026tloopformer,wang2026recurtrace}, adding loops to models pre-trained without them \citep{mcleish2025retrofitted,shapiro2026retrofitting,chen2026trainingfree,park2026loopus,marchenko2026closing}, and tool calling \citep{popescu2026tool}. The same idea appears above the level of layers: a self-improving agent can apply one fixed operation repeatedly to the result of its previous application and let convergence decide the depth \citep{kim2026metan}. Reading a prediction at several depths relates to adaptive computation \citep{graves2016act,banino2021pondernet} and early exit \citep{xin2020deebert,zhou2020bert}, where the predictions of several exits can also be combined \citep{sun2021early,wolczyk2021zero}, as we do when we average the loops (\S\ref{sec:ablations}); we do not propose a new halting rule but test existing ones, including Ouro's exit gate (Appendix~\ref{app:stop}). In networks with exits at several layers, later layers turn some right predictions into wrong ones \citep{kaya2019shallow}, and the layer at which a prediction settles measures how hard an example is \citep{baldock2021deep}; we find both for loops (\S\ref{sec:rq2}). More generated reasoning can make models overconfident \citep{lacombe2025dont,hiremath2026drift}; we find a related effect for loops (\S\ref{sec:analysis}).

\section{\sansi{}: A Looped Typed Decision Model}
\label{sec:method}

\paragraph{Task.} A typed decision is a call with three parts: a state $s$ (a passage, a set of rules or records), a question $q$, and $K \ge 2$ declared options. The model returns a distribution $p \in \Delta^{K}$ over the options and no text. The target distribution $y^{*}$ is one-hot when the item has a correct option, uniform when the state lacks the evidence needed to answer (an \emph{unanswerable} item), and the annotators' label distribution when the item was labelled by a crowd.

\paragraph{Looped backbone.} Ouro-1.4B \citep{zhu2025ouro} applies one stack $F_{\theta}$ of 24 transformer layers repeatedly (Figure~\ref{fig:model-train}). With token embeddings $h_{0}$ and the model's final normalisation $N$,
\begin{equation}
h_{t} = N\big(F_{\theta}(h_{t-1})\big), \qquad t = 1, \dots, T .
\end{equation}
The normalised state $h_{t}$ is both the output of loop $t$ and the input of loop $t+1$; the input tokens are not injected again. Running $T$ loops therefore costs $T$ passes through the stack and adds no parameters. Ouro was pre-trained with four loops.

\paragraph{Readout after every loop.} The item is rendered as a prompt that lists the options under the letters A, B, \dots{} and ends in ``Answer:''. Let $z_{t}$ be the row of $h_{t}$ at the last prompt token. The option logits are
\begin{equation}
\ell_{t} = (W + U_{t} V_{t})\, z_{t} / \tau_{t}, \quad p_{t} = \mathrm{softmax}(\ell_{t}),
\end{equation}
restricted to the $K$ declared options. $W$ holds the rows of the frozen language-model head for the option letters; the rank-16 correction $U_{t} V_{t}$ and the scale $\tau_{t}$ are trained, one per loop, and start at the identity (Appendix~\ref{app:training}). Since $p_{t}$ is computed from $h_{t}$ alone, one pass with $T$ loops yields the decisions of all budgets $1, \dots, T$ (Figure~\ref{fig:example}).

\paragraph{Training.} We train every loop towards the target with the sum of two proper scoring rules, cross-entropy and the Brier score:
\begin{equation}
\mathcal{L} = \frac{1}{T} \sum_{t=1}^{T} \Big[ -{\textstyle\sum_{k}}\, y^{*}_{k} \log p_{t,k} + {\textstyle\sum_{k}}\, (p_{t,k} - y^{*}_{k})^{2} \Big].
\label{eq:loss}
\end{equation}
For an item with one correct option, cross-entropy looks only at the probability of that option and penalises a confident error heavily. The Brier score looks at the probability of every option and is bounded. Both are smallest when the model outputs exactly the target probabilities. The loss of every loop is backpropagated through all the loops before it. The weights of the backbone are frozen. We train LoRA adapters \citep{hu2022lora} of rank 64 on all attention and feed-forward projections (60.6M parameters, shared by all loops) and the readouts (0.27M), for 1,000 steps of 16 items; the remaining settings are in Appendix~\ref{app:training}. \sansi{} is trained with $T = 8$ loops, twice the four loops of Ouro's pre-training, and is read after the eighth loop unless another loop is named.

\section{Experimental Setup}
\label{sec:setup}

\begin{table}[t]
\centering
\small
\setlength{\tabcolsep}{4pt}
\begin{tabular}{@{}lrrrr@{}}
\toprule
 & & \multicolumn{3}{c}{Test} \\
\cmidrule(l){3-5}
Item type & Train & In-dist. & Near & Far \\
\midrule
A\; Classification & 2,400 & 321 & -- & 768 \\
B\; Multi-step reasoning & 5,500 & 800 & 991 & 1,281 \\
C\; Uncertain evidence & 2,300 & 478 & 640 & 440 \\
D\; Long documents & 1,000 & 357 & -- & 544 \\
E\; Sentence pairs & 1,600 & 160 & 240 & 968 \\
F\; Knowledge & -- & -- & -- & 1,808 \\
\midrule
Total & 12,800 & 2,116 & 1,871 & 5,809 \\
\bottomrule
\end{tabular}
\caption{The decision suite: number of items by type and by distance from the training data. Type C contains the unanswerable and crowd-labelled items. The test set also contains the 231 public JevBench items (10,027 test items in total). Sources and an example of every type are in Appendix~\ref{app:data}.}
\label{tab:data}
\end{table}

\paragraph{Data.} We build one suite of typed decisions from public datasets, the training and transfer suites of Kev, and the public items of JevBench \citep{standhartinger2026jevbench}, all rendered in the prompt format above (Table~\ref{tab:data}; sources and an example of every type in Appendix~\ref{app:data}). Items fall into six types by what they demand; knowledge questions occur only in the test set. Test items are also grouped by distance from the 12,800 training items: \emph{in-distribution} items are new items from the 20 training sources; \emph{near transfer} items are harder or reworded trained types, such as CLUTRR with 5--10 hops (2--4 in training) \citep{sinha2019clutrr}; \emph{far transfer} items come from 30 sources unseen in training, including FOLIO, BBH, MMLU and QuALITY \citep{han2022folio,suzgun2023bbh,hendrycks2021mmlu,pang2022quality}. The 231 public JevBench items form a fourth group. The test set has 10,027 items; 382 are unanswerable and 766 carry a crowd distribution \citep{nie2020chaosnli}. Another 2,471 items form the development set.

\paragraph{Models.} We compare \sansi{} with four single-pass models, which are run once and read exactly as loop 1 of \sansi{} is. Two are controls of the same shape: \emph{Ouro-1.4B, one loop}, \sansi{}'s backbone trained and run with $T = 1$, and \emph{SmolLM2-1.7B} \citep{allal2025smollm2}, the closest non-looped model (Ouro's tokenizer, 24 layers, hidden size 2,048). Two are references from a newer family: \emph{Qwen3.5-2B} (1.9B parameters), which is close to \sansi{} in size, and \emph{Qwen3.5-4B} \citep{qwen2026qwen35} (4.2B), which has three times its parameters. A fifth single-pass model tests whether these references depend on our recipe: \emph{Kev-4B (our data)} is Qwen3.5-4B trained on our data with Kev's own code and recipe \citep{palmer2026kev}. We use the Base checkpoint of every backbone. All other models are fine-tuned with the same recipe: the same training items in the same order, LoRA of the same rank on the same kinds of modules, the same readout and the same loss, with three seeds each; this includes \sansi{}-2.6B, the same recipe on the larger looped backbone Ouro-2.6B. We report means over the seeds and, in the tables, the standard deviation ($\pm$). We also report the backbones without fine-tuning and two released models that were not trained on our data, Kev-4B and the commercial Jev API (jev-1.13.0); these are in Appendix~\ref{app:main}. Appendix~\ref{app:training} gives the sizes and the measured cost of every model.

\begin{table*}[t]
\centering
\small
\setlength{\tabcolsep}{4.5pt}
\begin{tabular}{@{}lccc cccc cc@{}}
\toprule
 & & & & \multicolumn{4}{c}{Accuracy (\%) $\uparrow$} & & Evidence \\
\cmidrule(lr){5-8}
Model & Params & Loops & Cost & All & In-dist. & Near & Far & ECE $\downarrow$ & AUROC $\uparrow$ \\
\midrule
\mkSmol{}SmolLM2-1.7B & 1.7B & 1 & 1.1 & 58.4\sd{0.7} & 76.1\sd{0.8} & 57.7\sd{0.6} & 52.3\sd{0.7} & \textbf{.069}\sd{.020} & .765\sd{.009} \\
\mkOuro{}Ouro-1.4B, one loop & 1.4B & 1 & 1.0 & 58.6\sd{0.2} & 77.3\sd{1.0} & 59.9\sd{0.9} & 51.3\sd{0.4} & .137\sd{.009} & .837\sd{.007} \\
\mkQwenS{}Qwen3.5-2B & 1.9B & 1 & 1.2 & 66.7\sd{0.7} & 84.3\sd{1.4} & 62.3\sd{2.5} & 61.7\sd{0.4} & .123\sd{.007} & .896\sd{.002} \\
\mkQwen{}Qwen3.5-4B & 4.2B & 1 & 2.4 & 73.8\sd{0.6} & 88.5\sd{0.4} & 68.9\sd{2.4} & 70.1\sd{0.3} & .113\sd{.003} & \textbf{.948}\sd{.003} \\
\mkKev{}Kev-4B (our data) & 4.2B & 1 & 2.1 & 74.3\sd{0.3} & \textbf{88.6}\sd{0.2} & 70.9\sd{0.7} & 70.3\sd{0.6} & .119\sd{.004} & .947\sd{.002} \\
\midrule
\rowcolor{sansibg}\mkSansi{}\sansi{}, read after loop 3 & 1.4B & 3 & 2.9 & 70.4\sd{0.6} & 85.6\sd{0.6} & 67.0\sd{1.6} & 65.9\sd{0.9} & .082\sd{.010} & .921\sd{.006} \\
\rowcolor{sansibg}\mkSansi{}\textbf{\sansi{}} & 1.4B & 8 & 7.7 & 72.0\sd{0.7} & 86.6\sd{0.6} & 67.7\sd{2.7} & 68.0\sd{0.1} & .093\sd{.012} & .935\sd{.004} \\
\rowcolor{sansibg}\mkSansiL{}\sansi{}-2.6B & 2.7B & 8 & 14.8 & \textbf{75.8}\sd{0.6} & 88.4\sd{0.8} & \textbf{74.4}\sd{2.5} & \textbf{71.6}\sd{0.3} & .078\sd{.000} & .943\sd{.000} \\
\bottomrule
\end{tabular}

\caption{Main results on the 10,027 test items: mean $\pm$ standard deviation over three training seeds (bold: best). All models are trained on the same data, with our recipe except Kev-4B (our data), which uses Kev's. \sansi{}-2.6B: the same recipe on Ouro-2.6B. Cost: GPU time of one pass over the test set, relative to Ouro-1.4B with one loop; all models are timed on one RTX A6000 with the same setting (Table~\ref{tab:params} in Appendix~\ref{app:training}; loop 3: three eighths of the eight-loop pass). Appendix~\ref{app:main} gives the full table. Shaded rows: \sansi{}.}
\label{tab:main}
\end{table*}

\paragraph{Metrics.} \emph{Accuracy} counts an item as right when the most probable option is the gold option (the majority option for crowd-labelled items). An unanswerable item is right when the model gives no \emph{hard answer}, that is, when its top probability is below $(1 + 1/K)/2$. Unanswerable items make up 3.8\% of the test set; accuracy on the answerable items alone, or a threshold a quarter or three quarters of the way from uniform to certain, changes none of the comparisons (Appendix~\ref{app:main-abstain}). \emph{Confidence} is the top probability, and \emph{ECE} the expected calibration error over ten equal-width bins of confidence, on answerable items. \emph{Evidence AUROC} is the probability that an item with its key evidence receives a higher confidence than an item without it, on the 786 items of the sources that contain both. We measure the \emph{cost} of a model by the GPU time of one pass over the test set (Appendix~\ref{app:training}). Intervals are 95\% bootstrap intervals over groups of related items. Appendix~\ref{app:metrics} gives the formulas and the detailed definitions.

\section{Results}
\label{sec:results}

We ask four questions: does looping help, and at what cost (\rqlabel{1}, \S\ref{sec:rq1}); how do the answers change from loop to loop (\rqlabel{2}, \S\ref{sec:rq2}); what do the loops do to the probabilities (\rqlabel{3}, \S\ref{sec:analysis}); and does looping buy reasoning depth (\rqlabel{4}, \S\ref{sec:depth})?

\subsection{\rqlabel{1}: Does looping help, and at what cost?}
\label{sec:rq1}

\paragraph{Gain at the same shape.} Table~\ref{tab:main} gives the main comparison. \sansi{} and SmolLM2-1.7B share the same shape (24 layers, hidden size 2,048), training items, recipe and readout; \sansi{} applies its layers eight times, SmolLM2 once. \sansi{} reaches 72.0\% against 58.4\%, a gain of 13.5 points (95\% interval [12.5, 14.5]; Table~\ref{tab:ci}), stable across seeds (12.9--14.8). The gain is broad: it holds in every test group and on 55 of 59 test sources (Table~\ref{tab:sources}). The backbone is not the cause: Ouro-1.4B trained and run with a single loop reaches 58.6\%, indistinguishable from SmolLM2 (+0.1 [$-$0.7, 1.0]). The 13.5 points come from the loops.

\paragraph{Where the gain is largest.} The gain is not the same for every kind of decision (Figure~\ref{fig:types-gain}; Table~\ref{tab:types} in Appendix~\ref{app:main-types}). It is smallest on classification (+3.1 points), where most items can be decided at a glance and one pass is already enough. It is large where a decision has to bring several pieces of information together: 15.1 points on multi-step reasoning, 16.7 on long documents and 14.2 on sentence pairs. Knowledge questions, which occur only in the test set, gain the most (17.6 points). The gain also varies with distance from training: 10.5 points on in-distribution items, 10.0 on near transfer and 15.8 on far transfer, whose sources are unseen in training (Table~\ref{tab:ci}). Looping thus helps least where one pass already suffices. The two types where the single-pass SmolLM2-1.7B is weakest, multi-step reasoning and knowledge questions (both near 53\%), are among the biggest gainers.

\begin{figure*}[t]
\centering
\begin{subfigure}[b]{0.49\textwidth}
\centering
\includegraphics[width=2.83in]{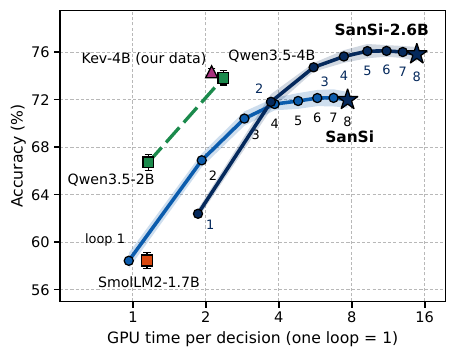}
\caption{Accuracy against computation}
\label{fig:cost}
\end{subfigure}\hfill
\begin{subfigure}[b]{0.49\textwidth}
\centering
\includegraphics[width=3.06in]{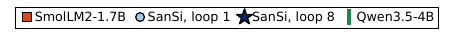}\\[1pt]
\includegraphics[width=3.03in]{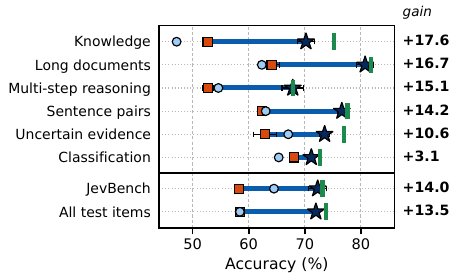}
\caption{Accuracy by item type}
\label{fig:types-gain}
\end{subfigure}
\caption{What looping costs and where it helps. (a) Accuracy against GPU time of one pass over the test set (unit: one loop of Ouro-1.4B). \sansi{} and \sansi{}-2.6B are read after each loop (bands: $\pm$1 s.d., three seeds); Ouro-1.4B trained with one loop coincides with loop 1 of \sansi{}. Kev-4B (our data): Qwen3.5-4B trained with Kev's recipe. All models timed on one RTX A6000 (Table~\ref{tab:params}). (b) Accuracy by item type, ordered by the gain of \sansi{} at loop 8 over SmolLM2-1.7B (thick line; number on the right: gain in points). Error bars: $\pm$1 s.d., three seeds; values in Table~\ref{tab:types}.}
\label{fig:cost-types}
\end{figure*}

\paragraph{What looping costs.} Each loop is another pass through the 24 layers: inference takes 7.7$\times$ and training 9.5$\times$ the GPU time of the one-loop model (Figure~\ref{fig:cost}; Table~\ref{tab:params}). Looping trades computation for parameters, and the trade is set at test time: three loops cost three eighths of the full pass and already reach 70.4\%, and a stopping rule can set the budget per item (\S\ref{sec:rq2}). It costs little memory, since every loop uses the same weights: the test pass needs 6.2 GB with eight loops and 5.3 GB with one, against 12.9 GB for Qwen3.5-4B (Table~\ref{tab:params}). \sansi{}-2.6B extends the curve to higher budgets (Appendix~\ref{app:backbone}): with two loops it equals \sansi{} with four at about the same cost (71.8\% and 71.6\%), and with three loops it passes Qwen3.5-4B (+0.9 points [0.3, 1.5]), at 5.6 times the cost of one loop against 2.4.

\paragraph{Newer and larger single-pass models.} The Qwen3.5 models are references, not controlled comparisons. Qwen3.5-2B, a newer backbone of similar size, is 8.1 points [7.2, 9.0] stronger than Ouro-1.4B run once; \sansi{} draws level at loop 2 (+0.2 [$-$0.5, 0.8]) and leads by 5.3 [4.5, 6.0] at loop 8. Qwen3.5-4B, with three times the parameters, stays ahead: by 1.8 points [1.2, 2.5] at loop 8, where \sansi{} uses 3.3$\times$ its GPU time, and by 3.4 [2.8, 4.1] at loop 3, where it uses 1.2$\times$. With Ouro's exit gate, \sansi{} keeps the accuracy of loop 8 at 1.8$\times$ the GPU time of Qwen3.5-4B if every item runs only its own loops (\S\ref{sec:rq2}). The remaining gap lies on knowledge questions and uncertain evidence, not on multi-step reasoning, where the two are level (Figure~\ref{fig:types-gain}; Appendix~\ref{app:main-types}). Loops improve how a model uses its knowledge, not how much it stores.

This single-pass reference does not depend on our recipe. Trained on the same data with Kev's own code, the same backbone reaches 74.3\% (Kev-4B (our data) in Table~\ref{tab:main}), 0.5 points above Qwen3.5-4B with our recipe, a difference whose interval includes zero ([0.0, 1.0]). \sansi{} is 2.4 points behind this model [1.7, 3.1] with a third of its parameters. \sansi{}-2.6B, our recipe on the larger looped backbone, is 1.5 points ahead of it [0.8, 2.2] with 63\% of its parameters (\S\ref{sec:ablations}; Figure~\ref{fig:pareto}).

The commercial Jev API, the dashed line in Figure~\ref{fig:pareto}, is more accurate than every model that we trained (78.9\% on all test items). Its size and its training data are not public, and it was not trained on our data. Appendix~\ref{app:main-api} therefore compares it with our models on the test items whose sources none of our models was trained on, where it reaches 83.5\%, against 71.9\% for \sansi{}-2.6B.

\begin{takeaway}
\takeawaylabel Eight loops add 13.5 points at fixed shape, data and recipe, paid in computation, not in parameters, and with little extra memory. Per parameter, the looped model beats even a newer single-pass model; per unit of computation, a larger one wins.
\end{takeaway}

\begin{figure*}[t]
\centering
\includegraphics[width=\textwidth]{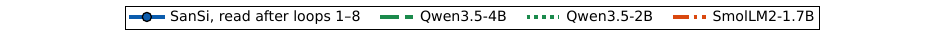}\\[1pt]
\begin{subfigure}[t]{0.2476\textwidth}\includegraphics[width=\linewidth]{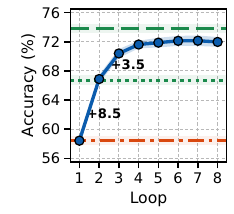}\caption{Accuracy}\end{subfigure}\hfill
\begin{subfigure}[t]{0.2476\textwidth}\includegraphics[width=\linewidth]{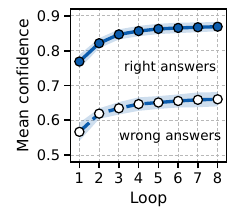}\caption{Confidence}\end{subfigure}\hfill
\begin{subfigure}[t]{0.2476\textwidth}\includegraphics[width=\linewidth]{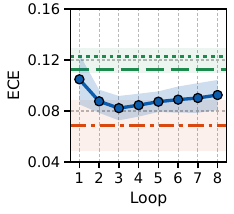}\caption{Calibration error}\end{subfigure}\hfill
\begin{subfigure}[t]{0.2476\textwidth}\includegraphics[width=\linewidth]{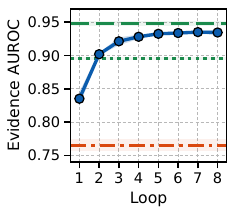}\caption{Missing evidence}\end{subfigure}
\caption{\sansi{} read after each loop on the 10,027 test items (mean $\pm$ s.d., three seeds); horizontal lines: single-pass models; Kev-4B (our data) is not drawn, as its values almost coincide with those of Qwen3.5-4B (Table~\ref{tab:main}). (a) Accuracy, with the gains of loops 2 and 3. (b) Mean confidence on right and wrong answers. (c) ECE on answerable items. (d) Evidence AUROC. Values in Table~\ref{tab:perloop}.}
\label{fig:perloop}
\end{figure*}

\subsection{\rqlabel{2}: How do the answers change from loop to loop?}
\label{sec:rq2}

\paragraph{The gain comes early.} Loop 1 matches the model trained with one loop (58.4\%; $-$0.2 [$-$0.6, 0.3]), so training eight loops costs nothing at loop 1 (Figure~\ref{fig:perloop}a; Table~\ref{tab:perloop} in Appendix~\ref{app:loops}). Loop 2 adds 8.5 points [7.6, 9.3], loop 3 adds 3.5 [3.0, 4.0], loop 4 adds 1.2 [0.9, 1.6], and loops 5--8 together add only 0.4 [$-$0.0, 0.7]. Three loops deliver 88\% of the gain. Accuracy is flat from loop 4 and declines only beyond the trained loops (\S\ref{sec:nloops}).

\begin{figure*}[t]
\centering
\begin{subfigure}[t]{0.245\textwidth}\centering\includegraphics[width=\linewidth]{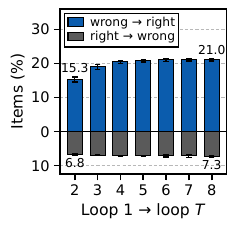}\caption{Changed since loop 1}\end{subfigure}\hfill
\begin{subfigure}[t]{0.245\textwidth}\centering\includegraphics[width=\linewidth]{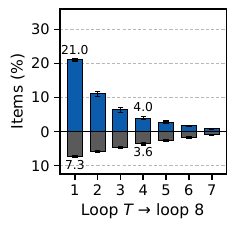}\caption{Still to change}\end{subfigure}\hfill
\begin{subfigure}[t]{0.245\textwidth}\centering\includegraphics[width=\linewidth]{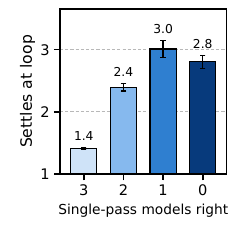}\caption{Harder items settle later}\end{subfigure}\hfill
\begin{subfigure}[t]{0.245\textwidth}\centering\includegraphics[width=\linewidth]{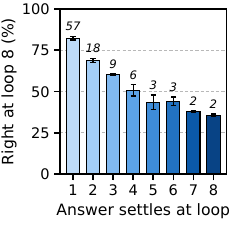}\caption{Late answers err more}\end{subfigure}
\caption{How \sansi{}'s answers change across loops (9,645 answerable items; mean $\pm$ s.d., three seeds). (a) Answers fixed or broken by loop $T$ relative to loop 1. (b) Answers that the loops after $T$ will still fix or break. (c) Mean settling loop by how many single-pass models (SmolLM2-1.7B, Qwen3.5-2B, Qwen3.5-4B) answer correctly. (d) Share of loop-8 answers that are right, by settling loop; above bars: share of items (\%). Values in Tables~\ref{tab:dynamics}, \ref{tab:settle} and~\ref{tab:difficulty}.}
\label{fig:dynamics}
\end{figure*}

\paragraph{Fixes outweigh breaks until loop 5.} On the 9,645 answerable items, loop 8 has fixed 21.0\% of the loop-1 answers and broken 7.3\%, about three fixes per break; loop 2 alone fixes 15.3\% and breaks 6.8\% (Figure~\ref{fig:dynamics}a,b; Table~\ref{tab:dynamics}). After loop 4, 90.2\% of the answers are final, and the remaining fixes (4.0\%) barely exceed the breaks (3.6\%); after loop 5 they are equal. Later loops change answers without improving them.

\paragraph{Harder items settle later, and late answers are unreliable.} An answer \emph{settles} at the first loop after which it no longer changes. 56.7\% of answers never change after loop 1, and 89.1\% have settled by loop 4. Settling tracks difficulty, measured independently by how many of the three single-pass models answer an item correctly (Figure~\ref{fig:dynamics}c; Table~\ref{tab:difficulty}): items solved by all three settle after 1.4 loops on average, by two after 2.4, by one after 3.0. (Items no single-pass model solves settle at 2.8, as \sansi{} often keeps its wrong first answer; Appendix~\ref{app:loops-settle}.) Settling also signals reliability (Figure~\ref{fig:dynamics}d; Table~\ref{tab:settle}): 82.2\% of the answers that never change are right, but only 35.6\% of those that settle at loop 8, and mean confidence falls (0.89 to 0.44).

\paragraph{Stopping early.} 11.7\% of the items are right at some loop but wrong at loop 8, which bounds what any loop-selection rule can gain (Appendix~\ref{app:loops-dynamics}). Of four stopping rules that we simulate on the stored probabilities (confidence, a stable answer, the KL divergence between loops, and Ouro's own exit gate; Appendix~\ref{app:stop}), only the gate and the KL rule beat a fixed budget, and the gate is the better of the two. The gate is a layer of the backbone that our training leaves frozen, and it needs no tuning: stopping once its cumulative exit probability passes one half, \sansi{} keeps the accuracy of loop 8 (72.0\% against 72.0\%; 0.0 [$-$0.3, 0.2]) with 4.5 loops on average, 44\% fewer than the full pass (cost 4.3 against 7.7; Table~\ref{tab:stop}), and thresholds from 0.2 to 0.7 give 71.7\% to 72.1\% with 3.5 to 5.4 loops. The gate also runs longer on harder items: 4.2 loops on average on the items that all three single-pass models solve, 4.7 on those that two solve and 4.9 on those that one solves (4.8 on those none solves), and 5.0 on unanswerable items against 4.5 on answerable ones (Appendix~\ref{app:stop}).

\begin{takeaway}
\takeawaylabel The loops work early: loops 2 and 3 bring 88\% of the gain, and 90\% of answers are final after loop 4. From loop 5 on, fixes and breaks balance. Easy items are decided at once, hard ones take more loops, and a late-settling answer is more often wrong than right. Ouro's frozen exit gate stops each item on its own, runs longer on harder items, and keeps the accuracy of loop 8 with 4.5 loops on average.
\end{takeaway}

\subsection{\rqlabel{3}: What do the loops do to the probabilities?}
\label{sec:analysis}

A caller accepts, escalates or rejects a decision according to its probability, so we track the probabilities loop by loop (Figure~\ref{fig:perloop}b--d; intervals in Table~\ref{tab:probci} in Appendix~\ref{app:probs}).

\paragraph{Confidence rises on right and wrong answers alike.} On the 8,879 items with one gold option, mean confidence rises from 0.769 to 0.869 on right answers and from 0.567 to 0.660 on wrong ones; the gap stays near 0.20. The AUROC for separating right from wrong answers improves only slightly (0.760 to 0.795; +0.035 [0.025, 0.045]). The loops make the model more accurate and more confident, but barely better at knowing when it is wrong, which is why confidence is a poor signal for stopping (\S\ref{sec:rq2}).

\paragraph{Calibration is best at loop 3.} While accuracy rises faster than confidence, the ECE falls (0.105 to 0.082 at loop 3; $-$0.023 [$-$0.032, $-$0.013]). Once the answers settle, confidence keeps rising and the ECE climbs back to 0.093 at loop 8 (+0.010 [0.005, 0.016]; Table~\ref{tab:confloop}), between SmolLM2-1.7B and the other single-pass models (Table~\ref{tab:main}). Averaging the probabilities of the loops removes most of this at no cost (\S\ref{sec:ablations}).

\paragraph{The loops detect missing evidence.} On the 382 items whose key evidence was removed (target: the uniform distribution), hard answers drop from 27.1\% at loop 1 to 17.5\% at loop 8 ($-$9.7 points [$-$13.4, $-$6.0]; Table~\ref{tab:unans}), and the evidence AUROC rises from 0.835 to 0.935 (+0.099 [0.075, 0.125]). The loops, not the training, cause this: the same backbone trained with one loop reaches 0.837, the value of loop 1. Of the single-pass models, only the two with three times the parameters, Qwen3.5-4B and Kev-4B (our data), are higher (0.948 and 0.947). See item 3 in Figure~\ref{fig:examples}.

\begin{takeaway}
\takeawaylabel Looping sharpens the probabilities where a decision needs a closer look at the evidence: \sansi{} detects missing evidence nearly as well as a model three times its size. It does not teach self-doubt: confidence rises on wrong answers too, error detection barely improves, and calibration peaks at loop 3.
\end{takeaway}

\begin{figure}[t]
\centering
\includegraphics[width=\columnwidth]{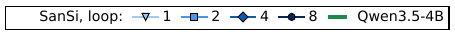}\\[1pt]
\begin{subfigure}[t]{0.5248\columnwidth}\includegraphics[width=\linewidth]{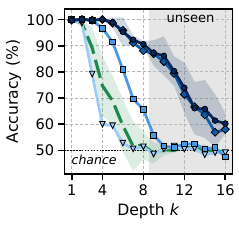}\caption{Liar chains}\end{subfigure}%
\begin{subfigure}[t]{0.4752\columnwidth}\includegraphics[width=\linewidth]{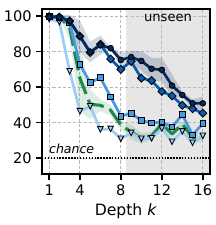}\caption{Object swaps}\end{subfigure}
\caption{Accuracy by depth $k$, with \sansi{} read after 1, 2, 4 and 8 loops. Shaded: depths unseen in training ($k > 8$). Dashed: single-pass Qwen3.5-4B trained on the same data. Mean of three seeds; bands: $\pm$1 s.d.\ for loop 8 and Qwen3.5-4B.}
\label{fig:depth}
\end{figure}

\subsection{\rqlabel{4}: Does looping buy reasoning depth?}
\label{sec:depth}

To isolate reasoning depth, we use two program-generated tasks in which every item needs exactly $k$ dependent steps (Figure~\ref{fig:depth}).

\paragraph{Tasks.} In a \emph{liar chain}, each person states that another tells the truth or lies, and the question is whether the last person tells the truth ($k$: chain length; chance 50\%). In \emph{object swaps}, modelled on the tracking-shuffled-objects task of BBH~\citep{suzgun2023bbh}, five people swap objects in pairs, and the question is who holds a given object at the end ($k$: number of times it changes hands; chance 20\%). Every item also contains $k$ irrelevant steps. We train \sansi{} and single-pass Qwen3.5-4B with our recipe (\S\ref{sec:method}) for 2,000 steps on $k \le 8$ and test on $k \le 16$ (120 items per depth, three seeds). Appendix~\ref{app:depth-tasks} gives examples and shortcut checks.

\paragraph{Each loop extends the reachable depth.} A model \emph{holds} depth $k$ if its accuracy is at least 75\% at every depth up to $k$. On liar chains, \sansi{} holds depth 3 after one loop, 6 after two and 11 after four, three steps beyond the longest training chain (Table~\ref{tab:depth-summary} in Appendix~\ref{app:depth-summary}). On the unseen depths (9--16), one loop is at chance (49.9\%) and eight loops reach 72.6\%. On object swaps, \sansi{} holds depth 2 after one loop and 9 after eight. Deeper items again settle later (Appendix~\ref{app:depth-settle}), and Ouro's exit gate follows the depth: with $q = 0.5$ it stops after 3.6 loops at $k = 1$, 6.5 at $k = 8$ and 7.7 at $k = 16$ on liar chains, and after 2.8, 4.8 and 6.2 on object swaps (Table~\ref{tab:depth-settle}; Figure~\ref{fig:gate-depth}).

\paragraph{Parameters do not replace loops.} Qwen3.5-4B holds depth 3 on both tasks, as \sansi{} does after one loop on liar chains and after two loops on object swaps. On the main suite, two loops of \sansi{} take less GPU time than one pass of Qwen3.5-4B (1.9 against 2.4; Tables~\ref{tab:stop} and~\ref{tab:main}). On the depth tasks, \sansi{} read after two loops is already more accurate on the trained depths of both tasks (+13.7 points [11.7, 15.6] on liar chains, +6.2 [4.1, 8.4] on object swaps; Table~\ref{tab:depth-summary}). On liar chains Qwen3.5-4B reaches 74.3\% on trained depths and chance (50.0\%) on unseen ones; \sansi{} at loop 8 leads by 22.7 points [20.8, 24.7] and 22.6 [20.5, 24.7]. On object swaps, \sansi{} leads by 20.2 [17.9, 22.6] and 29.5 [26.5, 32.7]. Longer training does not close the gap: trained for 8,000 steps, about the training time of \sansi{}, Qwen3.5-4B holds depth 5 and reaches 80.5\% on the trained depths of liar chains, but stays at chance on the unseen ones (51.5\%), where \sansi{} still leads by 21.2 points [18.6, 23.8] (Appendix~\ref{app:depth-long}; Figure~\ref{fig:depth-long}).

\begin{takeaway}
\takeawaylabel Looping buys reasoning depth, even beyond the training range, and on these tasks three times the parameters in a single pass do not replace it: Qwen3.5-4B leads by 1.8 points on the mixed suite but trails by more than 20 here, and four times the training leaves it at chance on unseen depths. Unseen depths remain unsolved (72.6\% on liar chains, 63.7\% on object swaps at loop 8), and running more than the eight trained loops does not help.
\end{takeaway}

\section{Examples and Errors}
\label{sec:examples}

\begin{figure*}[t]
\centering
\includegraphics[width=\textwidth]{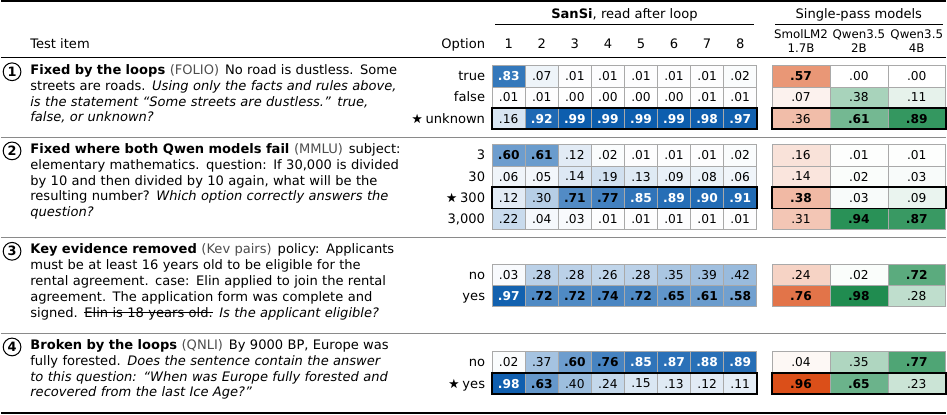}
\caption{Four test items: \sansi{}'s probability for each option after every loop, and the three single-pass models (seed 0). Correct option starred and outlined; column maxima in bold. Item 3 has no correct option: its deciding sentence (struck out) was removed. Full prompts in Appendix~\ref{app:examples}.}
\label{fig:examples}
\end{figure*}

\paragraph{Examples.} Figure~\ref{fig:examples} traces four items. In item 1, loop 1 follows a surface cue (as in Figure~\ref{fig:example}) and judges ``Some streets are dustless'' true (0.83); from loop 2 on, \sansi{} answers ``unknown'' (0.97 at loop 8). In item 2, loops 1--2 answer 3 and later loops 300, where both Qwen models answer 3,000. In item 3, whose deciding sentence was removed, loop 1 answers ``yes'' at 0.97; from loop 2 on the probability stays below the hard-answer threshold (0.72, falling to 0.58), while Qwen3.5-2B stays at 0.98. In item 4, the loops argue the model out of a right answer (``yes'': 0.98 to 0.11). Nor can the loops supply missing knowledge: a science question that Qwen3.5-4B answers correctly is wrong at every loop, with probability 0.96--0.99 (Table~\ref{tab:examples}).

\paragraph{Errors.} At loop 8, \sansi{} is wrong on 27.2\% of the 8,879 items with one gold option (Table~\ref{tab:errors} in Appendix~\ref{app:errors}). Most errors are not specific to looping: Qwen3.5-4B also fails on 68.0\% of them, and 60.9\% are wrong at every loop. One in five (19.7\%) carries a confidence of at least 0.9 and would pass any confidence threshold below 0.9, as expected from \S\ref{sec:analysis}.

\section{Ablations}
\label{sec:ablations}
\label{sec:training-choices}\label{sec:nloops}\label{sec:backbone}\label{sec:retrofit}\label{sec:larger}\label{sec:averaging}

\begin{figure}[t]
\centering
\includegraphics[width=0.95\columnwidth]{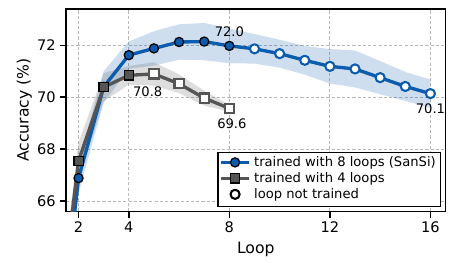}
\caption{Running beyond the trained loop size: accuracy from loop 2 on for \sansi{} (trained with 8 loops, run for 16) and a model trained with 4 (run for 8). Hollow: untrained loops; bands: $\pm$1 s.d., three seeds. Values in Table~\ref{tab:nloops}.}
\label{fig:nloops}
\end{figure}

We ablate how the loops are trained, how many to train and run, what the backbone contributes, and how the loops are read. All variants share \sansi{}'s data, recipe and seeds (Appendix~\ref{app:ablations}).

\begin{table*}[t]
\centering
\small
\setlength{\tabcolsep}{6pt}
\begin{tabular}{@{}l cccc cc@{}}
\toprule
 & \multicolumn{4}{c}{Accuracy (\%) read at loop} & \multicolumn{2}{c}{ECE at loop} \\
\cmidrule(lr){2-5}\cmidrule(lr){6-7}
Training & 1 & 2 & 4 & 8 & 4 & 8 \\
\midrule
\rowcolor{sansibg}\sansi{}: every loop, cross-entropy + Brier, $T{=}8$ & 58.4\sd{0.7} & 66.9\sd{0.6} & 71.6\sd{0.5} & 72.0\sd{0.7} & .085\sd{.009} & .093\sd{.012} \\
\rowcolor{groupbg}\multicolumn{7}{@{}l}{\emph{Loops that carry the loss}} \\
Loops 1, 2, 4, 8 only & 59.2\sd{0.8} & 67.0\sd{0.3} & 71.0\sd{0.2} & 71.2\sd{0.2} & .090\sd{.013} & .098\sd{.014} \\
Last loop only & 35.6\sd{4.7} & 51.9\sd{7.9} & 67.9\sd{1.5} & 70.9\sd{0.4} & .069\sd{.017} & .073\sd{.009} \\
Every loop, $T{=}4$ & 59.7\sd{0.5} & 67.6\sd{0.7} & 70.8\sd{0.4} & 69.6\sd{0.1} & .094\sd{.005} & .106\sd{.005} \\
\rowcolor{groupbg}\multicolumn{7}{@{}l}{\emph{The loss}} \\
Cross-entropy only (no Brier term) & 58.8\sd{0.2} & 67.1\sd{0.1} & 71.4\sd{0.4} & 71.6\sd{0.4} & .092\sd{.008} & .104\sd{.007} \\
Reinforcement learning & 58.1\sd{0.2} & 66.5\sd{0.4} & 71.3\sd{0.4} & 71.7\sd{0.1} & .068\sd{.015} & .076\sd{.015} \\
\bottomrule
\end{tabular}

\caption{Training choices: accuracy on the 10,027 test items and ECE, mean $\pm$ standard deviation over three seeds. Every variant differs from \sansi{} in one choice. For $T = 4$, loops 5--8 were never trained. For ``loops 1, 2, 4, 8 only'' (loss weight $1/4$ at each of these loops) and ``last loop only'', the loops without a loss are read with the frozen language-model head. ``Last loop only'' is the standard fine-tuning of the final output, with both terms of the loss. ``Cross-entropy only'': the loss of Equation~\ref{eq:loss} without the Brier term. ``Reinforcement learning'': the training with a reward of Appendix~\ref{app:rl}.}
\label{tab:ablation}
\end{table*}

\paragraph{Effect of the training objective.} With the loss on the last loop only, Ouro is fine-tuned the way a single-pass model is, through its final output alone. Its accuracy at loop 8 drops 1.1 points [0.7, 1.5] to 70.9\%, and its early loops collapse (35.6\% at loop 1, against 58.4\%; Table~\ref{tab:ablation}). The per-loop loss thus adds a point and, above all, makes every loop usable, which reading early and stopping early (\S\ref{sec:rq2}) rely on. The signal type matters less for accuracy: replacing supervision with outcome-only reinforcement learning after a 200-step supervised warm-up leaves accuracy unchanged (71.7\% against 72.0\%; $-$0.3 [$-$0.7, 0.1]; Appendix~\ref{app:rl}), and so does dropping the Brier term (71.6\%; $-$0.4 [$-$0.8, $-$0.0]). Training every loop costs some calibration at the last loop (ECE 0.093, against 0.073 with the last loop only; Appendix~\ref{app:train-loops}), which averaging the loops more than repairs (0.044; below). The Brier term matters for the probabilities: without it, the ECE at loop 8 rises from 0.093 to 0.104 (+0.011 [0.007, 0.016]), and the share of unanswerable items that receive a hard answer from 17.5\% to 20.0\% (+2.5 points [0.7, 4.3]; Appendix~\ref{app:train-loops}).

\paragraph{Effect of the number of loops.} Training four loops gives 70.8\% at loop 4, only 0.8 points [0.4, 1.2] below \sansi{} at loop 4 (Table~\ref{tab:nloops-ci}). Running past the trained loops hurts (Figure~\ref{fig:nloops}): \sansi{} falls from 72.0\% at loop 8 to 70.1\% at loop 16 ($-$1.8 [$-$2.2, $-$1.5]), and the four-loop model declines after loop 4. A looped model can be read early but not extended: the trained loops cap its compute.

\paragraph{Effect of the backbone.} Adding a loop to SmolLM2-1.7B, which was not pre-trained to loop, in either of two ways and training it with our recipe brings no gain (33.5\% and 58.3\% at loop 8, against 58.4\% without a loop; Table~\ref{tab:retrofit}). The gain rests on looped pre-training: our recipe turns a looped backbone into a decision model, but does not create the loops. A larger looped backbone helps further: \sansi{}-2.6B reaches 75.8\%, 2.0 points [1.4, 2.6] above Qwen3.5-4B with 63\% of its parameters (Table~\ref{tab:main}).

\paragraph{Effect of averaging the loops.} As confidence keeps rising after answers settle (\S\ref{sec:analysis}), averaging the option probabilities of the eight loops is a free fix (no labels or training). It keeps loop-8 accuracy (71.8\% against 72.0\%) and halves the ECE (0.044 against 0.093; $-$0.049 [$-$0.052, $-$0.045]; Table~\ref{tab:loopavg}), at no cost, as all loops are computed anyway.

\section{Use Case: \sansi{} as a Verifier}
\label{sec:verifier}

Typed decision models increasingly judge other models' outputs \citep{li2026judge}, and language-model judges are used where the quality of an output cannot be verified against a gold answer \citep{gan2026scaling}. We test if the loops matter when \sansi{} is the only reward for training a generator by reinforcement learning without gold answers.

\begin{figure}[t]
\centering
\includegraphics[width=0.95\columnwidth]{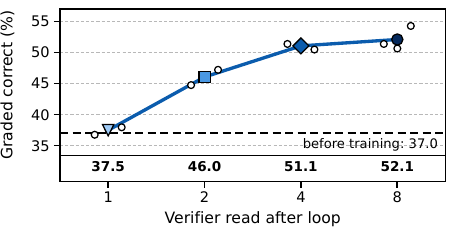}
\caption{GRPO on 2WikiMultiHopQA with \sansi{}'s probability as the only reward: share of the answers to the 3,000 test questions that are graded correct, by the loop at which \sansi{} is read (filled: mean of three seeds; hollow: seeds; dashed: before training). Values and F1 in Table~\ref{tab:grpo-ci}; exact match in Table~\ref{tab:grpo}.}
\label{fig:grpo}
\end{figure}

\paragraph{Setup.} We train SmolLM2-1.7B with LoRA and GRPO~\citep{shao2024grpo} on 2WikiMultiHopQA~\citep{ho2020twowiki}, not among \sansi{}'s data sources. Distinct sampled answers form the options of one typed call to the frozen \sansi{}, and an answer's reward is its probability; the gold answer is never used. We vary only the loop at which \sansi{} is read (1, 2, 4 or 8; three seeds each). For evaluation only, a language model (Claude Sonnet 5.5; Appendix~\ref{app:verifier-grading}) grades the greedy answer against the gold answer; on 500 answers graded a second time with a reworded instruction, 97.4\% receive the same label. We also report token F1 and exact match (Appendix~\ref{app:verifier}).

\paragraph{Result.} The generator follows the number of loops at which its verifier is read (Figure~\ref{fig:grpo}). Before training, 37.0\% of its answers are graded correct. Rewards read at loop 8 raise this share to 52.1\% (+15.1 points [13.3, 16.9]), at loop 4 to 51.1\% and at loop 2 to 46.0\%; rewards read at loop 1 leave it unchanged (37.5\%; +0.4 [$-$1.2, 2.2]). A likely cause is reward quality, a known weak point of language-model judges used as rewards \citep{kim2025evaluative}: in the first 20 training steps, the loop-1 reward separates gold-matching answers from the rest worse than the loop-4 or loop-8 reward (AUROC 0.78 against 0.91 and 0.90), so it more often ranks a wrong answer above a right one. Token F1 rises in the same order (39.5 before training, 47.3 at loop 8) but also penalises form: with the loop-1 reward the answers become twice as long, and F1 falls to 29.1 although as many answers are correct (Appendix~\ref{app:verifier-em}). As in the main results, loop 1 behaves like a single-pass model, and most of what the loops add is present by loop 4.

\section{Conclusion}

Looping lets a small typed decision model spend more computation on a decision instead of more parameters. With the same data and recipe, a looped 1.4B model is 13.5 points more accurate than a non-looped model of its shape and 5.3 points more accurate than a newer non-looped model of its size, and it comes within 1.8 points of a model with three times the parameters and a stronger backbone, at 3.3 times that model's GPU time and half its memory. Because every loop is read and trained, one model serves every budget from one loop to eight: three loops give most of the gain, most answers have settled by the fourth, and harder items settle later. The loops help the model notice missing evidence, and on two depth-controlled tasks they solve deeper problems than a single-pass model with three times the parameters, beyond the depths seen in training. Calibration improves only up to the third loop, since confidence keeps rising after the answers have settled, and running more loops than were trained lowers accuracy. These effects rest on a backbone that was pre-trained to loop. As the sole reward for training a generator, \sansi{} raises the share of correct answers from 37.0\% to 52.1\% when read after eight loops and leaves it unchanged when read after one. Ouro's exit gate already saves 44\% of the loops at the accuracy of loop 8, without tuning, and runs longer on harder and deeper items; a rule that chooses the loop for each item could go further, since 11.7\% of the items are right at some loop but wrong at the last.

\section*{Limitations}

Our findings hold within a defined scope. \sansi{} builds on one family of backbones that were pre-trained to loop, Ouro-1.4B and Ouro-2.6B, and most of our analyses use the 1.4B model. The gain of looping relies on this pre-training (\S\ref{sec:ablations}), so how far the findings carry over to other and larger looped backbones is a question for future work, as more such models become available; of our two sizes, the larger is the more accurate (Table~\ref{tab:main}). Looping buys accuracy with computation: a decision with eight loops takes about 3.3 times the GPU time of Qwen3.5-4B, and 1.6 times with four loops; \sansi{}-2.6B takes about 6.3 times. Its advantage in memory holds for one decision per call: answering several questions about the same state from a cached prefix would require the keys and values of every loop. We test stopping rules only on the stored probabilities (Appendix~\ref{app:stop}): with static batches the gate's saving falls from 44\% to about 25\% of the loops (Table~\ref{tab:stop-batch}), and serving in which finished items leave their batch is left to future work. The analysis of reasoning depth rests on two program-generated tasks. In both, deeper items are also longer, but even the deepest have at most 336 tokens, far fewer than the long documents of the main suite. The verifier case study uses one dataset and one generator; it is meant to illustrate one use of the loops, not to evaluate \sansi{} as a reward model in general. Finally, our suite is in English and is assembled from existing datasets. Its mix of item types is our choice, so we also report the results by type and by test group (Table~\ref{tab:main}; Figure~\ref{fig:types-gain}), and we cannot exclude that a backbone has seen some of these public datasets during pre-training. The models of the main comparison are trained with three seeds, so differences of one or two points on the smaller test groups should be read together with their intervals (Appendix~\ref{app:main}).

\section*{Ethical Considerations}

Typed decision models target automated decisions such as moderation and policy checks. Their probabilities invite automation, so their failure modes matter: we report where \sansi{} is overconfident (near transfer) and how often it answers without the evidence. All data come from public benchmarks; no new data were collected from people.

\bibliography{custom}

\appendix
\setcounter{dbltopnumber}{3}\setcounter{topnumber}{3}\setcounter{totalnumber}{5}
\renewcommand{\dbltopfraction}{0.9}\renewcommand{\topfraction}{0.9}\renewcommand{\textfraction}{0.08}
\renewcommand{\dblfloatpagefraction}{0.7}\renewcommand{\floatpagefraction}{0.7}

\section*{Appendix Overview}
The appendix follows the order of the paper. Appendix~\ref{app:data} describes the data, Appendix~\ref{app:training} the training and its cost, and Appendix~\ref{app:metrics} the metrics with their formulas. Appendices~\ref{app:main} to~\ref{app:depth} give additional results for the four research questions: the main comparison (Appendix~\ref{app:main}), the loops one by one (Appendix~\ref{app:loops}), the probabilities (Appendix~\ref{app:probs}) and the depth-controlled tasks (Appendix~\ref{app:depth}). Appendix~\ref{app:ablations} gives the details of the ablations, Appendix~\ref{app:verifier} those of the verifier case study, and Appendix~\ref{app:errors} those of the error analysis. Appendix~\ref{app:examples} gives examples: five items with the answers of \sansi{} and two single-pass models, and seven items followed through all loops.

\section{Data}
\label{app:data}

\paragraph{Item types and sources.} The training set draws on 20 sources and the test set on 59. Table~\ref{tab:data-examples} lists the main sources of each item type with one test item, Table~\ref{tab:data-groups} gives the sources of every type in training and in each test group with their numbers of items, and Table~\ref{tab:sources} lists every test source with the results on it. Type A (classification) uses topic, intent and sentiment datasets: AG News, DBpedia, Yelp and Amazon reviews \citep{zhang2015character}, IMDB \citep{maas2011imdb}, SST-5 \citep{socher2013sst}, TREC \citep{li2002trec} and Banking77 \citep{casanueva2020banking77}; its far-transfer sources are Emotion \citep{saravia2018carer}, TweetEval \citep{barbieri2020tweeteval} and Yahoo Answers \citep{zhang2015character}. Type B (multi-step reasoning) uses ProofWriter \citep{tafjord2021proofwriter}, CLUTRR \citep{sinha2019clutrr}, MuSiQue \citep{trivedi2022musique} and Kev's rule and policy items; near transfer adds longer CLUTRR chains, human paraphrases of ProofWriter and 4-hop MuSiQue questions; far transfer adds FOLIO \citep{han2022folio}, BBH \citep{suzgun2023bbh} and HoVer \citep{jiang2020hover}. Type C (uncertain evidence) uses MuSiQue minimal pairs (the same question with and without its key paragraph), SQuAD 2.0 \citep{rajpurkar2018squad2}, ChaosNLI \citep{nie2020chaosnli}, Kev's unknowable pairs and Sys1Cal \citep{porcedda2026sys1cal}. Type D (long documents) uses MuSiQue and HotpotQA \citep{yang2018hotpotqa} with distractor passages and, for far transfer, ContractNLI \citep{koreeda2021contractnli} and QuALITY \citep{pang2022quality}. Type E (sentence pairs) uses MNLI \citep{williams2018mnli} and BoolQ \citep{clark2019boolq} and, for far transfer, ANLI \citep{nie2020anli}, PAWS \citep{zhang2019paws}, WANLI \citep{liu2022wanli} and QNLI \citep{wang2019glue}. Type F (knowledge, test only) uses ARC \citep{clark2018arc}, CommonsenseQA \citep{talmor2019commonsenseqa}, MMLU \citep{hendrycks2021mmlu}, MMLU-Pro \citep{wang2024mmlupro} and SciQ \citep{welbl2017sciq}.

\begin{table*}[t]
\centering
\small
\setlength{\tabcolsep}{4pt}
\begin{tabular}{@{}>{\raggedright\arraybackslash}p{2.35cm}>{\raggedright\arraybackslash}p{4.75cm}>{\raggedright\arraybackslash}p{5.9cm}rr@{}}
\toprule
Item type & Sources (selection) & Example decision & Train & Test \\
\midrule
A\ \ Classification & AG News, DBpedia, Yelp \citep{zhang2015character}, IMDB \citep{maas2011imdb}, SST-5 \citep{socher2013sst}, TREC \citep{li2002trec}, Banking77 \citep{casanueva2020banking77}, TweetEval \citep{barbieri2020tweeteval} & ``Meh. I was unimpressed.'' \emph{Would this reviewer recommend the business?} (A) no \ (B) yes \ $\rightarrow$ A & 2,400 & 1,089 \\
\addlinespace
B\ \ Multi-step reasoning & CLUTRR \citep{sinha2019clutrr}, ProofWriter \citep{tafjord2021proofwriter}, MuSiQue \citep{trivedi2022musique}, FOLIO \citep{han2022folio}, BBH \citep{suzgun2023bbh} & ``No homework is fun. Some reading is homework.'' \emph{Is the statement ``Some reading is fun.'' true, false, or unknown?} (A) true \ (B) false \ (C) unknown \ $\rightarrow$ C & 5,500 & 3,072 \\
\addlinespace
C\ \ Uncertain evidence & ChaosNLI \citep{nie2020chaosnli}, Sys1Cal \citep{porcedda2026sys1cal}, question pairs with the evidence removed (from MuSiQue) & ``A hockey fight.'' \emph{Hypothesis: ``fighting on the ice''.} (A) entailment \ (B) neutral \ (C) contradiction \ $\rightarrow$ the annotators' label distribution & 2,300 & 1,558 \\
\addlinespace
D\ \ Long documents & HotpotQA \citep{yang2018hotpotqa}, ContractNLI \citep{koreeda2021contractnli}, QuALITY \citep{pang2022quality} & [two encyclopedia paragraphs] \emph{Are Anja Salomonowitz and Rod Lurie both directors?} (A) no \ (B) yes \ $\rightarrow$ B & 1,000 & 901 \\
\addlinespace
E\ \ Sentence pairs & MNLI \citep{williams2018mnli}, BoolQ \citep{clark2019boolq}, ANLI \citep{nie2020anli}, WANLI \citep{liu2022wanli}, PAWS \citep{zhang2019paws}, QNLI \citep{wang2019glue} & ``Revco was acquired in 1997 by CVS.'' \emph{Does this sentence mean the same thing: ``Revco was subsequently acquired by CVS in 1997.''} (A) no \ (B) yes \ $\rightarrow$ B & 1,600 & 1,368 \\
\addlinespace
F\ \ Knowledge & ARC \citep{clark2018arc}, CommonsenseQA \citep{talmor2019commonsenseqa}, MMLU \citep{hendrycks2021mmlu}, SciQ \citep{welbl2017sciq} & \emph{Coal is formed from} (A) seas that have evaporated \ \dots{} \ (D) plant remains decomposed under pressure \ $\rightarrow$ D & -- & 1,808 \\
\midrule
Total & & & 12,800 & 10,027 \\
\bottomrule
\end{tabular}
\caption{Item types, a selection of their sources, and one test item of each type (state, \emph{question}, options and target). The test total includes the 231 public items of JevBench.}
\label{tab:data-examples}
\end{table*}

\begin{table*}[t]
\centering
\footnotesize
\setlength{\tabcolsep}{4pt}
\begin{tabular}{@{}>{\raggedright\arraybackslash}p{2.15cm}>{\raggedright\arraybackslash}p{5.15cm}>{\raggedright\arraybackslash}p{3.0cm}>{\raggedright\arraybackslash}p{4.35cm}@{}}
\toprule
Item type & Seen in training (training / in-distribution test items) & Near transfer & Far transfer \\
\midrule
A\ \ Classification & \textbf{2,400 / 321}: AG News$^{\dagger}$ 300/41; Amazon$^{\dagger}$ 300/40; Banking77$^{\dagger}$ 300/40; DBpedia$^{\dagger}$ 300/40; IMDB$^{\dagger}$ 300/40; SST-5$^{\dagger}$ 300/40; TREC$^{\dagger}$ 300/40; Yelp$^{\dagger}$ 300/40 & -- & \textbf{768}: Emotion 240; TweetEval 240; Yahoo Answers 160; Emotion$^{\ddagger}$ 64; Offensive tweets$^{\ddagger}$ 64 \\
\addlinespace[3pt]
B\ \ Multi-step reasoning & \textbf{5,500 / 800}: ProofWriter 1,500/240; CLUTRR, 2--4 hops 1,200/160; Kev rules$^{\dagger}$ 1,110/120; MuSiQue, 2--3 hops 1,000/160; Kev policies$^{\dagger}$ 690/120 & \textbf{991}: CLUTRR, 5--10 hops 480; ProofWriter, paraphrased 319; MuSiQue, 4 hops 192 & \textbf{1,281}: BBH 480; HoVer 360; FOLIO 320; Kev policies, held out$^{\ddagger}$ 64; Kev rules, held out$^{\ddagger}$ 57 \\
\addlinespace[3pt]
C\ \ Uncertain evidence & \textbf{2,300 / 478}: MuSiQue pairs, 2--3 hops 1,000/158; SQuAD 2.0 800/240; ChaosNLI (SNLI, $\alpha$NLI) 500/80 & \textbf{640}: ChaosNLI (MNLI) 400; MuSiQue pairs, 4 hops 240 & \textbf{440}: Sys1Cal 292; Kev unknowable pairs$^{\ddagger}$ 148 \\
\addlinespace[3pt]
D\ \ Long documents & \textbf{1,000 / 357}: MuSiQue with distractors 600/237; HotpotQA 400/120 & -- & \textbf{544}: ContractNLI 240; QuALITY 240; Kev buried evidence$^{\ddagger}$ 64 \\
\addlinespace[3pt]
E\ \ Sentence pairs & \textbf{1,600 / 160}: BoolQ$^{\dagger}$ 800/80; MNLI$^{\dagger}$ 800/80 & \textbf{240}: MNLI, mismatched genres 240 & \textbf{968}: ANLI 240; PAWS 200; QNLI 200; WANLI 200; PAWS$^{\ddagger}$ 64; QNLI$^{\ddagger}$ 64 \\
\addlinespace[3pt]
F\ \ Knowledge & -- & -- & \textbf{1,808}: MMLU 480; ARC-Challenge 320; CommonsenseQA 240; MMLU-Pro 240; MMLU-Pro$^{\ddagger}$ 160; ARC-Easy 120; SciQ 120; MMLU$^{\ddagger}$ 64; SciQ$^{\ddagger}$ 64 \\
\midrule
Total & \textbf{12,800 / 2,116} & \textbf{1,871} & \textbf{5,809} \\
\bottomrule
\end{tabular}

\caption{The sources of every item type in training and in the three test groups, with their numbers of items (bold: the total of the type). For a source seen in training, the two numbers are its training items and its in-distribution test items; near transfer consists of harder or reworded versions of these sources, far transfer of sources that are never trained on. $^{\dagger}$: from Kev's training suite; $^{\ddagger}$: from Kev's transfer suite. The 231 public JevBench items (48 easy, 72 standard, 111 hard) form a fourth test group.}
\label{tab:data-groups}
\end{table*}

\paragraph{Unanswerable and crowd-labelled items.} An unanswerable item is an item whose key evidence has been removed; it has no correct option and its target is the uniform distribution. A crowd-labelled item (ChaosNLI, about 100 annotators per item; Sys1Cal, whose items state exact probabilities) has the label distribution as its target, and its accuracy uses the most probable label.

\paragraph{Splits and leakage controls.} No model was called while building the suite. Test candidates that share a state with a training item, or 80\% or more of its sentences, were not drawn. Development and test items were split by connected components of related items (items that share a group, state, story or sub-question), so that related items stay on one side. No two items have the same text.

\paragraph{Prompt format and number of options.} Every item is rendered as the state, an empty line, ``Question:'' with the question, ``Options:'' with the options as ``(A) \dots{} (B) \dots'', and ``Answer:''; Appendix~\ref{app:examples} shows complete prompts. The model is read at the last token. An option is named by one letter, which is one token, so an item can have at most 26 options. In our suite, 76\% of the test items and 78\% of the training items have two to four options; the largest sets are the 14 classes of DBpedia, the 18 kinship relations of CLUTRR and the 26 intents of Kev's Banking77 items.

\section{Training and Implementation Details}
\label{app:training}

\paragraph{Optimisation.} AdamW with $\beta = (0.9, 0.95)$ and no weight decay; learning rate $10^{-4}$ for the LoRA adapters and $10^{-3}$ for the readouts; 200 warm-up steps followed by a cosine decay to 10\% of the peak; gradient clipping at 1.0; bfloat16 autocast. A step is the next 16 items of the shuffled training set, so 1,000 steps cover 16,000 items (1.25 passes). Items are never truncated; the longest training item has 5,345 tokens. The option order of multiple-choice items is shuffled each time an item is drawn. LoRA uses rank 64, $\alpha = 128$ and dropout 0.05 on the query, key, value and output projections of attention and on the three projections of the feed-forward block (for Qwen3.5-4B also on the projections of its linear-attention layers). The two terms of the loss (Equation~\ref{eq:loss}) have equal weight in all our models except one variant trained with cross-entropy alone (Table~\ref{tab:ablation}; Appendix~\ref{app:train-loops}).

\paragraph{Parameters and time.} Table~\ref{tab:params} lists the sizes and the measured times. The test times are measured in one way for all models: one pass over the 10,027 test items on a single RTX A6000, in one process, with the same batching (items sorted by length, at most 32,000 tokens per batch), after a warm-up that is not counted. The time of a forward pass depends on the architecture and not on the values of the weights, so the models are timed with the adapters and the readout attached as in training; for Ouro-1.4B with one loop this gives 334 seconds, against 337 seconds in the test of the trained model. The Qwen3.5 models run in their own software environment (PyTorch 2.7.1 with the optimised kernel for their linear-attention layers; their convolution falls back to the reference implementation). \sansi{}-2.6B was timed on a second machine with the same card, on which Ouro-1.4B with one loop takes 316 seconds and \sansi{} 2,485 seconds; differences of a few per cent are therefore within the variation between machines. The training times are those of the original runs. They were not all made on the same cards and include the evaluations on the development set, so they are comparable only roughly; the comparison that the paper uses, \sansi{} against Ouro-1.4B with one loop, was made on the same cards with the same number of evaluations. The same passes record the peak GPU memory. Like the times, the memory of the Qwen3.5 models and of Kev-4B (our data) reflects their own software environment and model code.

\paragraph{Kev-4B trained on our data.} Kev-4B (our data) is trained with the training code of Kev \citep{palmer2026kev}, with the command of the base stage of the released Kev-4B: Qwen3.5-4B-Base, LoRA adapters of rank 16 (33.8M parameters) and Kev's pointer head, cross-entropy, a learning rate of $5 \times 10^{-5}$ with a one-cycle schedule, weight decay 0.01, and two epochs (3,200 steps of eight records) with Kev's augmentation of the options. Three things are changed, none of them in the loss, the optimiser or the augmentation: the training data are our 12,800 items, written as Kev requests (the target distributions of crowd-labelled and unanswerable items are passed as Kev's targets); the limit on the length of a state is raised so that no item is dropped; and long batches are run in several passes. We train three seeds, each for about 3.9 hours on one RTX A6000. Kev's recipe ends by fitting one temperature on development data. In the main comparison we read the model at temperature 1, like every other model, and Appendix~\ref{app:probs-cal} fits such a temperature for this model, Qwen3.5-4B and both sizes of \sansi{} (Table~\ref{tab:calib-dev}). The test time in Table~\ref{tab:params} is measured in the setting described above, with Kev's own model code and request format; with the adapters merged into the weights, as Kev serves its models, the pass takes 567 seconds.

\begin{table}[h]
\centering
\small
\setlength{\tabcolsep}{3pt}
\resizebox{\columnwidth}{!}{
\begin{tabular}{@{}lrrrrr@{}}
\toprule
 & Backbone & Trained & Training & Test & Test \\
Model & params & params & (GPU-min) & (GPU-s) & memory (GB) \\
\midrule
SmolLM2-1.7B & 1.71B & 72.6M & 39$^{\dagger}$ & 382 & 6.5 \\
Ouro-1.4B, one loop & 1.43B & 60.8M & 33 & 334 & 5.3 \\
\sansi{}, $T = 8$ & 1.43B & 60.8M & 313 & 2,571 & 6.2 \\
\sansi{}-2.6B, $T = 8$ & 2.67B & 121.4M & 558$^{\S}$ & 4,963 & 8.8 \\
Qwen3.5-2B & 1.88B & 67.5M & 48$^{\ddagger}$ & 387 & 6.5 \\
Qwen3.5-4B & 4.21B & 130.2M & 75 & 789 & 12.9 \\
Kev-4B (our data) & 4.21B & 33.8M & 234$^{\|}$ & 707 & 16.1 \\
\bottomrule
\end{tabular}}
\caption{Sizes and measured cost. Test: one pass over the 10,027 test items on one RTX A6000, the same setting for all models. Memory: the peak GPU memory allocated during that pass. Training: 1,000 steps on RTX A6000 cards (wall-clock time times the number of cards), except $^{\dagger}$ RTX A5500 and $^{\ddagger}$ two RTX A5000; the times include five evaluations on the development set ($^{\S}$ two). $^{\|}$: Kev's recipe, two epochs (3,200 steps), without evaluations.}
\label{tab:params}
\end{table}

\paragraph{The readout correction is small.} For the runs whose records keep both readings, the accuracy read with the frozen language-model head alone differs from the accuracy of the trained readout by at most 0.04 points (every loop of the $T = 4$ model, SmolLM2-1.7B and Qwen3.5-4B) and by at most 0.12 points for an eight-loop model trained on 80\% of the data.

\section{Metrics}
\label{app:metrics}

For an item with $K$ options, let $p \in \Delta^{K}$ be the distribution that the model returns at the loop that is read, $\hat{y} = \arg\max_{k} p_{k}$ its answer and $c = \max_{k} p_{k}$ its confidence. Let $d$ be the target distribution of the item. Probabilities are used as they are, without post-hoc calibration, unless stated otherwise.

\paragraph{Correctness and accuracy.} An item with a gold option $y$ is answered correctly when $\hat{y} = y$; for a crowd-labelled item, $y$ is the most probable label under $d$. An unanswerable item has no gold option. It is answered correctly when the model gives no \emph{hard answer}, that is, when
\begin{equation}
c < \tfrac{1}{2}\big(1 + \tfrac{1}{K}\big),
\end{equation}
the midpoint between the uniform distribution ($c = 1/K$) and certainty ($c = 1$). Accuracy is the share of items answered correctly, and the hard-answer rate is the share of unanswerable items that receive a hard answer.

\paragraph{Calibration.} ECE is computed on the $N$ answerable items. They are sorted by their confidence into ten bins $B_{1}, \dots, B_{10}$ of equal width, and
\begin{equation}
\mathrm{ECE} = \sum_{b=1}^{10} \frac{|B_{b}|}{N}\, \big|\, \mathrm{acc}(B_{b}) - \mathrm{conf}(B_{b}) \,\big| ,
\end{equation}
where $\mathrm{acc}(B_{b})$ is the accuracy and $\mathrm{conf}(B_{b})$ the mean confidence of the items in bin $b$. Overconfidence is the mean confidence minus the accuracy. Confident errors are the share of answerable items that are wrong with $c \ge 0.8$. NLL is the mean of $-\log p_{y}$ over the items with a single gold option. For crowd-labelled items we also report the total variation distance $\frac{1}{2} \sum_{k} |p_{k} - d_{k}|$.

\paragraph{AUROC.} For two sets of items $P$ and $Q$, let $s_{ij}$ be 1 if $c_{i} > c_{j}$, $\frac{1}{2}$ if $c_{i} = c_{j}$ and 0 otherwise. Then
\begin{equation}
\mathrm{AUROC}(P, Q) = \frac{1}{|P|\,|Q|} \sum_{i \in P} \sum_{j \in Q} s_{ij} .
\end{equation}
The right/wrong AUROC takes as $P$ and $Q$ the correctly and the wrongly answered items among those with a single gold option. The evidence AUROC takes the answerable and the unanswerable items of the four sources that contain both: MuSiQue minimal pairs (in-distribution and 4-hop), SQuAD 2.0 and Kev's unknowable pairs (786 items, 382 of them unanswerable). It is 1 when every answerable item receives a higher confidence than every unanswerable one, and 0.5 when confidence says nothing about missing evidence.

\paragraph{Intervals.} For the difference between two models we resample groups of related items with replacement 2,000 times and report the 2.5th and 97.5th percentiles of the difference. A resample is applied to all seeds and to both models of a comparison. For the ECE, the AUROC and the hard-answer rate, the statistic is recomputed on every resample and averaged over the seeds.

\paragraph{Depth-controlled tasks.} We report the accuracy at every depth $k$. The depth up to which a model holds is the largest $k$ such that its accuracy is at least 75\% at every depth from 1 to $k$. The loop at which an answer settles is the last loop whose answer differs from that of the loop before (1 if the answer never changes).

\paragraph{Verifier case study.} A generated answer is compared with the gold answer after both have been put in lower case and stripped of punctuation and of the articles \emph{a}, \emph{an} and \emph{the}. Exact match (EM) is 1 if the two are then equal and 0 otherwise. F1 is the harmonic mean of the precision and the recall of the words of the generated answer against the words of the gold answer. When a question has several accepted gold answers, the best match counts. ``Contains'' is the share of answers in which the gold answer occurs as a sequence of whole words.

\raggedbottom
\setcounter{dbltopnumber}{3}
\renewcommand{\dbltopfraction}{0.9}
\renewcommand{\dblfloatpagefraction}{0.7}
\renewcommand{\topfraction}{0.9}
\renewcommand{\textfraction}{0.07}
\section{Main Comparison: Additional Results}
\label{app:main}
\label{app:more}

This appendix supports \S\ref{sec:rq1}. It gives the full version of the main table (Appendix~\ref{app:main-all}), the intervals of the differences that the paper reports (Appendix~\ref{app:main-ci}), the comparison with the model of the same shape in detail (Appendix~\ref{app:main-shape}), accuracy on the answerable items and with other thresholds for a hard answer (Appendix~\ref{app:main-abstain}), the results on every test source (Appendix~\ref{app:main-sources}) and by item type (Appendix~\ref{app:main-types}), and the details of the comparisons with the Qwen3.5 models (Appendix~\ref{app:main-qwen}), with Kev's recipe on our data (Appendix~\ref{app:main-kev}) and with the released Kev-4B and the commercial API (Appendix~\ref{app:main-api}).

\subsection{All models}
\label{app:main-all}

Table~\ref{tab:main-full} extends Table~\ref{tab:main} of the main text in four ways. First, it gives the backbones before fine-tuning, read with the frozen language-model head. Qwen3.5-4B is the strongest backbone before any training (61.6\%), 13.7 points above Ouro-1.4B at its best loop (47.9\% at loop 4); SmolLM2-1.7B is the weakest (38.8\%). Fine-tuning with our recipe adds 19.6 points to SmolLM2-1.7B, 17.1 to Qwen3.5-2B and 12.2 to Qwen3.5-4B. Second, it contains two further rows of looped models: the model trained with four loops, which \S\ref{sec:ablations} discusses, and \sansi{} read at loop 4 (\sansi{} read at loop 3, a row of Table~\ref{tab:main}, is in Table~\ref{tab:perloop} with every other loop). Third, it adds two columns: the accuracy on the 231 public JevBench items, and the share of unanswerable items to which a model gives a hard answer. Fourth, it gives two released models as references, Kev-4B and the commercial Jev API; Appendix~\ref{app:main-api} explains why they can be compared with our models only on a part of the test set.

\begin{table*}[t]
\centering
\resizebox{\textwidth}{!}{
\begin{tabular}{@{}llcc ccccc ccc@{}}
\toprule
 & & & & \multicolumn{5}{c}{Accuracy (\%) $\uparrow$} & ECE $\downarrow$ & Hard ans. & Evidence \\
\cmidrule(lr){5-9}
Model & Params & Loops & Cost & All & In-dist. & Near & Far & JevBench & (all) & (\%) $\downarrow$ & AUROC $\uparrow$ \\
\midrule
\rowcolor{groupbg}\multicolumn{12}{@{}l}{\emph{Not fine-tuned}} \\
\mkSmol{}SmolLM2-1.7B & 1.7B & 1 & -- & 38.8 & 41.4 & 30.6 & 40.5 & 36.8 & .106 & 15.2 & .547 \\
\mkOuro{}Ouro-1.4B & 1.4B & 4 & -- & 47.9 & 49.8 & 37.4 & 50.6 & 48.1 & .079 & 34.0 & .587 \\
\mkQwenS{}Qwen3.5-2B & 1.9B & 1 & -- & 49.6 & 48.0 & 38.4 & 53.5 & 55.8 & .180 & 55.5 & .572 \\
\mkQwen{}Qwen3.5-4B & 4.2B & 1 & -- & 61.6 & 60.4 & 51.1 & 65.2 & 66.7 & .067 & 52.4 & .652 \\
\midrule
\rowcolor{groupbg}\multicolumn{12}{@{}l}{\emph{Fine-tuned, single pass (same data, recipe and seeds)}} \\
\mkSmol{}SmolLM2-1.7B & 1.7B & 1 & 1.1 & 58.4\sd{0.7} & 76.1\sd{0.8} & 57.7\sd{0.6} & 52.3\sd{0.7} & 58.3\sd{0.7} & \textbf{.069}\sd{.020} & 36.3\sd{4.2} & .765\sd{.009} \\
\mkOuro{}Ouro-1.4B, one loop & 1.4B & 1 & 1.0 & 58.6\sd{0.2} & 77.3\sd{1.0} & 59.9\sd{0.9} & 51.3\sd{0.4} & 59.7\sd{3.0} & .137\sd{.009} & 31.9\sd{2.2} & .837\sd{.007} \\
\mkQwenS{}Qwen3.5-2B & 1.9B & 1 & 1.2 & 66.7\sd{0.7} & 84.3\sd{1.4} & 62.3\sd{2.5} & 61.7\sd{0.4} & 67.5\sd{1.5} & .123\sd{.007} & 27.9\sd{1.0} & .896\sd{.002} \\
\mkQwen{}Qwen3.5-4B & 4.2B & 1 & 2.4 & 73.8\sd{0.6} & 88.5\sd{0.4} & 68.9\sd{2.4} & 70.1\sd{0.3} & 73.2\sd{1.9} & .113\sd{.003} & 18.2\sd{2.5} & \textbf{.948}\sd{.003} \\
\midrule
\rowcolor{groupbg}\multicolumn{12}{@{}l}{\emph{Trained on the same data with Kev's recipe, single pass}} \\
\mkKev{}Kev-4B (our data) & 4.2B & 1 & 2.1 & 74.3\sd{0.3} & \textbf{88.6}\sd{0.2} & 70.9\sd{0.7} & 70.3\sd{0.6} & 73.6\sd{0.4} & .119\sd{.004} & \textbf{16.3}\sd{0.6} & .947\sd{.002} \\
\midrule
\rowcolor{groupbg}\multicolumn{12}{@{}l}{\emph{Fine-tuned, looped (\sansi{} is trained with 8 loops unless noted)}} \\
\rowcolor{sansibg}\mkSansi{}\sansi{}, trained with 4 loops & 1.4B & 4 & 3.8 & 70.8\sd{0.4} & 86.4\sd{0.5} & 64.5\sd{2.4} & 67.3\sd{0.4} & 69.3\sd{0.7} & .094\sd{.005} & 20.6\sd{1.7} & .931\sd{.002} \\
\rowcolor{sansibg}\mkSansi{}\sansi{}, read at loop 4 & 1.4B & 4 & 3.8 & 71.6\sd{0.5} & 86.5\sd{0.5} & 67.2\sd{2.0} & 67.6\sd{0.5} & 72.6\sd{1.2} & .085\sd{.009} & 18.1\sd{0.5} & .928\sd{.003} \\
\rowcolor{sansibg}\mkSansi{}\sansi{}, read at loop 8 & 1.4B & 8 & 7.7 & 72.0\sd{0.7} & 86.6\sd{0.6} & 67.7\sd{2.7} & 68.0\sd{0.1} & 72.3\sd{1.6} & .093\sd{.012} & 17.5\sd{1.3} & .935\sd{.004} \\
\rowcolor{sansibg}\mkSansiL{}\sansi{}-2.6B, read at loop 8 & 2.7B & 8 & 14.8 & \textbf{75.8}\sd{0.6} & 88.4\sd{0.8} & \textbf{74.4}\sd{2.5} & \textbf{71.6}\sd{0.3} & \textbf{79.4}\sd{0.7} & .078\sd{.000} & 18.2\sd{2.4} & .943\sd{.000} \\
\midrule
\rowcolor{groupbg}\multicolumn{12}{@{}l}{\emph{Reference: released models, not trained on our data}} \\
\mkKevR{}Kev-4B (released) & 4.2B & 1 & -- & 68.2 & 70.6 & 53.1 & 71.9 & 74.5 & .036 & 29.1 & .742 \\
\mkJev{}Jev (jev-1.13.0) & n/a & 1 & -- & 78.9 & 77.5 & 65.7 & 83.3 & 87.9 & .066 & 62.8 & .743 \\
\bottomrule
\end{tabular}
}
\caption{Full results on the 10,027 test items (mean $\pm$ standard deviation over three seeds; the untuned backbones, the released Kev-4B and the Jev API are single runs; bold: best fine-tuned model). All fine-tuned models share the training data, recipe and seeds; Loops is the number of loops run at test time. Cost: GPU time of one pass over the test set, relative to Ouro-1.4B with one loop (Table~\ref{tab:params}; not measured for the untuned backbones and the released models). Hard ans.: share of the 382 unanswerable items that receive a hard answer. Evidence AUROC: separation of answerable from unanswerable items by confidence (786 items). The low hard-answer rate of the untuned SmolLM2-1.7B reflects its low confidence on all items. The released Kev-4B is read at the temperature it ships with. Shaded rows: looped models (\sansi{}); squares: model colours of the figures.}
\label{tab:main-full}
\end{table*}

\subsection{Intervals of the differences}
\label{app:main-ci}

Table~\ref{tab:ci} lists the differences between models and between loops that the paper reports, with their 95\% bootstrap intervals, for all test items and for every test group. The intervals come from resampling groups of related test items (Appendix~\ref{app:metrics}); the three seeds of a model are averaged. An interval therefore says how much a difference depends on the choice of test items. It does not cover the variation between training runs, which Table~\ref{tab:seeds} shows.

\begin{table*}[t]
\centering
\resizebox{\textwidth}{!}{
\begin{tabular}{@{}l ccccc@{}}
\toprule
Difference (points) & All & In-dist. & Near & Far & JevBench \\
\midrule
\sansi{} L8 $-$ SmolLM2-1.7B & \dbetter{+13.5}{+12.5}{+14.5} & \dbetter{+10.5}{+8.8}{+12.2} & \dbetter{+10.0}{+7.4}{+12.6} & \dbetter{+15.8}{+14.5}{+17.0} & \dbetter{+14.0}{+8.5}{+19.5} \\
\sansi{} L8 $-$ Ouro one loop & \dbetter{+13.4}{+12.4}{+14.4} & \dbetter{+9.3}{+7.7}{+11.0} & \dbetter{+7.9}{+5.7}{+10.1} & \dbetter{+16.7}{+15.3}{+18.0} & \dbetter{+12.6}{+7.5}{+17.6} \\
Ouro one loop $-$ SmolLM2-1.7B & \dns{+0.1}{-0.7}{+1.0} & \dns{+1.2}{-0.1}{+2.5} & \dbetter{+2.2}{+0.2}{+4.2} & \dns{-0.9}{-2.2}{+0.3} & \dns{+1.4}{-3.2}{+6.1} \\
\sansi{} L1 $-$ Ouro one loop & \dns{-0.2}{-0.6}{+0.3} & \dns{-0.9}{-1.9}{+0.0} & \dbetter{+1.2}{+0.1}{+2.4} & \dns{-0.5}{-1.1}{+0.1} & \dbetter{+4.8}{+1.9}{+7.8} \\
\sansi{} $T{=}4$ (L4) $-$ SmolLM2-1.7B & \dbetter{+12.4}{+11.4}{+13.4} & \dbetter{+10.3}{+8.5}{+12.1} & \dbetter{+6.8}{+3.8}{+9.8} & \dbetter{+15.0}{+13.8}{+16.3} & \dbetter{+11.0}{+5.3}{+16.2} \\
Qwen3.5-4B $-$ \sansi{} $T{=}4$ (L4) & \dbetter{+3.0}{+2.4}{+3.6} & \dbetter{+2.1}{+1.0}{+3.2} & \dbetter{+4.5}{+3.0}{+6.1} & \dbetter{+2.8}{+1.9}{+3.7} & \dbetter{+3.9}{+0.4}{+7.5} \\
Qwen3.5-4B $-$ \sansi{} L8 & \dbetter{+1.8}{+1.2}{+2.5} & \dbetter{+1.9}{+0.9}{+2.9} & \dns{+1.2}{-0.3}{+2.8} & \dbetter{+2.1}{+1.2}{+2.9} & \dns{+0.9}{-2.0}{+4.0} \\
Qwen3.5-4B $-$ \sansi{} L3 & \dbetter{+3.4}{+2.8}{+4.1} & \dbetter{+2.9}{+1.8}{+4.0} & \dbetter{+1.9}{+0.4}{+3.6} & \dbetter{+4.1}{+3.3}{+5.0} & \dns{+2.0}{-1.4}{+5.6} \\
Qwen3.5-4B $-$ SmolLM2-1.7B & \dbetter{+15.4}{+14.4}{+16.4} & \dbetter{+12.4}{+10.8}{+14.1} & \dbetter{+11.2}{+8.6}{+14.2} & \dbetter{+17.8}{+16.6}{+19.1} & \dbetter{+14.9}{+9.4}{+20.5} \\
\sansi{} L3 $-$ SmolLM2-1.7B & \dbetter{+12.0}{+11.0}{+12.9} & \dbetter{+9.5}{+7.8}{+11.0} & \dbetter{+9.3}{+6.6}{+12.5} & \dbetter{+13.7}{+12.4}{+15.0} & \dbetter{+12.8}{+7.6}{+18.2} \\
\sansi{} L2 $-$ Qwen3.5-2B & \dns{+0.2}{-0.5}{+0.8} & \dworse{-1.4}{-2.6}{-0.2} & \dbetter{+1.9}{+0.1}{+3.6} & \dns{+0.2}{-0.6}{+1.2} & \dns{-0.6}{-4.5}{+3.0} \\
\sansi{} L8 $-$ Qwen3.5-2B & \dbetter{+5.3}{+4.5}{+6.0} & \dbetter{+2.3}{+1.1}{+3.5} & \dbetter{+5.4}{+3.7}{+7.3} & \dbetter{+6.3}{+5.4}{+7.3} & \dbetter{+4.8}{+1.2}{+8.5} \\
\sansi{} $T{=}4$ (L4) $-$ Qwen3.5-2B & \dbetter{+4.1}{+3.4}{+4.8} & \dbetter{+2.1}{+1.0}{+3.3} & \dbetter{+2.1}{+0.5}{+3.9} & \dbetter{+5.6}{+4.6}{+6.6} & \dns{+1.7}{-2.5}{+5.6} \\
Qwen3.5-2B $-$ Ouro one loop & \dbetter{+8.1}{+7.2}{+9.0} & \dbetter{+7.0}{+5.5}{+8.6} & \dbetter{+2.5}{+0.5}{+4.4} & \dbetter{+10.4}{+9.0}{+11.6} & \dbetter{+7.8}{+3.0}{+12.3} \\
Qwen3.5-4B $-$ Qwen3.5-2B & \dbetter{+7.1}{+6.4}{+7.8} & \dbetter{+4.2}{+3.1}{+5.3} & \dbetter{+6.6}{+5.0}{+8.2} & \dbetter{+8.4}{+7.4}{+9.3} & \dbetter{+5.6}{+1.7}{+9.7} \\
\sansi{}: L2 $-$ L1 & \dbetter{+8.5}{+7.6}{+9.3} & \dbetter{+6.5}{+5.1}{+8.0} & \dbetter{+3.1}{+1.0}{+5.3} & \dbetter{+11.1}{+10.1}{+12.2} & \dns{+2.5}{-1.0}{+6.1} \\
\sansi{}: L3 $-$ L2 & \dbetter{+3.5}{+3.0}{+4.0} & \dbetter{+2.7}{+1.9}{+3.7} & \dbetter{+2.8}{+1.6}{+4.0} & \dbetter{+4.0}{+3.3}{+4.7} & \dbetter{+4.2}{+0.9}{+7.6} \\
\sansi{}: L4 $-$ L3 & \dbetter{+1.2}{+0.9}{+1.6} & \dbetter{+1.0}{+0.4}{+1.6} & \dns{+0.2}{-0.6}{+0.9} & \dbetter{+1.6}{+1.2}{+2.1} & \dns{+1.4}{-0.7}{+3.5} \\
\sansi{}: L8 $-$ L4 & \dns{+0.4}{-0.0}{+0.7} & \dns{+0.0}{-0.6}{+0.7} & \dns{+0.6}{-0.4}{+1.5} & \dns{+0.4}{-0.0}{+0.9} & \dns{-0.3}{-3.0}{+2.7} \\
Every loop $-$ last loop only (L8) & \dbetter{+1.1}{+0.7}{+1.5} & \dbetter{+2.0}{+1.1}{+2.8} & \dbetter{+1.3}{+0.1}{+2.4} & \dbetter{+0.7}{+0.2}{+1.3} & \dns{+0.3}{-2.5}{+3.2} \\
Every loop $-$ last loop only (L4) & \dbetter{+3.7}{+3.2}{+4.2} & \dbetter{+5.8}{+4.8}{+6.9} & \dbetter{+3.6}{+2.2}{+5.0} & \dbetter{+2.9}{+2.2}{+3.6} & \dbetter{+4.6}{+1.6}{+8.1} \\
Loops 1, 2, 4, 8 only $-$ every loop (L8) & \dworse{-0.7}{-1.1}{-0.4} & \dns{-0.4}{-1.1}{+0.3} & \dworse{-2.3}{-3.4}{-1.2} & \dns{-0.3}{-0.8}{+0.2} & \dns{-1.9}{-4.3}{+0.6} \\
Loops 1, 2, 4, 8 only $-$ every loop (L1) & \dbetter{+0.8}{+0.4}{+1.2} & \dbetter{+1.9}{+0.9}{+2.9} & \dns{+0.1}{-1.1}{+1.1} & \dbetter{+0.7}{+0.2}{+1.2} & \dworse{-2.5}{-4.8}{-0.1} \\
Reinforcement learning $-$ every loop (L8) & \dns{-0.3}{-0.7}{+0.1} & \dns{-0.5}{-1.2}{+0.3} & \dns{-0.9}{-1.8}{+0.1} & \dns{-0.1}{-0.6}{+0.5} & \dns{-0.0}{-2.3}{+2.3} \\
Cross-entropy only $-$ \sansi{} (L8) & \dns{-0.4}{-0.8}{-0.0} & \dns{-0.2}{-0.9}{+0.5} & \dns{-1.0}{-2.0}{+0.0} & \dns{-0.2}{-0.8}{+0.3} & \dns{-2.2}{-4.9}{+0.4} \\
$T{=}4$ $-$ $T{=}8$, both read at L4 & \dworse{-0.8}{-1.2}{-0.4} & \dns{-0.2}{-0.9}{+0.6} & \dworse{-2.7}{-3.7}{-1.7} & \dns{-0.3}{-0.8}{+0.2} & \dworse{-3.3}{-5.8}{-1.2} \\
$T{=}4$: L8 $-$ L4 & \dworse{-1.3}{-1.7}{-0.9} & \dworse{-1.5}{-2.3}{-0.8} & \dns{-0.9}{-2.0}{+0.3} & \dworse{-1.4}{-1.9}{-0.8} & \dns{+0.4}{-2.5}{+3.3} \\
Jev API $-$ \sansi{} L8 & \dbetter{+6.9}{+5.7}{+8.1} & \dworse{-9.1}{-10.9}{-7.3} & \dns{-2.0}{-5.4}{+1.5} & \dbetter{+15.3}{+14.1}{+16.5} & \dbetter{+15.6}{+11.0}{+20.6} \\
\sansi{}-2.6B L8 $-$ Qwen3.5-4B & \dbetter{+2.0}{+1.4}{+2.6} & \dns{-0.1}{-1.1}{+0.8} & \dbetter{+5.5}{+3.9}{+7.3} & \dbetter{+1.5}{+0.6}{+2.4} & \dbetter{+6.2}{+2.6}{+9.7} \\
\sansi{}-2.6B L8 $-$ \sansi{} L8 & \dbetter{+3.8}{+3.3}{+4.4} & \dbetter{+1.8}{+0.9}{+2.7} & \dbetter{+6.7}{+5.1}{+8.5} & \dbetter{+3.5}{+2.7}{+4.3} & \dbetter{+7.1}{+3.6}{+10.7} \\
\sansi{}-2.6B: L8 $-$ L1 & \dbetter{+13.4}{+12.6}{+14.3} & \dbetter{+9.5}{+7.9}{+11.2} & \dbetter{+12.6}{+10.5}{+14.9} & \dbetter{+14.9}{+13.8}{+16.0} & \dbetter{+18.8}{+13.4}{+24.4} \\
\sansi{}-2.6B: L8 $-$ L4 & \dns{+0.2}{-0.1}{+0.5} & \dns{-0.4}{-0.9}{+0.1} & \dbetter{+0.8}{+0.1}{+1.6} & \dns{+0.1}{-0.3}{+0.5} & \dbetter{+2.9}{+0.9}{+5.1} \\
\sansi{}-2.6B L3 $-$ Qwen3.5-4B & \dbetter{+0.9}{+0.3}{+1.5} & \dns{-0.6}{-1.7}{+0.4} & \dbetter{+3.6}{+2.1}{+5.2} & \dns{+0.6}{-0.3}{+1.4} & \dns{+1.4}{-1.9}{+4.6} \\
\sansi{}-2.6B L1 $-$ Qwen3.5-2B & \dworse{-4.3}{-5.1}{-3.6} & \dworse{-5.4}{-6.8}{-4.0} & \dns{-0.5}{-2.4}{+1.7} & \dworse{-5.1}{-6.1}{-4.0} & \dworse{-6.9}{-11.4}{-2.5} \\
Kev-4B (our data) $-$ Qwen3.5-4B & \dns{+0.5}{-0.0}{+1.0} & \dns{+0.1}{-0.7}{+0.9} & \dbetter{+1.9}{+0.8}{+3.1} & \dns{+0.2}{-0.6}{+1.0} & \dns{+0.4}{-2.3}{+3.2} \\
Kev-4B (our data) $-$ \sansi{} L8 & \dbetter{+2.4}{+1.7}{+3.1} & \dbetter{+2.0}{+1.0}{+3.1} & \dbetter{+3.2}{+1.6}{+4.8} & \dbetter{+2.3}{+1.3}{+3.3} & \dns{+1.3}{-2.6}{+4.9} \\
Kev-4B (our data) $-$ \sansi{} L3 & \dbetter{+3.9}{+3.2}{+4.6} & \dbetter{+3.0}{+1.9}{+4.2} & \dbetter{+3.9}{+2.2}{+5.5} & \dbetter{+4.3}{+3.4}{+5.3} & \dns{+2.5}{-1.3}{+6.5} \\
\sansi{}-2.6B L8 $-$ Kev-4B (our data) & \dbetter{+1.5}{+0.8}{+2.2} & \dns{-0.3}{-1.3}{+0.7} & \dbetter{+3.5}{+2.0}{+5.2} & \dbetter{+1.3}{+0.3}{+2.3} & \dbetter{+5.8}{+2.0}{+9.7} \\
\bottomrule
\end{tabular}
}
\caption{Accuracy differences in points with 95\% bootstrap intervals. L$t$: read at loop $t$. Colours: \textcolor{better}{\textbf{teal}}, the first term is significantly better (the interval excludes 0); \textcolor{worse}{\textbf{red}}, significantly worse; \textcolor{black!55}{grey}, the interval includes 0.}
\label{tab:ci}
\end{table*}

\subsection{The comparison with the model of the same shape}
\label{app:main-shape}

The difference of 13.5 points between \sansi{} and SmolLM2-1.7B is 12.9, 14.8 and 12.9 points in the three seeds (Table~\ref{tab:seeds}). It is far larger than the variation between seeds, which is at most 1.3 points for the overall accuracy of any model. By test group, the difference is 10.5 points in distribution, 10.0 on near transfer, 15.8 on far transfer and 14.0 on JevBench (Table~\ref{tab:ci}).

We use SmolLM2 as the main control because it was pre-trained without loops. A backbone that was pre-trained to loop might be at a disadvantage when it is run only once, and a comparison with it alone could then overstate the gain. This is not the case: Ouro-1.4B trained and run with one loop reaches 58.6\%, which cannot be distinguished from SmolLM2 (+0.1 points [$-$0.7, 1.0]), and \sansi{} at loop 8 is 13.4 points [12.4, 14.4] above this one-loop model.

The gain is also not an effect of a short training. Table~\ref{tab:steps} repeats the training of seed 0 with 2,000 instead of 1,000 steps. The longer training adds 1.9 points to SmolLM2-1.7B [1.1, 2.7] and 0.9 points to \sansi{} at loop 8, so that the difference between the two stays above 11 points. Longer training makes the calibration of both models worse.

\begin{table*}[t]
\centering
\small
\begin{minipage}[t]{0.48\textwidth}\vspace{0pt}
\centering
\resizebox{\linewidth}{!}{
\begin{tabular}{@{}l ccc c@{}}
\toprule
 & Seed 0 & Seed 1 & Seed 2 & Mean\sd{sd} \\
\midrule
\rowcolor{groupbg}\multicolumn{5}{@{}l}{\emph{Accuracy (\%), all test items}} \\
SmolLM2-1.7B & 58.4 & 57.8 & 59.1 & 58.4\sd{0.7} \\
Ouro-1.4B, one loop & 58.3 & 58.7 & 58.7 & 58.6\sd{0.2} \\
\sansi{} trained with 4 loops, loop 4 & 70.4 & 71.1 & 71.0 & 70.8\sd{0.4} \\
\sansi{}, loop 4 & 71.2 & 72.2 & 71.4 & 71.6\sd{0.5} \\
\sansi{}, loop 8 & 71.3 & 72.6 & 72.0 & 72.0\sd{0.7} \\
Loops 1, 2, 4, 8 only, loop 8 & 71.5 & 71.2 & 71.1 & 71.2\sd{0.2} \\
Last loop only, loop 8 & 70.5 & 71.0 & 71.2 & 70.9\sd{0.4} \\
Reinforcement learning, loop 8 & 71.6 & 71.7 & 71.7 & 71.7\sd{0.1} \\
Cross-entropy only, loop 8 & 71.8 & 71.7 & 71.1 & 71.6\sd{0.4} \\
Qwen3.5-2B & 66.6 & 67.4 & 66.1 & 66.7\sd{0.7} \\
Qwen3.5-4B & 74.3 & 73.2 & 74.1 & 73.8\sd{0.6} \\
Kev-4B (our data) & 74.2 & 74.7 & 74.1 & 74.3\sd{0.3} \\
\sansi{}-2.6B, loop 8 & 75.2 & 76.4 & 75.8 & 75.8\sd{0.6} \\
\midrule
\rowcolor{groupbg}\multicolumn{5}{@{}l}{\emph{Accuracy on near transfer (\%)}} \\
SmolLM2-1.7B & 57.5 & 57.2 & 58.4 & 57.7\sd{0.6} \\
\sansi{} trained with 4 loops, loop 4 & 61.7 & 65.5 & 66.2 & 64.5\sd{2.4} \\
\sansi{}, loop 8 & 64.8 & 70.2 & 68.1 & 67.7\sd{2.7} \\
Qwen3.5-4B & 71.2 & 66.4 & 69.2 & 68.9\sd{2.4} \\
\sansi{}-2.6B, loop 8 & 71.7 & 76.8 & 74.8 & 74.4\sd{2.5} \\
\midrule
\rowcolor{groupbg}\multicolumn{5}{@{}l}{\emph{Differences (points), all test items}} \\
\sansi{} loop 8 $-$ SmolLM2-1.7B & $+12.9$ & $+14.8$ & $+12.9$ & $+13.5$\sd{1.1} \\
\sansi{} (4 loops) $-$ SmolLM2-1.7B & $+12.0$ & $+13.3$ & $+11.8$ & $+12.4$\sd{0.8} \\
Qwen3.5-4B $-$ \sansi{} (4 loops) & $+3.8$ & $+2.0$ & $+3.1$ & $+3.0$\sd{0.9} \\
Qwen3.5-4B $-$ \sansi{} loop 8 & $+3.0$ & $+0.5$ & $+2.1$ & $+1.8$\sd{1.2} \\
\sansi{} (4 loops) $-$ \sansi{}, both at loop 4 & $-0.8$ & $-1.1$ & $-0.5$ & $-0.8$\sd{0.3} \\
\sansi{} loop 8 $-$ last loop only & $+0.8$ & $+1.7$ & $+0.7$ & $+1.1$\sd{0.5} \\
Loops 1, 2, 4, 8 only $-$ \sansi{}, loop 8 & $+0.1$ & $-1.5$ & $-0.9$ & $-0.7$\sd{0.8} \\
Reinforcement learning $-$ \sansi{}, loop 8 & $+0.3$ & $-0.9$ & $-0.3$ & $-0.3$\sd{0.6} \\
\sansi{} loop 8 $-$ Qwen3.5-2B & $+4.7$ & $+5.2$ & $+5.9$ & $+5.3$\sd{0.6} \\
\sansi{}-2.6B loop 8 $-$ Qwen3.5-4B & $+0.9$ & $+3.3$ & $+1.8$ & $+2.0$\sd{1.2} \\
\sansi{}-2.6B loop 8 $-$ \sansi{} loop 8 & $+3.9$ & $+3.8$ & $+3.9$ & $+3.8$\sd{0.1} \\
Kev-4B (our data) $-$ \sansi{} loop 8 & $+2.9$ & $+2.0$ & $+2.1$ & $+2.4$\sd{0.5} \\
\sansi{}-2.6B loop 8 $-$ Kev-4B (our data) & $+1.0$ & $+1.7$ & $+1.7$ & $+1.5$\sd{0.4} \\
\midrule
\rowcolor{groupbg}\multicolumn{5}{@{}l}{\emph{Differences (points), near transfer}} \\
Qwen3.5-4B $-$ \sansi{} (4 loops) & $+9.5$ & $+0.9$ & $+3.0$ & $+4.5$\sd{4.5} \\
Qwen3.5-4B $-$ \sansi{} loop 8 & $+6.4$ & $-3.8$ & $+1.1$ & $+1.2$\sd{5.1} \\
\sansi{} (4 loops) $-$ \sansi{}, both at loop 4 & $-3.3$ & $-3.6$ & $-1.2$ & $-2.7$\sd{1.3} \\
\sansi{}-2.6B loop 8 $-$ Qwen3.5-4B & $+0.5$ & $+10.4$ & $+5.6$ & $+5.5$\sd{4.9} \\
\bottomrule
\end{tabular}
}
\caption{Results seed by seed, with the mean and the standard deviation over the seeds. Overall accuracy is stable across seeds; accuracy on near transfer (1,871 items, of which 480 come from one source) is not, and neither are differences that rest on it.}
\label{tab:seeds}
\end{minipage}\hfill
\begin{minipage}[t]{0.48\textwidth}\vspace{0pt}
\centering
\resizebox{\linewidth}{!}{
\begin{tabular}{@{}ll ccc ccc@{}}
\toprule
 & & \multicolumn{3}{c}{Accuracy (\%) at loop} & \multicolumn{3}{c}{ECE at the last loop} \\
\cmidrule(lr){3-5}\cmidrule(lr){6-8}
Model & Steps & 1 & 4 & 8 & All & Near & Far \\
\midrule
\sansi{} & 1,000 & 57.7 & 71.2 & 71.3 & .099 & .183 & .089 \\
\sansi{} & 2,000 & 60.1 & 71.6 & 72.2 & .163 & .247 & .167 \\
SmolLM2-1.7B & 1,000 & 58.4 & -- & -- & .061 & .034 & .098 \\
SmolLM2-1.7B & 2,000 & 60.3 & -- & -- & .116 & .107 & .143 \\
\bottomrule
\end{tabular}
}
\caption{Training for 2,000 instead of 1,000 steps (seed 0; the learning-rate schedule is stretched accordingly). The 12,800 training items are passed once every 800 steps. Accuracy rises by one to two points for both models and the gap between them stays (12.9 points after 1,000 steps, 11.9 after 2,000), while the calibration error grows by more than half.}
\label{tab:steps}
\end{minipage}
\end{table*}

\subsection{Answerable items and the abstention threshold}
\label{app:main-abstain}

Accuracy combines two abilities: choosing the right option on the 9,645 answerable test items, and giving no hard answer on the 382 unanswerable ones (3.8\% of the test set), with a threshold set halfway between the uniform distribution and certainty (Appendix~\ref{app:metrics}). Table~\ref{tab:abstain} separates the two and moves the threshold to a quarter and to three quarters of that way. On the answerable items alone, \sansi{} is 13.3 points above SmolLM2-1.7B, 13.4 above Ouro-1.4B with one loop, 5.1 above Qwen3.5-2B and 1.9 below Qwen3.5-4B, against 13.5, 13.4, 5.3 and $-$1.8 on all items. The threshold moves the accuracy on all items by at most 0.7 points and changes no ranking. The hard-answer rates depend on it more, but at every threshold \sansi{} gives fewer hard answers than the single-pass models of its size and about as few as Qwen3.5-4B.

\begin{table*}[t]
\centering
\small
\setlength{\tabcolsep}{5pt}
\begin{tabular}{@{}l c ccc ccc@{}}
\toprule
 & Answerable & \multicolumn{3}{c}{Accuracy (\%) on all items} & \multicolumn{3}{c}{Hard answers (\%)} \\
\cmidrule(lr){2-2}\cmidrule(lr){3-5}\cmidrule(l){6-8}
Model & accuracy (\%) & $\alpha = \frac{1}{4}$ & $\alpha = \frac{1}{2}$ & $\alpha = \frac{3}{4}$ & $\alpha = \frac{1}{4}$ & $\alpha = \frac{1}{2}$ & $\alpha = \frac{3}{4}$ \\
\midrule
SmolLM2-1.7B & 58.2\sd{0.6} & 57.7\sd{0.7} & 58.4\sd{0.7} & 58.9\sd{0.7} & 56.5\sd{4.4} & 36.3\sd{4.2} & 23.2\sd{3.7} \\
Ouro-1.4B, one loop & 58.2\sd{0.3} & 58.1\sd{0.2} & 58.6\sd{0.2} & 58.9\sd{0.2} & 45.3\sd{3.4} & 31.9\sd{2.2} & 22.9\sd{1.2} \\
Qwen3.5-2B & 66.5\sd{0.7} & 66.3\sd{0.6} & 66.7\sd{0.7} & 67.0\sd{0.6} & 37.6\sd{2.1} & 27.9\sd{1.0} & 20.2\sd{2.7} \\
Qwen3.5-4B & 73.5\sd{0.7} & 73.5\sd{0.6} & 73.8\sd{0.6} & 74.1\sd{0.6} & 26.3\sd{3.9} & 18.2\sd{2.5} & 12.3\sd{1.6} \\
\rowcolor{sansibg}\sansi{} & 71.6\sd{0.7} & 71.6\sd{0.7} & 72.0\sd{0.7} & 72.2\sd{0.7} & 27.8\sd{0.8} & 17.5\sd{1.3} & 11.3\sd{2.2} \\
\rowcolor{sansibg}\sansi{}-2.6B & 75.6\sd{0.7} & 75.4\sd{0.6} & 75.8\sd{0.6} & 76.0\sd{0.7} & 28.4\sd{3.5} & 18.2\sd{2.4} & 13.7\sd{0.8} \\
\bottomrule
\end{tabular}

\caption{Accuracy on the 9,645 answerable test items, and accuracy on all 10,027 items and the rate of hard answers on the 382 unanswerable items with the threshold for a hard answer at a quarter, half (as in the paper) and three quarters of the way from the uniform distribution to certainty: a top probability of at least $1/K + \alpha(1 - 1/K)$, $\alpha \in \{\frac{1}{4}, \frac{1}{2}, \frac{3}{4}\}$. Mean $\pm$ standard deviation over three seeds; shaded rows: \sansi{}.}
\label{tab:abstain}
\end{table*}

\subsection{Every test source}
\label{app:main-sources}

Table~\ref{tab:sources} gives the accuracy on each of the 59 test sources. \sansi{} at loop 8 is more accurate than SmolLM2-1.7B on 55 of them. The four exceptions are two small in-distribution classification sources with 40 items each (\texttt{kev\_amazon\_id} and \texttt{kev\_imdb\_id}), the easy tier of JevBench, on which both models answer every item correctly, and ProofWriter with rules in natural language (\texttt{proofwriter\_natlang}, near transfer, 319 items), on which \sansi{} is 7.2 points below SmolLM2-1.7B (67.2\% against 74.4\%).

\begin{table*}[t]
\centering
\footnotesize
\setlength{\tabcolsep}{1.9pt}
\renewcommand{\arraystretch}{0.9}
\begin{tabular}{@{}lllr cc>{\columncolor{sansibg}}c cc@{}}
\toprule
Source & Type & Tier & $n$ & \mkSmol{}SmolLM2-1.7B & \mkOuro{}Ouro, one loop & \mkSansi{}\sansi{} & \mkQwen{}Qwen3.5-4B & \mkJev{}Jev API \\
\midrule
\texttt{kev\_agnews\_id} & A & ID & 41 & 88.6\sd{1.4} & 87.8\sd{4.9} & 94.3\sd{3.7} & 94.3\sd{1.4} & 90.2 \\
\texttt{kev\_amazon\_id} & A & ID & 40 & 62.5\sd{5.0} & 63.3\sd{5.8} & \textcolor{worse}{59.2}\sd{3.8} & 62.5\sd{8.7} & 55.0 \\
\texttt{kev\_banking77\_id} & A & ID & 40 & 81.7\sd{5.2} & 82.5\sd{0.0} & 84.2\sd{2.9} & 84.2\sd{3.8} & 92.5 \\
\texttt{kev\_dbpedia14\_id} & A & ID & 40 & 97.5\sd{2.5} & 95.8\sd{1.4} & 100.0\sd{0.0} & 98.3\sd{2.9} & 97.5 \\
\texttt{kev\_imdb\_id} & A & ID & 40 & 95.8\sd{2.9} & 93.3\sd{1.4} & \textcolor{worse}{87.5}\sd{2.5} & 88.3\sd{1.4} & 85.0 \\
\texttt{kev\_sst5\_id} & A & ID & 40 & 53.3\sd{7.6} & 49.2\sd{3.8} & 59.2\sd{5.8} & 50.8\sd{1.4} & 62.5 \\
\texttt{kev\_trec\_id} & A & ID & 40 & 80.0\sd{2.5} & 83.3\sd{2.9} & 90.0\sd{6.6} & 95.8\sd{1.4} & 85.0 \\
\texttt{kev\_yelp\_id} & A & ID & 40 & 73.3\sd{3.8} & 75.8\sd{2.9} & 78.3\sd{1.4} & 75.0\sd{2.5} & 70.0 \\
\texttt{emotion} & A & Far & 240 & 57.1\sd{1.5} & 59.7\sd{2.6} & 58.6\sd{0.2} & 60.4\sd{0.7} & 62.9 \\
\texttt{kev\_v9\_emotion} & A & Far & 64 & 56.2\sd{2.7} & 61.5\sd{2.4} & 56.8\sd{2.4} & 57.8\sd{2.7} & 59.4 \\
\texttt{kev\_v9\_tweet\_offensive} & A & Far & 64 & 71.9\sd{2.7} & 75.0\sd{2.7} & 77.1\sd{5.9} & 74.0\sd{2.4} & 79.7 \\
\texttt{tweet\_eval} & A & Far & 240 & 67.2\sd{1.3} & 65.6\sd{5.5} & 72.8\sd{0.9} & 77.2\sd{0.9} & 72.9 \\
\texttt{yahoo\_answers} & A & Far & 160 & 66.9\sd{1.1} & 63.1\sd{1.9} & 70.0\sd{1.1} & 72.7\sd{2.0} & 75.0 \\
\addlinespace[2pt]
\texttt{clutrr\_id} & B & ID & 160 & 30.2\sd{8.3} & 30.0\sd{10.0} & 63.7\sd{1.3} & 71.2\sd{3.8} & 50.0 \\
\texttt{kev\_compositional\_id} & B & ID & 120 & 56.9\sd{5.4} & 53.6\sd{4.6} & 78.1\sd{3.4} & 78.9\sd{3.5} & 81.7 \\
\texttt{kev\_legacy\_policy\_id} & B & ID & 120 & 86.9\sd{2.7} & 88.1\sd{3.2} & 93.1\sd{1.0} & 96.9\sd{2.1} & 92.5 \\
\texttt{musique\_id} & B & ID & 160 & 96.5\sd{1.0} & 97.9\sd{1.0} & 98.5\sd{0.4} & 98.8\sd{0.6} & 95.6 \\
\texttt{proofwriter\_id} & B & ID & 240 & 73.3\sd{0.8} & 72.5\sd{2.6} & 82.6\sd{1.3} & 87.1\sd{1.7} & 78.8 \\
\texttt{clutrr\_long} & B & Near & 480 & 32.2\sd{1.0} & 24.4\sd{4.3} & 46.3\sd{6.3} & 41.5\sd{6.4} & 38.8 \\
\texttt{musique\_4hop} & B & Near & 192 & 88.2\sd{2.6} & 92.2\sd{0.9} & 96.7\sd{0.6} & 97.9\sd{0.5} & 83.3 \\
\texttt{proofwriter\_natlang} & B & Near & 319 & 74.4\sd{0.7} & 71.3\sd{2.8} & \textcolor{worse}{67.2}\sd{10.6} & 69.2\sd{6.7} & 78.4 \\
\texttt{bbh} & B & Far & 480 & 22.5\sd{2.3} & 28.4\sd{0.8} & 52.4\sd{1.6} & 54.6\sd{0.9} & 96.5 \\
\texttt{folio} & B & Far & 320 & 45.2\sd{1.0} & 46.4\sd{1.6} & 61.0\sd{1.8} & 56.0\sd{2.1} & 75.6 \\
\texttt{hover} & B & Far & 360 & 54.1\sd{3.9} & 66.9\sd{1.8} & 70.3\sd{0.0} & 69.1\sd{2.4} & 68.9 \\
\texttt{kev\_v9\_composition\_holdout} & B & Far & 57 & 53.2\sd{4.1} & 55.6\sd{2.7} & 84.2\sd{7.6} & 77.8\sd{11.7} & 91.2 \\
\texttt{kev\_v9\_legacy\_holdout} & B & Far & 64 & 47.9\sd{0.9} & 45.3\sd{4.1} & 79.7\sd{4.7} & 78.6\sd{3.3} & 96.9 \\
\addlinespace[2pt]
\texttt{chaosnli\_id} & C & ID & 80 & 65.8\sd{4.7} & 60.0\sd{2.5} & 72.9\sd{1.4} & 76.7\sd{4.0} & 80.0 \\
\texttt{musique\_pairs\_id} & C & ID & 158 & 83.1\sd{2.0} & 85.9\sd{1.5} & 93.9\sd{0.4} & 94.5\sd{1.3} & 62.7 \\
\texttt{squad2\_id} & C & ID & 240 & 83.2\sd{3.1} & 85.7\sd{0.9} & 93.2\sd{2.1} & 93.6\sd{1.0} & 56.7 \\
\texttt{chaosnli\_mnli} & C & Near & 400 & 46.5\sd{2.5} & 55.6\sd{3.9} & 58.2\sd{0.6} & 65.6\sd{3.5} & 67.0 \\
\texttt{musique\_pairs\_4hop} & C & Near & 240 & 70.0\sd{2.2} & 77.2\sd{1.0} & 86.1\sd{1.0} & 87.9\sd{1.4} & 65.4 \\
\texttt{kev\_v9\_unknowable\_pairs} & C & Far & 148 & 62.8\sd{4.7} & 65.1\sd{1.7} & 83.3\sd{5.1} & 77.9\sd{6.9} & 83.8 \\
\texttt{sys1cal} & C & Far & 292 & 51.3\sd{6.7} & 45.9\sd{2.2} & 52.3\sd{2.9} & 60.2\sd{1.2} & 69.2 \\
\addlinespace[2pt]
\texttt{hotpotqa\_lengths} & D & ID & 120 & 62.5\sd{2.2} & 71.9\sd{2.1} & 87.8\sd{1.7} & 88.3\sd{0.0} & 91.7 \\
\texttt{musique\_lengths} & D & ID & 237 & 96.2\sd{1.1} & 95.6\sd{0.5} & 96.5\sd{0.6} & 97.7\sd{1.7} & 82.7 \\
\texttt{contract\_nli} & D & Far & 240 & 49.0\sd{4.6} & 53.3\sd{3.3} & 76.0\sd{2.4} & 74.4\sd{0.2} & 77.5 \\
\texttt{kev\_v9\_buried} & D & Far & 64 & 52.6\sd{2.4} & 55.2\sd{5.9} & 63.5\sd{2.4} & 72.9\sd{4.5} & 76.6 \\
\texttt{quality} & D & Far & 240 & 51.3\sd{0.8} & 42.5\sd{2.2} & 71.1\sd{2.4} & 72.6\sd{1.0} & 90.4 \\
\addlinespace[2pt]
\texttt{kev\_boolq\_id} & E & ID & 80 & 77.5\sd{1.3} & 76.7\sd{1.9} & 85.0\sd{3.3} & 88.3\sd{1.4} & 90.0 \\
\texttt{kev\_mnli\_id} & E & ID & 80 & 70.0\sd{2.5} & 86.3\sd{2.5} & 92.9\sd{1.9} & 95.0\sd{1.3} & 93.8 \\
\texttt{mnli\_mismatched} & E & Near & 240 & 68.5\sd{3.6} & 79.4\sd{1.3} & 85.6\sd{0.6} & 86.9\sd{2.4} & 87.1 \\
\texttt{anli\_dev} & E & Far & 240 & 38.1\sd{1.3} & 36.8\sd{2.1} & 56.9\sd{0.9} & 52.5\sd{3.8} & 70.4 \\
\texttt{kev\_v9\_paws} & E & Far & 64 & 55.2\sd{3.6} & 39.1\sd{1.6} & 68.2\sd{3.3} & 77.1\sd{1.8} & 78.1 \\
\texttt{kev\_v9\_qnli} & E & Far & 64 & 78.1\sd{0.0} & 81.8\sd{1.8} & 88.5\sd{2.4} & 88.0\sd{3.9} & 93.8 \\
\texttt{paws} & E & Far & 200 & 60.2\sd{2.3} & 55.5\sd{1.5} & 80.2\sd{1.8} & 82.5\sd{1.7} & 85.0 \\
\texttt{qnli} & E & Far & 200 & 83.5\sd{0.9} & 82.8\sd{2.6} & 87.8\sd{3.0} & 87.7\sd{2.8} & 93.0 \\
\texttt{wanli} & E & Far & 200 & 53.8\sd{1.6} & 58.2\sd{2.5} & 63.7\sd{3.3} & 67.2\sd{0.3} & 73.0 \\
\addlinespace[2pt]
\texttt{arc\_challenge} & F & Far & 320 & 60.2\sd{2.1} & 52.6\sd{1.0} & 86.1\sd{0.8} & 90.0\sd{0.8} & 97.2 \\
\texttt{arc\_easy} & F & Far & 120 & 77.2\sd{1.3} & 70.0\sd{3.8} & 91.1\sd{2.1} & 95.8\sd{0.8} & 100.0 \\
\texttt{commonsense\_qa} & F & Far & 240 & 58.9\sd{1.3} & 53.1\sd{0.9} & 77.5\sd{1.9} & 82.5\sd{0.8} & 90.8 \\
\texttt{kev\_v9\_mmlu} & F & Far & 64 & 45.3\sd{4.1} & 31.2\sd{0.0} & 57.3\sd{3.3} & 75.0\sd{1.6} & 93.8 \\
\texttt{kev\_v9\_mmlu\_pro} & F & Far & 160 & 18.5\sd{2.6} & 13.3\sd{1.3} & 39.4\sd{1.1} & 44.8\sd{1.0} & 86.2 \\
\texttt{kev\_v9\_sciq} & F & Far & 64 & 89.1\sd{3.1} & 93.2\sd{0.9} & 95.8\sd{0.9} & 96.9\sd{0.0} & 98.4 \\
\texttt{mmlu} & F & Far & 480 & 51.0\sd{1.9} & 43.5\sd{1.6} & 67.2\sd{2.4} & 73.1\sd{0.6} & 92.5 \\
\texttt{mmlu\_pro} & F & Far & 240 & 20.7\sd{1.6} & 18.9\sd{1.0} & 39.7\sd{2.8} & 44.0\sd{1.6} & 85.4 \\
\texttt{sciq} & F & Far & 120 & 96.1\sd{1.7} & 97.8\sd{0.5} & 99.7\sd{0.5} & 99.7\sd{0.5} & 100.0 \\
\addlinespace[2pt]
\texttt{jevbench\_easy} & JB & JB & 48 & 100.0\sd{0.0} & 99.3\sd{1.2} & 100.0\sd{0.0} & 100.0\sd{0.0} & 100.0 \\
\texttt{jevbench\_hard} & JB & JB & 111 & 34.5\sd{1.0} & 38.4\sd{3.6} & 49.2\sd{2.6} & 47.4\sd{4.3} & 75.7 \\
\texttt{jevbench\_standard} & JB & JB & 72 & 67.1\sd{0.8} & 66.2\sd{4.9} & 89.4\sd{1.6} & 94.9\sd{0.8} & 98.6 \\
\bottomrule
\end{tabular}

\caption{Accuracy (\%) on each of the 59 test sources: mean $\pm$ standard deviation over three seeds (the Jev API is called once). Type: A--F as in Table~\ref{tab:data}; Tier: in-distribution (ID), near transfer, far transfer, JevBench (JB). \sansi{} is read at loop 8. Shaded column: \sansi{}; \textcolor{worse}{red}: below SmolLM2-1.7B.}
\label{tab:sources}
\end{table*}

\subsection{Item types}
\label{app:main-types}

Figure~\ref{fig:types-gain} and Table~\ref{tab:types} split the test set by item type, and \S\ref{sec:rq1} describes the gain over SmolLM2-1.7B for five of the six types. The sixth, items with uncertain evidence, gain 10.6 points. On knowledge questions, the first loop of \sansi{} is right on 47.1\% of the items and the eighth on 70.2\%.

The gap that remains to Qwen3.5-4B is not spread evenly over the item types. At loop 8, \sansi{} is level with Qwen3.5-4B on multi-step reasoning (0.0 points [$-$1.1, 1.1]) and cannot be distinguished from it on classification, long documents and sentence pairs (1.0 to 1.5 points behind, with intervals that include zero). It is behind on knowledge questions (4.9 points [3.4, 6.5]) and on items with uncertain evidence (3.5 points [1.7, 5.4]).

\begin{table*}[t]
\centering
\small
\setlength{\tabcolsep}{4.5pt}
\begin{tabular}{@{}lr c>{\columncolor{sansibg}}c>{\columncolor{sansibg}}c c c c@{}}
\toprule
 & & & \multicolumn{2}{c}{\sansi{}} & & \mkKev{}Kev-4B & $\Delta$ (loop 8 \\
\cmidrule(lr){4-5}
Item type & $n$ & \mkSmol{}SmolLM2-1.7B & loop 1 & loop 8 & \mkQwen{}Qwen3.5-4B & (our data) & $-$ SmolLM2) \\
\midrule
A\; Classification & 1,089 & 68.1\sd{0.7} & 65.4\sd{0.6} & 71.2\sd{0.2} & 72.7\sd{0.8} & 74.0\sd{0.7} & \textcolor{better}{$+3.1$}\sd{0.5} \\
B\; Multi-step reasoning & 3,072 & 52.8\sd{1.0} & 54.6\sd{0.8} & 67.9\sd{1.9} & 67.9\sd{0.5} & 69.5\sd{0.3} & \textcolor{better}{$+15.1$}\sd{2.2} \\
C\; Uncertain evidence & 1,558 & 62.9\sd{2.0} & 67.1\sd{0.3} & 73.5\sd{0.9} & 77.0\sd{0.8} & 75.6\sd{1.2} & \textcolor{better}{$+10.6$}\sd{1.8} \\
D\; Long documents & 901 & 64.1\sd{1.3} & 62.3\sd{1.6} & 80.8\sd{1.5} & 81.8\sd{0.9} & 82.1\sd{0.2} & \textcolor{better}{$+16.7$}\sd{2.6} \\
E\; Sentence pairs & 1,368 & 62.4\sd{0.7} & 63.0\sd{0.7} & 76.6\sd{0.8} & 77.6\sd{1.4} & 78.9\sd{1.0} & \textcolor{better}{$+14.2$}\sd{0.1} \\
F\; Knowledge & 1,808 & 52.7\sd{0.9} & 47.1\sd{1.2} & 70.2\sd{1.6} & 75.1\sd{0.4} & 74.5\sd{0.4} & \textcolor{better}{$+17.6$}\sd{2.2} \\
JevBench & 231 & 58.3\sd{0.7} & 64.5\sd{0.7} & 72.3\sd{1.6} & 73.2\sd{1.9} & 73.6\sd{0.4} & \textcolor{better}{$+14.0$}\sd{1.7} \\
\midrule
All & 10,027 & 58.4\sd{0.7} & 58.4\sd{0.7} & 72.0\sd{0.7} & 73.8\sd{0.6} & 74.3\sd{0.3} & \textcolor{better}{$+13.5$}\sd{1.1} \\
\bottomrule
\end{tabular}

\caption{Accuracy (\%) by item type, all groups pooled: mean $\pm$ standard deviation over three seeds. $\Delta$: \sansi{} at loop 8 minus SmolLM2-1.7B, paired by seed. Shaded columns: \sansi{}.}
\label{tab:types}
\end{table*}

\subsection{The Qwen3.5 models}
\label{app:main-qwen}

The Qwen3.5 models are stronger backbones than Ouro-1.4B already before fine-tuning (Table~\ref{tab:main-full}): Qwen3.5-4B is 13.7 points above Ouro-1.4B at its best loop (61.6\% against 47.9\%). After fine-tuning, Qwen3.5-4B reaches 73.8\%, 15.4 points above SmolLM2-1.7B [14.4, 16.4]. \sansi{} at loop 8 is 1.8 points behind it [1.2, 2.5]; in the three seeds the difference is 3.0, 0.5 and 2.1 points. Read after three loops, \sansi{} is 3.4 points behind [2.8, 4.1].

Qwen3.5-2B, a newer backbone of \sansi{}'s size, reaches 66.7\%. It is 8.1 points above Ouro-1.4B run once [7.2, 9.0], so the newer backbone is clearly the better one. \sansi{} draws level with it at the second loop (+0.2 points [$-$0.5, 0.8]) and is 5.3 points above it at loop 8 [4.5, 6.0]. This holds in every seed (4.7, 5.2 and 5.9 points) and in every test group (2.3 points in distribution, 5.4 on near transfer and 6.3 on far transfer). Table~\ref{tab:ci} gives all of these differences.

\subsection{Kev's recipe on our data}
\label{app:main-kev}

Our single-pass models are trained with our own recipe. To test whether the comparison depends on it, we train the backbone of Qwen3.5-4B on our 12,800 items with the code and the recipe of Kev (Appendix~\ref{app:training}); we call this model Kev-4B (our data). It reaches 74.3\% (74.2\%, 74.7\% and 74.1\% in the three seeds), which is 0.5 points above Qwen3.5-4B with our recipe, with an interval that includes zero ([0.0, 1.0]; Table~\ref{tab:ci}). The two recipes thus give the same accuracy on this backbone, and the single-pass reference of the main text is not weakened by our recipe. \sansi{} is 2.4 points behind Kev-4B (our data) at loop 8 [1.7, 3.1] and 3.9 points behind at loop 3 [3.2, 4.6], where it has used 1.4 times the GPU time of that model. \sansi{}-2.6B is 1.5 points ahead of it [0.8, 2.2]. As it is trained, Kev-4B (our data) has an ECE of 0.119, close to that of Qwen3.5-4B with our recipe (0.113). With the temperature that Kev's recipe fits on development data its accuracy is 74.5\% and its ECE 0.029; Appendix~\ref{app:probs-cal} shows that the same step gives the other models about the same ECE. By item type, Kev-4B (our data) stays within 1.6 points of Qwen3.5-4B with our recipe (Table~\ref{tab:types}).

\subsection{Released models: Kev-4B and the Jev API}
\label{app:main-api}

Two released models are natural references: Kev-4B \citep{palmer2026kev}, an open model on the backbone of Qwen3.5-4B, and the commercial Jev API. Neither was trained on our data, and a comparison on all test items would not be fair in either direction. Our in-distribution test items of the Kev sources (721 items) are cut from the file on which Kev was trained, so they are training items for the released Kev-4B and held-out items for our models. The other in-distribution and near-transfer items (3,266) come from sources on which only our models are trained. The comparison that favours neither side is on the 6,040 items of far transfer and JevBench, whose sources neither our models nor the released Kev-4B were trained on. Of these, 877 are development items of Kev, on which it selects its checkpoints; the last column of Table~\ref{tab:kev-who} leaves them out.

\begin{table*}[t]
\centering
\small
\setlength{\tabcolsep}{5pt}
\begin{tabular}{@{}ll ccccc@{}}
\toprule
 & & & \multicolumn{4}{c}{By who trained on the source of the item} \\
\cmidrule(l){4-7}
 & & All & Kev's training & Only our & Neither & Neither, without \\
Model & Params & (10,027) & items (721) & models (3,266) & (6,040) & Kev's dev.\ items (5,163) \\
\midrule
\rowcolor{groupbg}\multicolumn{7}{@{}l}{\emph{Released models, trained on other data}} \\
\mkJev{}Jev API (jev-1.13.0) & n/a & 78.9 & 84.9 & 69.1 & 83.5 & 83.2 \\
\mkKevR{}Kev-4B (released) & 4.2B & 68.2 & 96.1 & 54.9 & 72.0 & 71.0 \\
\midrule
\rowcolor{groupbg}\multicolumn{7}{@{}l}{\emph{Trained on our data}} \\
\mkKev{}Kev-4B (our data) & 4.2B & 74.3\sd{0.3} & 86.3\sd{0.6} & 79.0\sd{0.4} & 70.4\sd{0.6} & 70.0\sd{0.5} \\
\mkQwen{}Qwen3.5-4B & 4.2B & 73.8\sd{0.6} & 85.8\sd{0.4} & 77.9\sd{1.6} & 70.2\sd{0.2} & 70.0\sd{0.4} \\
\mkQwenS{}Qwen3.5-2B & 1.9B & 66.7\sd{0.7} & 80.4\sd{1.8} & 72.6\sd{1.8} & 61.9\sd{0.4} & 62.1\sd{0.3} \\
\mkSmol{}SmolLM2-1.7B & 1.7B & 58.4\sd{0.7} & 75.5\sd{0.7} & 65.7\sd{1.0} & 52.5\sd{0.7} & 52.3\sd{0.7} \\
\rowcolor{sansibg}\mkSansiL{}\sansi{}-2.6B, loop 8 & 2.7B & 75.8\sd{0.6} & 87.1\sd{1.0} & 80.7\sd{1.9} & 71.9\sd{0.3} & 71.3\sd{0.2} \\
\rowcolor{sansibg}\mkSansi{}\sansi{}, loop 8 & 1.4B & 72.0\sd{0.7} & 84.6\sd{1.4} & 76.2\sd{1.7} & 68.2\sd{0.1} & 68.0\sd{0.2} \\
\bottomrule
\end{tabular}

\caption{Accuracy (\%) by who was trained on the source of an item: mean $\pm$ standard deviation over three seeds (the released Kev-4B and the Jev API are single runs). Kev-4B (our data) is trained on our items with Kev's recipe and read at temperature 1; the released Kev-4B is read at the temperature it ships with.}
\label{tab:kev-who}
\end{table*}

On the 6,040 items that neither side was trained on, the released Kev-4B reaches 72.0\%. \sansi{}-2.6B is level with it (71.9\%; $-$0.1 points [$-$1.1, 0.8]) with 63\% of its parameters, and \sansi{} is 3.8 points behind it [2.9, 4.8]. Kev-4B trained on our 12,800 items reaches 70.4\% on these items, 1.6 points below the released model [0.7, 2.5]. The Jev API is far ahead of all these models (83.5\%; 11.5 points above the released Kev-4B [10.4, 12.7]).

The Jev API (jev-1.13.0) is more accurate than all models we trained (78.9\% overall; Table~\ref{tab:main-full}). Its lead comes from far transfer and JevBench (83.3\% and 87.9\%), in particular from knowledge questions. On in-distribution items \sansi{} is 9.1 points higher. We treat the API as a reference only: its size and its training data are not public, and it was not trained to give a uniform distribution on unanswerable items, to 62.8\% of which it gives a hard answer.

Table~\ref{tab:jevbench} splits the public JevBench items into their three tiers. All models answer nearly every item of the easy tier correctly (99.3\% to 100\%). On the standard tier, \sansi{} rises from 76.9\% at loop 1 to 93.1\% at loop 4 and ends at 89.4\% at loop 8, 5.5 points below Qwen3.5-4B (94.9\%). On the hard tier, \sansi{} and the two models on the Qwen3.5-4B backbone stay near 50\% (49.2\%, 47.4\% and 50.5\%), \sansi{}-2.6B reaches 60.1\%, and the API 75.7\%.

\begin{table*}[t]
\centering
\small
\setlength{\tabcolsep}{4pt}
\begin{tabular}{@{}l ccc@{}}
\toprule
Model & Easy (48) & Standard (72) & Hard (111) \\
\midrule
\mkSmol{}SmolLM2-1.7B & 100.0\sd{0.0} & 67.1\sd{0.8} & 34.5\sd{1.0} \\
\mkOuro{}Ouro-1.4B, one loop & 99.3\sd{1.2} & 66.2\sd{4.9} & 38.4\sd{3.6} \\
\rowcolor{sansibg}\mkSansi{}\sansi{}, loop 1 & 100.0\sd{0.0} & 76.9\sd{4.2} & 41.1\sd{1.9} \\
\rowcolor{sansibg}\mkSansi{}\sansi{}, loop 4 & 100.0\sd{0.0} & 93.1\sd{2.4} & 47.4\sd{2.3} \\
\rowcolor{sansibg}\mkSansi{}\sansi{}, loop 8 & 100.0\sd{0.0} & 89.4\sd{1.6} & 49.2\sd{2.6} \\
\rowcolor{sansibg}\mkSansiL{}\sansi{}-2.6B, loop 8 & 100.0\sd{0.0} & 95.4\sd{0.8} & 60.1\sd{1.4} \\
\mkQwen{}Qwen3.5-4B & 100.0\sd{0.0} & 94.9\sd{0.8} & 47.4\sd{4.3} \\
\mkKev{}Kev-4B (our data) & 99.3\sd{1.2} & 92.1\sd{0.8} & 50.5\sd{0.9} \\
\mkJev{}Jev API & 100.0 & 98.6 & 75.7 \\
\bottomrule
\end{tabular}

\caption{Accuracy (\%) on the public JevBench items by tier: mean $\pm$ standard deviation over three seeds (the Jev API is a single run).}
\label{tab:jevbench}
\end{table*}

\section{Loop by Loop: Additional Results}
\label{app:loops}

This appendix supports \S\ref{sec:rq2}. It gives every measure of \sansi{} after every loop (Appendix~\ref{app:loops-every}), the changes of the answers between consecutive loops (Appendix~\ref{app:loops-flips}), the changes seen from the first and from the last loop (Appendix~\ref{app:loops-dynamics}), the loop at which answers settle (Appendix~\ref{app:loops-settle}), the confidence of the items by whether their answer changed (Appendix~\ref{app:loops-groups}), and rules that stop early (Appendix~\ref{app:stop}).

\subsection{Every loop}
\label{app:loops-every}

Table~\ref{tab:perloop} gives all measures of \sansi{} after every loop: the accuracy on all test items and in every test group, the ECE by group, the mean confidence of right and of wrong answers, the AUROC with which confidence separates right from wrong answers, the evidence AUROC, the hard-answer rate, and the share of answers that differ from the previous loop.

\begin{table*}[t]
\centering
\resizebox{\textwidth}{!}{
\begin{tabular}{@{}c ccccc ccc cc ccc c@{}}
\toprule
 & \multicolumn{5}{c}{Accuracy (\%)} & \multicolumn{3}{c}{ECE} & \multicolumn{2}{c}{Confidence} & AUROC & Evid. & Hard & Answers \\
\cmidrule(lr){2-6}\cmidrule(lr){7-9}\cmidrule(lr){10-11}
Loop & All & In-dist. & Near & Far & JevB. & In-dist. & Near & Far & right & wrong & r/w & AUROC & (\%) & changed (\%) \\
\midrule
1 & 58.4 & 76.3 & 61.1 & 50.8 & 64.5 & .020 & .053 & .156 & .769 & .567 & .760 & .835 & 27.1 & -- \\
 & \sd{0.7} & \sd{0.9} & \sd{1.3} & \sd{0.5} & \sd{0.7} & \sd{.006} & \sd{.008} & \sd{.019} & \sd{.020} & \sd{.026} & \sd{.005} & \sd{.011} & \sd{2.1} &  \\[2pt]
2 & 66.9 & 82.8 & 64.2 & 61.9 & 67.0 & .044 & .110 & .095 & .822 & .619 & .777 & .902 & 19.1 & 29.2 \\
 & \sd{0.6} & \sd{0.4} & \sd{1.2} & \sd{0.8} & \sd{1.5} & \sd{.005} & \sd{.013} & \sd{.016} & \sd{.014} & \sd{.015} & \sd{.008} & \sd{.002} & \sd{1.4} & \sd{0.3} \\[2pt]
3 & 70.4 & 85.6 & 67.0 & 65.9 & 71.1 & .047 & .119 & .086 & .847 & .634 & .788 & .921 & 18.8 & 14.9 \\
 & \sd{0.6} & \sd{0.6} & \sd{1.6} & \sd{0.9} & \sd{2.0} & \sd{.006} & \sd{.016} & \sd{.016} & \sd{.015} & \sd{.018} & \sd{.010} & \sd{.006} & \sd{1.1} & \sd{0.4} \\[2pt]
4 & 71.6 & 86.5 & 67.2 & 67.6 & 72.6 & .046 & .143 & .084 & .857 & .647 & .788 & .928 & 18.1 & 7.8 \\
 & \sd{0.5} & \sd{0.5} & \sd{2.0} & \sd{0.5} & \sd{1.2} & \sd{.007} & \sd{.018} & \sd{.017} & \sd{.015} & \sd{.019} & \sd{.009} & \sd{.003} & \sd{0.5} & \sd{0.4} \\[2pt]
5 & 71.9 & 86.6 & 67.2 & 68.0 & 72.0 & .044 & .149 & .084 & .863 & .651 & .792 & .933 & 17.1 & 4.5 \\
 & \sd{0.7} & \sd{0.7} & \sd{2.4} & \sd{0.2} & \sd{1.1} & \sd{.006} & \sd{.024} & \sd{.017} & \sd{.015} & \sd{.018} & \sd{.009} & \sd{.003} & \sd{1.5} & \sd{0.3} \\[2pt]
6 & 72.1 & 86.8 & 67.8 & 68.2 & 72.2 & .047 & .147 & .087 & .866 & .656 & .793 & .934 & 17.0 & 3.3 \\
 & \sd{0.7} & \sd{0.7} & \sd{2.4} & \sd{0.1} & \sd{1.6} & \sd{.004} & \sd{.026} & \sd{.018} & \sd{.015} & \sd{.019} & \sd{.008} & \sd{.002} & \sd{2.7} & \sd{0.3} \\[2pt]
7 & 72.1 & 86.9 & 67.8 & 68.2 & 72.2 & .049 & .148 & .089 & .867 & .659 & .793 & .935 & 16.8 & 2.6 \\
 & \sd{0.7} & \sd{0.6} & \sd{2.5} & \sd{0.2} & \sd{1.2} & \sd{.004} & \sd{.027} & \sd{.020} & \sd{.014} & \sd{.021} & \sd{.006} & \sd{.002} & \sd{2.0} & \sd{0.3} \\[2pt]
8 & 72.0 & 86.6 & 67.7 & 68.0 & 72.3 & .048 & .149 & .092 & .869 & .660 & .795 & .935 & 17.5 & 2.3 \\
 & \sd{0.7} & \sd{0.6} & \sd{2.7} & \sd{0.1} & \sd{1.6} & \sd{.004} & \sd{.031} & \sd{.019} & \sd{.013} & \sd{.020} & \sd{.005} & \sd{.004} & \sd{1.3} & \sd{0.2} \\[2pt]
\bottomrule
\end{tabular}
}
\caption{\sansi{} after every loop (10,027 test items; means of three seeds, with the standard deviation over the seeds on the line below). Confidence right/wrong: mean top probability on items answered correctly/wrongly at that loop. AUROC r/w: right against wrong by confidence. Answers changed: share of answerable items whose answer differs from the previous loop.}
\label{tab:perloop}
\end{table*}

\begin{figure*}[t]
\centering
\includegraphics[width=\textwidth]{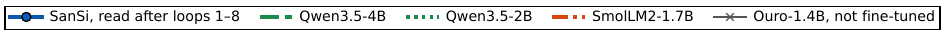}\\[1pt]
\begin{subfigure}[t]{0.2476\textwidth}\includegraphics[width=\linewidth]{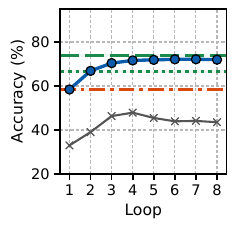}\caption{All test items}\end{subfigure}\hfill
\begin{subfigure}[t]{0.2476\textwidth}\includegraphics[width=\linewidth]{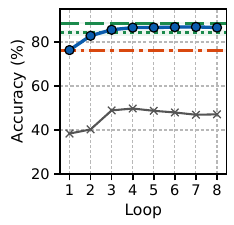}\caption{In-distribution}\end{subfigure}\hfill
\begin{subfigure}[t]{0.2476\textwidth}\includegraphics[width=\linewidth]{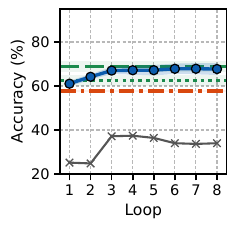}\caption{Near transfer}\end{subfigure}\hfill
\begin{subfigure}[t]{0.2476\textwidth}\includegraphics[width=\linewidth]{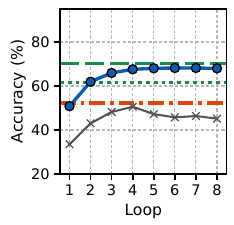}\caption{Far transfer}\end{subfigure}
\caption{Accuracy of \sansi{} after every loop, by distance from the training data (10,027, 2,116, 1,871 and 5,809 items; line: mean of three seeds; band: one standard deviation). Horizontal lines: the single-pass models. Grey: Ouro-1.4B before fine-tuning.}
\label{fig:tiers}
\end{figure*}

Figure~\ref{fig:tiers} shows the accuracy after every loop for every test group, with the single-pass models as horizontal lines and the backbone before fine-tuning in grey. The untuned backbone peaks at the four loops of its pre-training and then declines (47.9\% at loop 4, 43.5\% at loop 8). After fine-tuning, the curve of every group rises steeply up to loop 3 and is flat from loop 4. The gain from loop 1 to loop 8 is largest on far transfer (50.8\% to 68.0\%) and smallest on near transfer (61.1\% to 67.7\%); in distribution the accuracy rises from 76.3\% to 86.6\%.

\begin{figure*}[t]
\centering
\includegraphics[width=\textwidth]{figures/fig_perloop_legend.pdf}\\[1pt]
\begin{subfigure}[t]{0.3333\textwidth}\includegraphics[width=\linewidth]{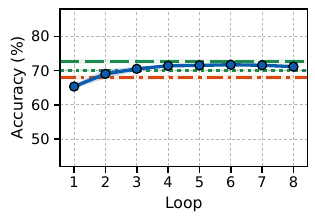}\caption{Classification}\end{subfigure}%
\begin{subfigure}[t]{0.3333\textwidth}\includegraphics[width=\linewidth]{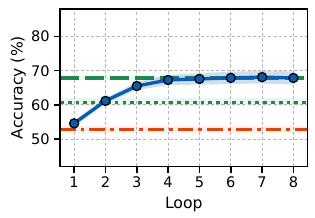}\caption{Multi-step reasoning}\end{subfigure}%
\begin{subfigure}[t]{0.3333\textwidth}\includegraphics[width=\linewidth]{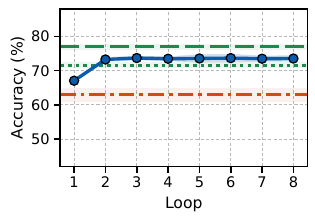}\caption{Uncertain evidence}\end{subfigure}\\[4pt]
\begin{subfigure}[t]{0.3333\textwidth}\includegraphics[width=\linewidth]{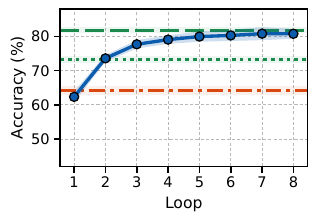}\caption{Long documents}\end{subfigure}%
\begin{subfigure}[t]{0.3333\textwidth}\includegraphics[width=\linewidth]{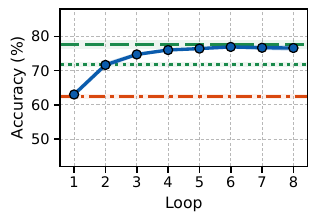}\caption{Sentence pairs}\end{subfigure}%
\begin{subfigure}[t]{0.3333\textwidth}\includegraphics[width=\linewidth]{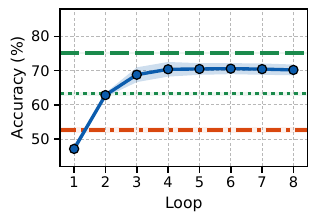}\caption{Knowledge}\end{subfigure}
\caption{Accuracy of \sansi{} after every loop for each item type, over all test groups (1,089, 3,072, 1,558, 901, 1,368 and 1,808 items; line: mean of three seeds; band: one standard deviation). Horizontal lines: the single-pass models.}
\label{fig:types}
\end{figure*}

Figure~\ref{fig:types} shows the same curves for every item type. Classification is nearly flat, and the other types gain most of their accuracy in loops 2 and 3.

\subsection{Changes between consecutive loops}
\label{app:loops-flips}

Table~\ref{tab:flips} gives, for every pair of consecutive loops, the share of answerable items whose answer changes, and the shares that go from wrong to right and from right to wrong. For \sansi{}, the share of changed answers roughly halves from one step to the next at first: 29.2\% between loops 1 and 2, 14.9\% between loops 2 and 3, and 7.8\% between loops 3 and 4. It falls to 2.3\% between loops 7 and 8. Up to loop 5, every step fixes more answers than it breaks; in the last two steps the two are about equal. The right half of the table gives the model that was trained with the loss on the last loop only (\S\ref{sec:ablations}). Its answers change much more between the early loops (58.4\% between loops 1 and 2), because its early loops were not trained to answer.

\begin{table*}[t]
\centering
\small
\setlength{\tabcolsep}{5pt}
\begin{tabular}{@{}c ccc ccc@{}}
\toprule
 & \multicolumn{3}{c}{Every loop trained (\sansi{})} & \multicolumn{3}{c}{Last loop only} \\
\cmidrule(lr){2-4}\cmidrule(lr){5-7}
Loops & changed & \textcolor{better}{wrong$\to$right} & \textcolor{worse}{right$\to$wrong} & changed & \textcolor{better}{wrong$\to$right} & \textcolor{worse}{right$\to$wrong} \\
\midrule
1 $\to$ 2 & 29.2\sd{0.3} & 15.3\sd{0.8} & 6.8\sd{0.2} & 58.4\sd{8.9} & 29.2\sd{5.5} & 11.7\sd{2.2} \\
2 $\to$ 3 & 14.9\sd{0.4} & 7.6\sd{0.2} & 3.9\sd{0.1} & 31.5\sd{13.3} & 17.9\sd{7.9} & 6.0\sd{2.1} \\
3 $\to$ 4 & 7.8\sd{0.4} & 3.6\sd{0.4} & 2.4\sd{0.1} & 15.4\sd{4.9} & 8.5\sd{3.2} & 3.4\sd{0.7} \\
4 $\to$ 5 & 4.5\sd{0.3} & 1.9\sd{0.2} & 1.7\sd{0.1} & 8.2\sd{1.8} & 4.0\sd{0.8} & 2.3\sd{0.5} \\
5 $\to$ 6 & 3.3\sd{0.3} & 1.4\sd{0.2} & 1.1\sd{0.2} & 5.3\sd{0.8} & 2.5\sd{0.5} & 1.6\sd{0.2} \\
6 $\to$ 7 & 2.6\sd{0.3} & 1.0\sd{0.1} & 1.0\sd{0.1} & 4.0\sd{0.6} & 1.6\sd{0.3} & 1.4\sd{0.3} \\
7 $\to$ 8 & 2.3\sd{0.2} & 0.8\sd{0.1} & 1.0\sd{0.1} & 3.3\sd{0.2} & 1.3\sd{0.2} & 1.3\sd{0.2} \\
\bottomrule
\end{tabular}

\caption{Share of answerable items (\%) whose answer changes between consecutive loops: mean $\pm$ standard deviation over three seeds.}
\label{tab:flips}
\end{table*}

\subsection{Changes seen from the first and from the last loop}
\label{app:loops-dynamics}

Table~\ref{tab:dynamics} gives the numbers behind Figure~\ref{fig:dynamics}a,b for every loop, on the 9,645 answerable test items. The left half compares loop $T$ with loop 1: how many answers have changed, how many of them were fixed and how many were broken. The right half compares loop $T$ with loop 8: how many answers are already final, and how many the later loops will still fix or break. The middle column gives the share of answers that have settled by loop $T$.

\begin{table*}[t]
\centering
\small
\setlength{\tabcolsep}{6pt}
\begin{tabular}{@{}c ccc c ccc@{}}
\toprule
 & \multicolumn{3}{c}{Since loop 1 (\%)} & Settled & \multicolumn{3}{c}{Up to loop 8 (\%)} \\
\cmidrule(lr){2-4}\cmidrule(l){6-8}
Loop $T$ & changed & \textcolor{better}{fixed} & \textcolor{worse}{broken} & by $T$ (\%) & already final & \textcolor{better}{to be fixed} & \textcolor{worse}{to be broken} \\
\midrule
1 & 0.0\sd{0.0} & 0.0\sd{0.0} & 0.0\sd{0.0} & 56.7\sd{0.2} & 64.5\sd{0.8} & 21.0\sd{0.3} & 7.3\sd{0.3} \\
2 & 29.2\sd{0.3} & 15.3\sd{0.8} & 6.8\sd{0.2} & 74.2\sd{1.2} & 78.7\sd{0.7} & 11.1\sd{0.7} & 5.9\sd{0.2} \\
3 & 33.2\sd{0.4} & 19.0\sd{0.8} & 6.9\sd{0.1} & 83.6\sd{1.1} & 85.8\sd{0.8} & 6.4\sd{0.7} & 4.8\sd{0.2} \\
4 & 34.6\sd{0.4} & 20.5\sd{0.5} & 7.1\sd{0.1} & 89.1\sd{0.9} & 90.2\sd{0.9} & 4.0\sd{0.5} & 3.6\sd{0.4} \\
5 & 35.0\sd{0.5} & 20.8\sd{0.3} & 7.2\sd{0.2} & 92.5\sd{0.8} & 93.1\sd{0.7} & 2.7\sd{0.4} & 2.6\sd{0.3} \\
6 & 35.4\sd{0.6} & 21.0\sd{0.4} & 7.2\sd{0.3} & 95.3\sd{0.5} & 95.5\sd{0.5} & 1.7\sd{0.2} & 1.8\sd{0.2} \\
7 & 35.4\sd{0.9} & 21.0\sd{0.3} & 7.2\sd{0.4} & 97.7\sd{0.2} & 97.7\sd{0.2} & 0.8\sd{0.1} & 1.0\sd{0.1} \\
8 & 35.5\sd{0.8} & 21.0\sd{0.3} & 7.3\sd{0.3} & 100.0\sd{0.0} & 100.0\sd{0.0} & 0.0\sd{0.0} & 0.0\sd{0.0} \\
\bottomrule
\end{tabular}

\caption{How the answers of \sansi{} move over the loops (9,645 answerable test items; mean $\pm$ standard deviation over three seeds). Since loop 1: share of items whose answer at loop $T$ differs from that at loop 1, and the shares that went from wrong to right (fixed) and from right to wrong (broken). Settled by $T$: the answer does not change from loop $T$ on. Up to loop 8: share of items whose answer at loop $T$ is already that of loop 8, and the shares that the later loops will still fix or break.}
\label{tab:dynamics}
\end{table*}

Following single items through all eight loops gives five groups: 46.7\% of the answerable items are right at every loop, 20.0\% are fixed once and stay right, 6.5\% are broken once and stay wrong, 10.1\% move between right and wrong more than once, and 16.7\% are wrong at every loop. In total, 11.7\% of the items are right at some loop and wrong at loop 8. Choosing the best loop for every item would therefore make 83.3\% of the answerable items right, instead of the 71.6\% at loop 8 (Table~\ref{tab:confloop}). This is an upper bound that requires the gold answer, and we do not propose such a rule.

\subsection{When answers settle}
\label{app:loops-settle}

Table~\ref{tab:settle} groups the answerable items by the loop at which their answer settles, and gives for every group its share of the items, how often the answer at loop 8 is right, and the mean confidence at loop 8. More than half of the answers never change (56.7\%), and these are right in 82.2\% of the cases. The later an answer settles, the less often it is right and the lower its confidence.

\begin{table*}[t]
\centering
\small
\setlength{\tabcolsep}{5pt}
\begin{tabular}{@{}c ccc@{}}
\toprule
Settles at loop & Items (\%) & Right at loop 8 (\%) & Confidence at loop 8 \\
\midrule
1 & 56.7\sd{0.2} & \cellcolor{sansi!43}82.2\sd{1.1} & .894\sd{.014} \\
2 & 17.5\sd{0.9} & \cellcolor{sansi!32}68.7\sd{1.2} & .795\sd{.020} \\
3 & 9.4\sd{0.2} & \cellcolor{sansi!25}60.1\sd{0.5} & .714\sd{.017} \\
4 & 5.5\sd{0.3} & \cellcolor{sansi!17}50.7\sd{3.3} & .645\sd{.031} \\
5 & 3.4\sd{0.2} & \cellcolor{sansi!11}43.6\sd{4.3} & .581\sd{.032} \\
6 & 2.8\sd{0.3} & \cellcolor{sansi!12}44.1\sd{2.5} & .540\sd{.032} \\
7 & 2.4\sd{0.3} & \cellcolor{sansi!6}37.7\sd{0.6} & .497\sd{.021} \\
8 & 2.3\sd{0.2} & \cellcolor{sansi!5}35.6\sd{0.7} & .439\sd{.013} \\
\bottomrule
\end{tabular}

\caption{Items by the loop at which the answer of \sansi{} settles (9,645 answerable test items; mean $\pm$ standard deviation over three seeds). Shading of the accuracy column: darker is more often right.}
\label{tab:settle}
\end{table*}

Table~\ref{tab:difficulty} groups the items by how many of the three single-pass models (SmolLM2-1.7B, Qwen3.5-2B and Qwen3.5-4B; for each seed of \sansi{} we use the same seed of these models) answer them correctly. The items that no single-pass model answers correctly settle slightly earlier than those that one answers correctly (2.8 against 3.0 loops): \sansi{} answers only 19.8\% of them correctly at loop 8, and on 39.0\% of them it keeps its first answer. The loops help most between the extremes. The accuracy rises from 52.3\% at loop 1 to 77.4\% at loop 8 on the items that two single-pass models answer correctly and from 28.0\% to 47.1\% on those that one answers correctly, against 86.3\% to 95.0\% and 13.8\% to 19.8\% at the two ends.

\begin{table*}[t]
\centering
\small
\setlength{\tabcolsep}{4pt}
\begin{tabular}{@{}c c c c cc@{}}
\toprule
Single-pass & Items & Settles at loop & Never changes & \multicolumn{2}{c}{Accuracy (\%) at loop} \\
\cmidrule(l){5-6}
models right & (\%) & (mean) & (\%) & 1 & 8 \\
\midrule
3 & 45.5\sd{0.1} & 1.41\sd{0.02} & 80.8\sd{0.6} & 86.3\sd{0.9} & 95.0\sd{0.2} \\
2 & 22.9\sd{0.4} & 2.39\sd{0.06} & 41.4\sd{1.5} & 52.3\sd{1.6} & 77.4\sd{1.5} \\
1 & 16.0\sd{0.3} & 3.01\sd{0.14} & 27.5\sd{1.9} & 28.0\sd{0.6} & 47.1\sd{2.0} \\
0 & 15.6\sd{0.3} & 2.80\sd{0.10} & 39.0\sd{2.1} & 13.8\sd{0.1} & 19.8\sd{2.3} \\
\bottomrule
\end{tabular}

\caption{Items by the number of single-pass models that answer them correctly (9,645 answerable test items; mean $\pm$ standard deviation over three seeds). Never changes: \sansi{} keeps the answer of loop 1 through all eight loops.}
\label{tab:difficulty}
\end{table*}

\subsection{Confidence by whether the answer changed}
\label{app:loops-groups}

Table~\ref{tab:groups} splits the 8,879 items with one gold option into four groups, by whether their answer ever changed after loop 1 and by whether it is right at loop 8, and gives the mean confidence of every group at loops 1, 2, 4 and 8. The answers that never change are held with the highest confidence. This holds also when they are wrong: the wrong answers that never change (9.5\% of the items) end with a mean confidence of 0.782, which is higher than that of the right answers that were reached by a change (0.763).

\begin{table*}[t]
\centering
\small
\setlength{\tabcolsep}{5pt}
\begin{tabular}{@{}l c cccc@{}}
\toprule
 & Share & \multicolumn{4}{c}{Mean confidence at loop} \\
\cmidrule(lr){3-6}
Items (single gold answer) & (\%) & 1 & 2 & 4 & 8 \\
\midrule
Never changed, right at loop 8 & 48.0\sd{0.5} & .805\sd{.019} & .883\sd{.013} & .918\sd{.013} & .924\sd{.011} \\
Never changed, wrong at loop 8 & 9.5\sd{0.7} & .669\sd{.031} & .756\sd{.022} & .779\sd{.024} & .782\sd{.022} \\
Changed at least once, right at loop 8 & 24.7\sd{0.4} & .567\sd{.024} & .647\sd{.015} & .734\sd{.018} & .763\sd{.017} \\
Changed at least once, wrong at loop 8 & 17.7\sd{0.3} & .530\sd{.023} & .559\sd{.020} & .575\sd{.020} & .595\sd{.021} \\
\bottomrule
\end{tabular}

\caption{\sansi{}: four groups of items by whether the answer ever changed after loop 1 and whether it is right at loop 8, with their mean confidence (mean $\pm$ standard deviation over three seeds).}
\label{tab:groups}
\end{table*}

\subsection{Stopping rules}
\label{app:stop}

\begin{table*}[t]
\centering
\small
\setlength{\tabcolsep}{4pt}
\begin{minipage}[t]{0.49\textwidth}\centering
\begin{tabular}[t]{@{}l ccc@{}}
\toprule
Rule and setting & Accuracy (\%) & Loops & Cost \\
\midrule
\rowcolor{groupbg}\multicolumn{4}{@{}l}{\emph{Fixed budget}} \\
$T = 1$ & 58.4\sd{0.7} & 1.00\sd{0.00} & 1.0 \\
$T = 2$ & 66.9\sd{0.6} & 2.00\sd{0.00} & 1.9 \\
$T = 3$ & 70.4\sd{0.6} & 3.00\sd{0.00} & 2.9 \\
$T = 4$ & 71.6\sd{0.5} & 4.00\sd{0.00} & 3.8 \\
$T = 5$ & 71.9\sd{0.7} & 5.00\sd{0.00} & 4.8 \\
$T = 6$ & 72.1\sd{0.7} & 6.00\sd{0.00} & 5.8 \\
$T = 7$ & 72.1\sd{0.7} & 7.00\sd{0.00} & 6.7 \\
$T = 8$ & 72.0\sd{0.7} & 8.00\sd{0.00} & 7.7 \\
\rowcolor{groupbg}\multicolumn{4}{@{}l}{\emph{Confidence reaches $\tau$}} \\
$\tau = 0.5$ & 63.3\sd{0.7} & 1.72\sd{0.14} & 1.6 \\
$\tau = 0.6$ & 65.7\sd{0.5} & 2.18\sd{0.18} & 2.1 \\
$\tau = 0.7$ & 67.7\sd{0.4} & 2.86\sd{0.23} & 2.7 \\
$\tau = 0.8$ & 69.5\sd{0.4} & 3.73\sd{0.22} & 3.6 \\
$\tau = 0.85$ & 70.2\sd{0.5} & 4.23\sd{0.22} & 4.1 \\
$\tau = 0.9$ & 70.9\sd{0.6} & 4.81\sd{0.22} & 4.6 \\
$\tau = 0.93$ & 71.3\sd{0.6} & 5.25\sd{0.22} & 5.0 \\
$\tau = 0.95$ & 71.6\sd{0.6} & 5.61\sd{0.21} & 5.4 \\
$\tau = 0.97$ & 71.8\sd{0.6} & 6.08\sd{0.20} & 5.8 \\
$\tau = 0.98$ & 71.9\sd{0.6} & 6.40\sd{0.16} & 6.1 \\
$\tau = 0.99$ & 72.0\sd{0.6} & 6.84\sd{0.13} & 6.6 \\
$\tau = 0.995$ & 72.0\sd{0.6} & 7.18\sd{0.11} & 6.9 \\
$\tau = 0.999$ & 72.0\sd{0.7} & 7.77\sd{0.06} & 7.5 \\
\rowcolor{groupbg}\multicolumn{4}{@{}l}{\emph{Same answer for $m$ consecutive loops}} \\
$m = 2$ & 68.7\sd{0.6} & 2.39\sd{0.01} & 2.3 \\
$m = 3$ & 71.1\sd{0.6} & 3.65\sd{0.01} & 3.5 \\
$m = 4$ & 71.8\sd{0.7} & 4.79\sd{0.02} & 4.6 \\
$m = 5$ & 71.9\sd{0.6} & 5.80\sd{0.02} & 5.6 \\
$m = 6$ & 72.0\sd{0.6} & 6.68\sd{0.01} & 6.4 \\
$m = 7$ & 72.0\sd{0.6} & 7.43\sd{0.00} & 7.1 \\
$m = 8$ & 72.0\sd{0.7} & 8.00\sd{0.00} & 7.7 \\
\bottomrule
\end{tabular}
\end{minipage}\hfill
\begin{minipage}[t]{0.49\textwidth}\centering
\begin{tabular}[t]{@{}l ccc@{}}
\toprule
Rule and setting & Accuracy (\%) & Loops & Cost \\
\midrule
\rowcolor{groupbg}\multicolumn{4}{@{}l}{\emph{KL divergence from the previous loop below $\varepsilon$}} \\
$\varepsilon = 0.3$ & 68.4\sd{0.7} & 2.29\sd{0.01} & 2.2 \\
$\varepsilon = 0.1$ & 69.7\sd{0.9} & 2.67\sd{0.04} & 2.6 \\
$\varepsilon = 0.05$ & 70.6\sd{0.8} & 2.99\sd{0.04} & 2.9 \\
$\varepsilon = 0.02$ & 71.3\sd{0.8} & 3.49\sd{0.05} & 3.3 \\
$\varepsilon = 0.01$ & 71.6\sd{0.7} & 3.92\sd{0.06} & 3.8 \\
$\varepsilon = 0.005$ & 71.8\sd{0.7} & 4.40\sd{0.08} & 4.2 \\
$\varepsilon = 0.002$ & 72.0\sd{0.7} & 5.06\sd{0.09} & 4.9 \\
$\varepsilon = 0.001$ & 72.0\sd{0.8} & 5.55\sd{0.10} & 5.3 \\
$\varepsilon = 0.0005$ & 72.0\sd{0.7} & 5.99\sd{0.11} & 5.8 \\
$\varepsilon = 0.0002$ & 71.9\sd{0.7} & 6.53\sd{0.12} & 6.3 \\
$\varepsilon = 0.0001$ & 71.9\sd{0.7} & 6.85\sd{0.11} & 6.6 \\
\rowcolor{groupbg}\multicolumn{4}{@{}l}{\emph{Ouro's exit gate}} \\
$q = 0.05$ & 70.9\sd{0.6} & 2.93\sd{0.17} & 2.8 \\
$q = 0.1$ & 71.3\sd{0.5} & 3.14\sd{0.16} & 3.0 \\
$q = 0.2$ & 71.7\sd{0.6} & 3.50\sd{0.13} & 3.4 \\
$q = 0.3$ & 71.8\sd{0.7} & 3.84\sd{0.08} & 3.7 \\
$q = 0.4$ & 71.9\sd{0.7} & 4.14\sd{0.10} & 4.0 \\
\rowcolor{sansibg}$q = 0.5$ & 72.0\sd{0.8} & 4.51\sd{0.11} & 4.3 \\
$q = 0.6$ & 72.0\sd{0.8} & 4.91\sd{0.11} & 4.7 \\
$q = 0.7$ & 72.1\sd{0.7} & 5.39\sd{0.16} & 5.2 \\
$q = 0.8$ & 72.2\sd{0.7} & 6.08\sd{0.18} & 5.8 \\
$q = 0.9$ & 72.1\sd{0.6} & 7.14\sd{0.22} & 6.9 \\
$q = 0.95$ & 72.0\sd{0.7} & 7.83\sd{0.09} & 7.5 \\
$q = 0.99$ & 72.0\sd{0.7} & 8.00\sd{0.00} & 7.7 \\
\bottomrule
\end{tabular}
\end{minipage}

\caption{Stopping rules, simulated on the stored option probabilities of the test items: every setting that we evaluated (Figure~\ref{fig:stop}a). Each item stops at the first loop that meets the rule, and at loop 8 at the latest. All three seeds use the same setting; accuracy and loops are the mean $\pm$ standard deviation over the seeds. Loops: loops run per item on average. Cost: GPU time as in Table~\ref{tab:main} (one loop of Ouro-1.4B = 1), that is, the mean number of loops times 0.96, assuming that a finished item leaves its batch. KL: the KL divergence of the option probabilities from those of the previous loop \citep{geiping2025recurrent}. Ouro's exit gate stops once its cumulative exit probability reaches $q$ \citep{zhu2025ouro}; with $q = 1$ it never stops early and equals $T = 8$. Shaded: $q = 0.5$.}
\label{tab:stop}
\end{table*}

\begin{table}[t]
\centering
\small
\setlength{\tabcolsep}{3pt}
\begin{tabular}{@{}l cc@{}}
\toprule
Items & Share (\%) & Gate stops \\
\midrule
\rowcolor{groupbg}\multicolumn{3}{@{}l}{\emph{Answerable items, by the loop at which the answer settles}} \\
Settles at loop 1 & 56.7 & 4.22\sd{0.09} \\
Settles at loop 2 & 17.5 & 4.59\sd{0.14} \\
Settles at loop 3 & 9.4 & 4.91\sd{0.10} \\
Settles at loop 4 & 5.5 & 5.10\sd{0.15} \\
Settles at loop 5 & 3.4 & 5.09\sd{0.15} \\
Settles at loop 6 & 2.8 & 5.02\sd{0.12} \\
Settles at loop 7 & 2.4 & 4.96\sd{0.15} \\
Settles at loop 8 & 2.3 & 5.02\sd{0.07} \\
\rowcolor{groupbg}\multicolumn{3}{@{}l}{\emph{Answerable items, by difficulty}} \\
All three & 45.5 & 4.17\sd{0.08} \\
Two & 22.9 & 4.67\sd{0.14} \\
One & 16.0 & 4.85\sd{0.12} \\
None & 15.6 & 4.78\sd{0.12} \\
\rowcolor{groupbg}\multicolumn{3}{@{}l}{\emph{Answerable items, by test group}} \\
In-distribution & 19.9 & 4.14\sd{0.12} \\
Near transfer & 18.2 & 4.42\sd{0.18} \\
Far transfer & 59.6 & 4.62\sd{0.09} \\
JevBench & 2.4 & 4.60\sd{0.06} \\
\rowcolor{groupbg}\multicolumn{3}{@{}l}{\emph{All test items, by group}} \\
Answerable & 96.2 & 4.49\sd{0.11} \\
Answerable, right at loop 8 & 68.8 & 4.36\sd{0.10} \\
Answerable, wrong at loop 8 & 27.4 & 4.81\sd{0.13} \\
Unanswerable & 3.8 & 5.01\sd{0.19} \\
\bottomrule
\end{tabular}

\caption{Where Ouro's exit gate ($q = 0.5$) stops: the mean loop $\pm$ standard deviation over three seeds. Upper blocks: answerable items, by the loop from which their answer no longer changes, by difficulty (how many of the three single-pass models answer the item correctly, as in Table~\ref{tab:difficulty}) and by test group, with their share of the answerable items. Lower block: groups of items, with their share of all test items.}
\label{tab:stop-settle}
\end{table}

\begin{table}[t]
\centering
\small
\setlength{\tabcolsep}{3pt}
\begin{tabular}{@{}l ccc@{}}
\toprule
Serving & Loops & Saving (\%) & Cost \\
\midrule
Per item & 4.51\sd{0.11} & 44 & 4.3 \\
Static batches of 8 items & 5.43\sd{0.14} & 32 & 5.2 \\
Static batches of 16 items & 5.70\sd{0.18} & 29 & 5.5 \\
Static batches of 32 items & 5.97\sd{0.19} & 25 & 5.7 \\
Static batches of 64 items & 6.19\sd{0.21} & 23 & 5.9 \\
Static batches of the test pass & 5.91\sd{0.30} & 26 & 5.7 \\
\bottomrule
\end{tabular}

\caption{Loops per item with Ouro's exit gate ($q = 0.5$) under different ways of serving (10,027 test items; mean $\pm$ standard deviation over three seeds). Per item: every item stops at its own loop, as with one call per item or with a batch that drops finished items. Static batches: every item runs as many loops as the longest item of its batch; batches of 8 to 64 items are formed in random order (mean over five orderings), and the batches of the test pass are those of Table~\ref{tab:params} (items sorted by length, at most 32,000 tokens and 64 items per batch). Saving: fewer loops than the fixed budget $T = 8$. Cost: as in Table~\ref{tab:stop}.}
\label{tab:stop-batch}
\end{table}

\begin{figure*}[t]
\centering
\begin{subfigure}[t]{3.55in}\includegraphics[width=\linewidth]{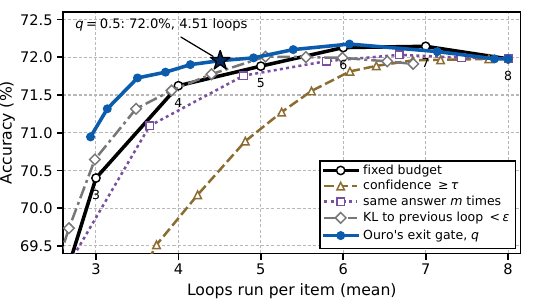}\caption{Accuracy against the mean number of loops}\end{subfigure}\hfill
\begin{subfigure}[t]{2.6in}\includegraphics[width=\linewidth]{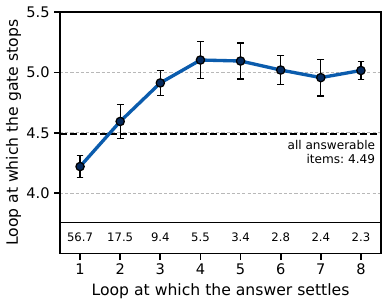}\caption{Where Ouro's exit gate stops}\end{subfigure}
\caption{Stopping rules on the test items (means of three seeds). (a) Each line is one rule and each point one of its settings (values in Table~\ref{tab:stop}); numbers below the black line: the fixed budget $T$; star: the gate with $q = 0.5$. The vertical axis starts at 69.4\%; the settings below it are in the table. (b) Answerable items, by the loop from which their answer no longer changes: the mean loop at which the gate ($q = 0.5$) stops (error bars: $\pm$1 standard deviation over the seeds); dashed: all answerable items; numbers above the axis: share of the answerable items (\%). Values in Table~\ref{tab:stop-settle}.}
\label{fig:stop}
\end{figure*}

A model that is read after every loop can stop early on items that need few loops. We simulate four rules on the stored option probabilities of the test items, without running the model again: stop at the first loop whose confidence reaches $\tau$, as in early exit based on confidence \citep{xin2020deebert}; stop once the answer has been the same for $m$ consecutive loops, as in patience-based early exit \citep{zhou2020bert}; stop once the KL divergence of the option probabilities from those of the previous loop falls below $\varepsilon$, the rule that \citet{geiping2025recurrent} use for their recurrent-depth model; and Ouro's own exit gate \citep{zhu2025ouro}, a linear layer of the backbone whose weights our training leaves frozen, read at the position of the readout. At loop $t$ the gate gives $\lambda_t$, the probability that the computation stops there given that it has not stopped before, so the probability of stopping at loop $t$ is $\lambda_t \prod_{j<t} (1 - \lambda_j)$; the model stops at the first loop at which the sum of these probabilities reaches $q$. Ouro's default, $q = 1$, never stops early. Every item stops at loop 8 at the latest. Table~\ref{tab:stop} lists every setting that we evaluated and Figure~\ref{fig:stop}a plots them; each setting is the same for all three seeds, so no setting is chosen on data.

The gate is a usable rule and needs no tuning. With $q = 0.5$, \sansi{} keeps the accuracy of loop 8 (72.0\% against 72.0\%; 0.0 [$-$0.3, 0.2]) with 4.5 loops on average, and is 0.3 points [0.1, 0.6] more accurate than a fixed budget of four loops; thresholds from 0.2 to 0.7 change the accuracy by less than half a point. The rules based on confidence and on a stable answer do not beat a fixed budget at any setting: with $\tau = 0.90$, for example, \sansi{} runs 4.8 loops on average and reaches 70.9\%, below the 71.6\% of four fixed loops. Confidence keeps rising after the answers settle and rises on wrong answers too (\S\ref{sec:analysis}), so it does not tell when to stop. The rule based on the KL divergence is slightly better than a fixed budget but not better than the gate: with $\varepsilon = 0.005$, \sansi{} runs 4.40 loops on average and reaches 71.8\%, while the gate with $q = 0.5$ reaches 72.0\% with 4.51 loops.

The gate also gives more loops to the items that need them (Figure~\ref{fig:stop}b; Table~\ref{tab:stop-settle}). Answerable items whose answer settles at loop 1, more than half of them, stop after 4.22 loops on average, and items whose answer settles at loop 4 after 5.10. Unanswerable items stop after 5.01 loops, against 4.49 for answerable ones, and answerable items that are wrong at loop 8 after 4.81, against 4.36 for those that are right. By difficulty, measured as in Table~\ref{tab:difficulty} by how many of the three single-pass models answer an item correctly, the gate stops after 4.17 loops on the items that all three solve, 4.67 on those that two solve, 4.85 on those that one solves and 4.78 on those that none solves; by test group, it stops earliest on in-distribution items (4.14) and latest on far transfer (4.62). Items whose answer settles at loop 6 or later also stop after about five loops, before their answer settles; since the last loops fix about as many answers as they break (\S\ref{sec:rq2}), this costs little accuracy.

\paragraph{From loops to time.} The saving in loops is a saving in time when every item runs only its own loops, as with one call per item or with a batch that drops finished items; our test pass does neither (Appendix~\ref{app:training}), and serving of this kind is left to future work. With static batches, every item runs as many loops as the longest item of its batch. With $q = 0.5$, the gate stops 7.2\% of the items at loop 3, 41.7\% at loop 4, 44.8\% at loop 5, 6.1\% at loop 6 and 0.4\% at loops 7 and 8, and none earlier. Table~\ref{tab:stop-batch} gives the loops per item for the gate with $q = 0.5$ under both kinds of serving: random static batches of 32 items run 5.97 loops on average and the batches of our test pass 5.91, a saving of 25--26\% instead of 44\%, because the gate stops 86.5\% of the items at loop 4 or 5 and only 0.4\% at loop 7 or 8, so that few batches wait for an item that runs to the end.

\section{Probabilities: Additional Results}
\label{app:probs}

This appendix supports \S\ref{sec:analysis}. It gives the intervals of the differences in the probability measures (Appendix~\ref{app:probs-ci}), the confidence of right and wrong answers and selective prediction (Appendix~\ref{app:probs-conf}), the calibration of \sansi{} and of the other models (Appendix~\ref{app:probs-cal}), and the results on items with missing evidence (Appendix~\ref{app:probs-evid}).

\subsection{Intervals of the differences}
\label{app:probs-ci}

Table~\ref{tab:probci} gives the differences in the probability measures that \S\ref{sec:analysis} reports, with their 95\% bootstrap intervals: the ECE, the AUROC with which confidence separates right from wrong answers, the evidence AUROC and the hard-answer rate. Every row gives the two values that are compared and their difference. As for the accuracy (Appendix~\ref{app:main-ci}), groups of related test items are resampled and the three seeds are averaged; the measure is recomputed on every resample.

\begin{table*}[t]
\centering
\small
\setlength{\tabcolsep}{8pt}
\begin{tabular}{@{}l cc c@{}}
\toprule
Difference (first $-$ second) & First & Second & Difference [95\% interval] \\
\midrule
\rowcolor{groupbg}\multicolumn{4}{@{}l}{\emph{ECE (answerable items) $\downarrow$}} \\
\sansi{} loop 3 $-$ loop 1 & 0.082 & 0.105 & \dbetter{-0.023}{-0.032}{-0.013} \\
\sansi{} loop 8 $-$ loop 3 & 0.093 & 0.082 & \dworse{+0.010}{+0.005}{+0.016} \\
\sansi{} loop 1 $-$ Ouro one loop & 0.105 & 0.137 & \dbetter{-0.032}{-0.038}{-0.027} \\
\sansi{} loop 8 $-$ Ouro one loop & 0.093 & 0.137 & \dbetter{-0.045}{-0.054}{-0.034} \\
\sansi{} loop 8 $-$ SmolLM2-1.7B & 0.093 & 0.069 & \dworse{+0.024}{+0.015}{+0.033} \\
\sansi{} loop 8 $-$ Qwen3.5-2B & 0.093 & 0.123 & \dbetter{-0.030}{-0.037}{-0.023} \\
\sansi{} loop 8 $-$ Qwen3.5-4B & 0.093 & 0.113 & \dbetter{-0.020}{-0.027}{-0.014} \\
Mean of loops 1--8 $-$ loop 8 & 0.044 & 0.093 & \dbetter{-0.049}{-0.052}{-0.045} \\
Mean of loops 1--8 $-$ loop 8, near transfer & 0.084 & 0.149 & \dbetter{-0.065}{-0.075}{-0.056} \\
\midrule
\rowcolor{groupbg}\multicolumn{4}{@{}l}{\emph{AUROC, right against wrong answers (items with one gold option) $\uparrow$}} \\
\sansi{} loop 8 $-$ loop 1 & 0.795 & 0.760 & \dbetter{+0.035}{+0.025}{+0.045} \\
Mean of loops 1--8 $-$ loop 8 & 0.790 & 0.795 & \dns{-0.004}{-0.009}{+0.000} \\
\midrule
\rowcolor{groupbg}\multicolumn{4}{@{}l}{\emph{Evidence AUROC (786 items) $\uparrow$}} \\
\sansi{} loop 3 $-$ loop 1 & 0.921 & 0.835 & \dbetter{+0.086}{+0.064}{+0.108} \\
\sansi{} loop 8 $-$ loop 1 & 0.935 & 0.835 & \dbetter{+0.099}{+0.075}{+0.125} \\
\sansi{} loop 1 $-$ Ouro one loop & 0.835 & 0.837 & \dns{-0.002}{-0.014}{+0.010} \\
\sansi{} loop 8 $-$ Ouro one loop & 0.935 & 0.837 & \dbetter{+0.098}{+0.075}{+0.122} \\
\sansi{} loop 8 $-$ SmolLM2-1.7B & 0.935 & 0.765 & \dbetter{+0.170}{+0.142}{+0.199} \\
\sansi{} loop 8 $-$ Qwen3.5-2B & 0.935 & 0.896 & \dbetter{+0.039}{+0.027}{+0.053} \\
Qwen3.5-4B $-$ \sansi{} loop 8 & 0.948 & 0.935 & \dbetter{+0.014}{+0.003}{+0.025} \\
\midrule
\rowcolor{groupbg}\multicolumn{4}{@{}l}{\emph{Hard-answer rate on the 382 unanswerable items (\%) $\downarrow$}} \\
\sansi{} loop 8 $-$ loop 1 & 17.5 & 27.1 & \dbetter{-9.7}{-13.4}{-6.0} \\
\sansi{} loop 8 $-$ Ouro one loop & 17.5 & 31.9 & \dbetter{-14.5}{-18.5}{-10.8} \\
\sansi{} loop 8 $-$ SmolLM2-1.7B & 17.5 & 36.3 & \dbetter{-18.8}{-23.6}{-14.2} \\
\sansi{} loop 8 $-$ Qwen3.5-4B & 17.5 & 18.2 & \dns{-0.7}{-2.9}{+1.5} \\
\bottomrule
\end{tabular}

\caption{Differences in the probability metrics with 95\% bootstrap intervals (means of three seeds; the statistic is recomputed on every resample of the groups of related test items). First, second: the two values that are compared. Colours: \textcolor{better}{\textbf{teal}}, the first term is significantly better (taking the direction of the metric into account; the interval excludes 0); \textcolor{worse}{\textbf{red}}, significantly worse; \textcolor{black!55}{grey}, the interval includes 0.}
\label{tab:probci}
\end{table*}

\subsection{Confidence of right and wrong answers}
\label{app:probs-conf}

Table~\ref{tab:confloop} gives, for every loop, the accuracy and the mean confidence of \sansi{} on the answerable items, their difference and the ECE. The difference between confidence and accuracy is smallest at loop 3 and grows again afterwards, and the ECE follows it closely. Table~\ref{tab:perloop} in Appendix~\ref{app:loops-every} gives the mean confidence of right and of wrong answers separately at every loop.

\begin{table*}[t]
\centering
\small
\setlength{\tabcolsep}{8pt}
\begin{tabular}{@{}c cccc@{}}
\toprule
Loop & Accuracy (\%) & Mean confidence & Confidence $-$ accuracy & ECE \\
\midrule
1 & 57.9\sd{0.7} & .683\sd{.022} & $+0.105$ & .105\sd{.020} \\
2 & 66.3\sd{0.6} & .750\sd{.015} & $+0.086$ & .088\sd{.009} \\
3 & 70.0\sd{0.6} & .780\sd{.016} & $+0.080$ & .082\sd{.010} \\
4 & 71.2\sd{0.6} & .793\sd{.016} & $+0.081$ & .085\sd{.009} \\
5 & 71.4\sd{0.7} & .798\sd{.016} & $+0.084$ & .087\sd{.008} \\
6 & 71.7\sd{0.7} & .802\sd{.016} & $+0.085$ & .089\sd{.010} \\
7 & 71.7\sd{0.8} & .804\sd{.017} & $+0.087$ & .090\sd{.012} \\
8 & 71.6\sd{0.7} & .805\sd{.016} & $+0.090$ & .093\sd{.012} \\
\bottomrule
\end{tabular}

\caption{\sansi{} after every loop on the 9,645 answerable test items: accuracy, mean confidence, their difference, and the ECE (mean $\pm$ standard deviation over three seeds).}
\label{tab:confloop}
\end{table*}

The loop at which an answer settles is a weaker signal of a wrong answer than confidence. On the 8,879 items with one gold option, the AUROC with which the settling loop separates right from wrong answers is 0.685, against 0.795 for the confidence at loop 8.

\paragraph{Selective prediction.} A model that answers only the items on which it is most confident can be judged by the area under its risk--coverage curve (AURC, in \%; lower is better), the mean error over all shares of answered items \citep{kamath2020selective}. On the 8,879 items with one gold option, the AURC of \sansi{} falls with every loop, from 22.0 at loop 1 to 12.2 at loop 3 and 10.9 at loop 8, unlike the ECE, which rises after loop 3: the later loops add accuracy and keep right answers ahead of wrong ones, while their probabilities become too confident. At loop 8, \sansi{} (10.9) is far below the single-pass models of its size (22.3 for SmolLM2-1.7B, 21.5 for Ouro-1.4B with one loop, 14.9 for Qwen3.5-2B) and close to Qwen3.5-4B (9.8) and Kev-4B (our data) (9.6); \sansi{}-2.6B is the lowest (8.6).

\subsection{Calibration}
\label{app:probs-cal}

The fall of the ECE from loop 1 to loop 3 and its rise from loop 3 to loop 8 occur in each of the three seeds. At every loop, the ECE of \sansi{} is below that of the same backbone trained with one loop (0.137) and of the two Qwen3.5 models (0.123 and 0.113), and above that of SmolLM2-1.7B (0.069; Table~\ref{tab:calib}). The table also gives the other measures of the probabilities for all models: the ECE by test group, overconfidence, confident errors, the confidence of right and of wrong answers, the AUROC, the negative log-likelihood and the distance to the crowd distribution.

\begin{table*}[t]
\centering
\resizebox{\textwidth}{!}{
\begin{tabular}{@{}l cccc c c cc c c c@{}}
\toprule
 & \multicolumn{4}{c}{ECE $\downarrow$} & Over- & Conf. & \multicolumn{2}{c}{Confidence} & AUROC & NLL & TV \\
\cmidrule(lr){2-5}\cmidrule(lr){8-9}
Model & All & In-dist. & Near & Far & conf. & errors (\%) & right & wrong & r/w $\uparrow$ & $\downarrow$ & crowd $\downarrow$ \\
\midrule
\mkSmol{}SmolLM2-1.7B & .069 & .021 & .041 & .101 & $+0.069$ & 5.1 & .744 & .531 & .762 & 1.044 & .269 \\
 & \sd{.020} & \sd{.006} & \sd{.015} & \sd{.022} & \sd{.020} & \sd{1.0} & \sd{.016} & \sd{.018} & \sd{.004} & \sd{0.027} & \sd{.006} \\[2pt]
\mkOuro{}Ouro-1.4B, one loop & .137 & .036 & .082 & .187 & $+0.137$ & 9.0 & .802 & .600 & .762 & 1.090 & .315 \\
 & \sd{.009} & \sd{.012} & \sd{.014} & \sd{.015} & \sd{.009} & \sd{0.8} & \sd{.008} & \sd{.014} & \sd{.006} & \sd{0.031} & \sd{.008} \\[2pt]
\rowcolor{sansibg}\mkSansi{}\sansi{}, loop 4 & .085 & .046 & .143 & .084 & $+0.081$ & 8.6 & .857 & .647 & .788 & 0.782 & .274 \\
\rowcolor{sansibg} & \sd{.009} & \sd{.007} & \sd{.018} & \sd{.017} & \sd{.012} & \sd{0.7} & \sd{.015} & \sd{.019} & \sd{.009} & \sd{0.016} & \sd{.010} \\[2pt]
\rowcolor{sansibg}\mkSansi{}\sansi{}, loop 8 & .093 & .048 & .149 & .092 & $+0.090$ & 9.2 & .869 & .660 & .795 & 0.788 & .278 \\
\rowcolor{sansibg} & \sd{.012} & \sd{.004} & \sd{.031} & \sd{.019} & \sd{.013} & \sd{0.8} & \sd{.013} & \sd{.020} & \sd{.005} & \sd{0.018} & \sd{.009} \\[2pt]
\rowcolor{sansibg}\mkSansiL{}\sansi{}-2.6B, loop 8 & .078 & .046 & .106 & .082 & $+0.073$ & 8.5 & .885 & .669 & .804 & 0.685 & .276 \\
\rowcolor{sansibg} & \sd{.000} & \sd{.009} & \sd{.027} & \sd{.011} & \sd{.003} & \sd{0.3} & \sd{.004} & \sd{.014} & \sd{.004} & \sd{0.008} & \sd{.011} \\[2pt]
\mkSansiV{}Last loop only, loop 8 & .073 & .041 & .100 & .077 & $+0.070$ & 7.4 & .841 & .623 & .792 & 0.774 & .279 \\
 & \sd{.009} & \sd{.012} & \sd{.019} & \sd{.009} & \sd{.009} & \sd{1.0} & \sd{.008} & \sd{.011} & \sd{.008} & \sd{0.030} & \sd{.014} \\[2pt]
\mkSansiV{}Reinforcement learning, loop 8 & .076 & .035 & .129 & .076 & $+0.074$ & 8.2 & .851 & .632 & .793 & 0.783 & .288 \\
 & \sd{.015} & \sd{.010} & \sd{.047} & \sd{.010} & \sd{.018} & \sd{0.8} & \sd{.018} & \sd{.023} & \sd{.002} & \sd{0.030} & \sd{.020} \\[2pt]
\mkSansiV{}Cross-entropy only, loop 8 & .104 & .055 & .153 & .108 & $+0.103$ & 9.9 & .877 & .676 & .791 & 0.829 & .279 \\
 & \sd{.007} & \sd{.006} & \sd{.029} & \sd{.004} & \sd{.008} & \sd{1.4} & \sd{.008} & \sd{.019} & \sd{.007} & \sd{0.030} & \sd{.008} \\[2pt]
\mkQwenS{}Qwen3.5-2B & .123 & .051 & .176 & .131 & $+0.122$ & 10.7 & .851 & .662 & .773 & 0.963 & .298 \\
 & \sd{.007} & \sd{.017} & \sd{.024} & \sd{.013} & \sd{.006} & \sd{1.1} & \sd{.008} & \sd{.014} & \sd{.012} & \sd{0.026} & \sd{.014} \\[2pt]
\mkQwen{}Qwen3.5-4B & .113 & .054 & .178 & .112 & $+0.109$ & 11.3 & .895 & .703 & .802 & 0.788 & .302 \\
 & \sd{.003} & \sd{.003} & \sd{.022} & \sd{.008} & \sd{.002} & \sd{0.1} & \sd{.008} & \sd{.005} & \sd{.005} & \sd{0.005} & \sd{.004} \\[2pt]
\mkKev{}Kev-4B (our data) & .119 & .063 & .150 & .129 & $+0.118$ & 11.0 & .917 & .730 & .797 & 0.859 & .254 \\
 & \sd{.004} & \sd{.003} & \sd{.012} & \sd{.009} & \sd{.004} & \sd{0.3} & \sd{.004} & \sd{.009} & \sd{.005} & \sd{0.025} & \sd{.016} \\[2pt]
\mkJev{}Jev API & .066 & .055 & .094 & .065 & $+0.064$ & 7.2 & .913 & .696 & .830 & 0.669 & .276 \\
\bottomrule
\end{tabular}
}
\caption{Probability quality: means of three seeds, with the standard deviation over the seeds on the line below (the Jev API is a single run). Overconf.: mean confidence minus accuracy. Conf.\ errors: wrong with top probability $\ge 0.8$. TV crowd: total variation distance to the crowd distribution (766 items).}
\label{tab:calib}
\end{table*}

Kev's recipe ends with a step that our recipe does not have: one temperature is fitted on the development split (the 2,190 items with one gold option that can be answered) and applied to the probabilities. Table~\ref{tab:calib-dev} applies this step to four models. As trained, their ECE lies between 0.078 and 0.119. With the fitted temperature it lies between 0.022 and 0.029 for all of them, and their hard-answer rates fall to between 11.0\% and 14.7\%, while the evidence AUROC does not change. The differences in calibration between the models as trained therefore largely disappear once labelled development data are used, and none of the recipes has an advantage in calibration (see also the comparison with temperature scaling in Appendix~\ref{app:averaging}).

\begin{table*}[t]
\centering
\small
\resizebox{\textwidth}{!}{
\begin{tabular}{@{}l c cc cccc cc@{}}
\toprule
 & & \multicolumn{2}{c}{As trained} & \multicolumn{6}{c}{With the temperature fitted on the development split} \\
\cmidrule(lr){3-4}\cmidrule(l){5-10}
 & Fitted & & Hard & \multicolumn{4}{c}{ECE $\downarrow$} & Hard & Evidence \\
\cmidrule(lr){5-8}
Model & temperature & ECE $\downarrow$ & ans.\ (\%) & All & In-dist. & Near & Far & ans.\ (\%) & AUROC $\uparrow$ \\
\midrule
\mkKev{}Kev-4B (our data) & 1.99 & .119\sd{.004} & 16.3\sd{0.6} & .029\sd{.005} & .025\sd{.006} & .043\sd{.006} & .034\sd{.005} & 11.0\sd{0.7} & .945\sd{.002} \\
\mkQwen{}Qwen3.5-4B & 1.61 & .113\sd{.003} & 18.2\sd{2.5} & .028\sd{.002} & .032\sd{.007} & .076\sd{.031} & .035\sd{.004} & 12.8\sd{1.8} & .948\sd{.003} \\
\rowcolor{sansibg}\mkSansiL{}\sansi{}-2.6B, loop 8 & 1.40 & .078\sd{.000} & 18.2\sd{2.4} & .022\sd{.003} & .023\sd{.006} & .039\sd{.020} & .033\sd{.003} & 14.7\sd{1.0} & .942\sd{.001} \\
\rowcolor{sansibg}\mkSansi{}\sansi{}, loop 8 & 1.47 & .093\sd{.012} & 17.5\sd{1.3} & .026\sd{.003} & .024\sd{.004} & .067\sd{.029} & .031\sd{.003} & 13.4\sd{1.9} & .934\sd{.003} \\
\bottomrule
\end{tabular}
}
\caption{Four models as trained and with one temperature fitted on the development split, the last step of Kev's recipe (10,027 test items; mean $\pm$ standard deviation over three seeds; fitted temperature: mean of the seeds).}
\label{tab:calib-dev}
\end{table*}

\begin{figure*}[t]
\centering
\includegraphics[width=\textwidth]{figures/fig_perloop_legend.pdf}\\[1pt]
\begin{subfigure}[t]{0.3333\textwidth}\includegraphics[width=\linewidth]{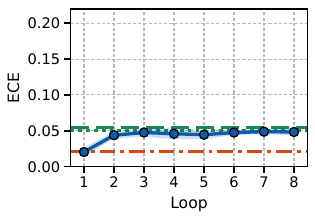}\caption{In-distribution}\end{subfigure}%
\begin{subfigure}[t]{0.3333\textwidth}\includegraphics[width=\linewidth]{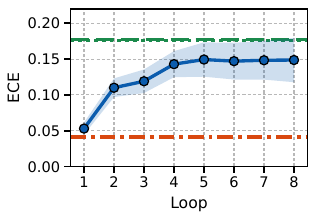}\caption{Near transfer}\end{subfigure}%
\begin{subfigure}[t]{0.3333\textwidth}\includegraphics[width=\linewidth]{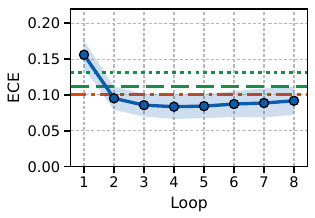}\caption{Far transfer}\end{subfigure}
\caption{ECE of \sansi{} after every loop, by distance from the training data (answerable items; line: mean of three seeds; band: one standard deviation). Horizontal lines: the single-pass models, whose standard deviations are in Table~\ref{tab:calib}; the lines of the two Qwen3.5 models nearly coincide in (a) and (b).}
\label{fig:ece-tier}
\end{figure*}

Figure~\ref{fig:ece-tier} splits the ECE of Figure~\ref{fig:perloop}c by distance from the training data. On far transfer the first loop is confident and often wrong (ECE 0.156); the next loops raise the accuracy by 17 points, and the ECE falls to 0.084 at loop 4. On near transfer the accuracy rises by only 6.6 points while the confidence rises as much as elsewhere, and the ECE grows from 0.053 to 0.149. This is not specific to looping: Qwen3.5-4B has an ECE of 0.178 on near transfer.

\subsection{Missing evidence}
\label{app:probs-evid}

Table~\ref{tab:unans} gives the hard-answer rate on the unanswerable items by test group, and the evidence AUROC. \sansi{} gives a hard answer to 27.1\% of the unanswerable items at loop 1, to 18.1\% at loop 4 and to 17.5\% at loop 8, and its evidence AUROC is 0.835, 0.928 and 0.935 at these loops. The same backbone trained with one loop reaches 0.837 and gives a hard answer to 31.9\% of the unanswerable items. SmolLM2-1.7B, which sees the same training targets, reaches 0.765 and 36.3\%, and Qwen3.5-2B 0.896 and 27.9\%. Qwen3.5-4B remains slightly ahead in one pass (0.948; +0.014 [0.003, 0.025]) with about the same hard-answer rate (18.2\%). Kev-4B (our data) is close to it on both measures (0.947 and 16.3\%). The Jev API was not trained to give a uniform distribution on such items and gives a hard answer to 62.8\% of them.

\begin{table*}[t]
\centering
\small
\setlength{\tabcolsep}{5pt}
\begin{tabular}{@{}l cccc c@{}}
\toprule
 & \multicolumn{4}{c}{Hard-answer rate (\%) $\downarrow$} & Evidence \\
\cmidrule(lr){2-5}
Model & All (382) & In-dist. (200) & Near (120) & Far (62) & AUROC $\uparrow$ \\
\midrule
\mkSmol{}SmolLM2-1.7B & 36.3\sd{4.2} & 25.3\sd{2.3} & 46.7\sd{5.1} & 51.6\sd{16.4} & .765\sd{.009} \\
\mkOuro{}Ouro-1.4B, one loop & 31.9\sd{2.2} & 23.3\sd{3.4} & 37.8\sd{1.7} & 48.4\sd{5.6} & .837\sd{.007} \\
\rowcolor{sansibg}\mkSansi{}\sansi{}, loop 1 & 27.1\sd{2.1} & 22.2\sd{0.8} & 28.6\sd{6.3} & 40.3\sd{2.8} & .835\sd{.011} \\
\rowcolor{sansibg}\mkSansi{}\sansi{}, loop 4 & 18.1\sd{0.5} & 11.7\sd{0.3} & 23.1\sd{3.2} & 29.0\sd{5.8} & .928\sd{.003} \\
\rowcolor{sansibg}\mkSansi{}\sansi{}, loop 8 & 17.5\sd{1.3} & 10.7\sd{2.3} & 23.6\sd{2.4} & 27.4\sd{11.3} & .935\sd{.004} \\
\mkQwen{}Qwen3.5-4B & 18.2\sd{2.5} & 9.3\sd{1.3} & 21.9\sd{2.1} & 39.2\sd{15.2} & .948\sd{.003} \\
\mkKev{}Kev-4B (our data) & 16.3\sd{0.6} & 9.3\sd{1.2} & 20.6\sd{0.5} & 30.6\sd{4.3} & .947\sd{.002} \\
\mkJev{}Jev API & 62.8 & 77.0 & 54.2 & 33.9 & .743 \\
\bottomrule
\end{tabular}

\caption{Unanswerable items by group (number of items in parentheses) and evidence AUROC: mean $\pm$ standard deviation over three seeds. The Jev API (a single run) was not trained to give a uniform distribution on such items.}
\label{tab:unans}
\end{table*}

\section{Depth-Controlled Tasks: Additional Results}
\label{app:depth}

This appendix supports \S\ref{sec:depth}. It describes the two tasks and how the models are trained on them (Appendix~\ref{app:depth-tasks}), summarises the results (Appendix~\ref{app:depth-summary}), gives the accuracy at every depth (Appendix~\ref{app:depth-byk}) and the loop at which the answers settle (Appendix~\ref{app:depth-settle}), reports Qwen3.5-4B trained four times as long (Appendix~\ref{app:depth-long}), and gives the results of the Jev API on the two tasks (Appendix~\ref{app:wall}).

\subsection{Tasks, training and checks}
\label{app:depth-tasks}

In a liar chain, answering requires following the chain from a person whose honesty is given, keeping the verdict at every link that calls the next person honest and flipping it at every link that calls them a liar. The labels are balanced and computed by the generator, the sentences are shuffled, and every item contains a second, irrelevant chain of the same length. In object swaps, every item contains $k$ swaps of the queried object and $k$ further swaps that do not involve it. Both tasks draw their names from the same list of 80 first names and use several wordings for every kind of sentence. No test item occurs in the training set. Table~\ref{tab:depth-examples} shows one test item of each task at depths 1, 2, 4 and 8.

\begin{table*}[t]
\centering
\footnotesize
\setlength{\tabcolsep}{5pt}
\begin{tabular}{@{}c>{\raggedright\arraybackslash}p{0.80\textwidth}l@{}}
\toprule
$k$ & Item & Answer \\
\midrule
\rowcolor{groupbg}\multicolumn{3}{@{}l}{\emph{Liar chains (options: yes, no)}} \\
1 & Lee is honest. Wes is honest. According to Flo, Lee tells the truth. According to Abe, Wes lies. \emph{Does Abe tell the truth?} & no \\
\addlinespace
2 & Uma always tells the truth. Seth always tells the truth. Fred says that Seth lies. Kurt says Fred is a liar. Bert says that Uma lies. Ben says Bert is a liar. \emph{Is Ben telling the truth?} & yes \\
\addlinespace
4 & Omar is honest. Ege always lies. According to Bert, Omar tells the truth. Yves says that Eli lies. Ben says that Bert lies. Liv says that Nate lies. Eli says that Liv tells the truth. Jon says Vera is honest. Vera says that Ben lies. According to Nate, Ege tells the truth. \emph{Is Yves telling the truth?} & no \\
\addlinespace
8 & Omar is a liar. Seth is a liar. Vera says Iris is a liar. Bea says Flo is a liar. According to Eli, Wade tells the truth. According to Yul, Ida tells the truth. Nia says that Nate lies. According to Bert, Omar lies. Ida says that Nia lies. Iris says Bert is honest. According to Wade, Meg lies. Nate says Vera is honest. According to Meg, Bo tells the truth. Ben says that Seth tells the truth. According to Kim, Bea tells the truth. Flo says Ben is a liar. According to Ola, Yul lies. Bo says that Kim lies. \emph{Is Eli telling the truth?} & no \\
\midrule
\rowcolor{groupbg}\multicolumn{3}{@{}l}{\emph{Object swaps (options: the five people)}} \\
1 & Dan has the cup, Ana has the scarf, Gail has the umbrella, Lee has the coin, and Jill has the pen. Then Jill and Gail swap. Then Gail swaps with Ana. \emph{Who has the scarf at the end?} & Gail \\
\addlinespace
2 & Dov holds the coin, Eli holds the hat, Bo holds the scarf, Kim holds the key, and Jill holds the cup. Then Dov swaps with Kim. Then Eli and Jill swap. Then Jill and Kim swap. Then Eli swaps with Dov. \emph{Who has the cup at the end?} & Dov \\
\addlinespace
4 & Sam holds the scarf, Max holds the coin, Lee holds the cup, Iris holds the hat, and Dan holds the key. Then Lee and Sam trade. Then Iris swaps with Max. Then Lee swaps with Dan. Then Max swaps with Dan. Then Sam and Iris trade. Then Iris and Max trade. Then Sam and Max swap. Then Dan and Max swap. \emph{At the end, who holds the cup?} & Sam \\
\addlinespace
8 & Tara holds the box, Cleo holds the coin, Bo holds the scarf, Eli holds the ball, and Ned holds the hat. Then Ned swaps with Bo. Then Cleo and Bo swap. Then Eli swaps with Tara. Then Ned and Cleo swap. Then Cleo and Ned trade. Then Eli swaps with Bo. Then Cleo and Bo swap. Then Tara and Cleo trade. Then Eli swaps with Tara. Then Ned swaps with Tara. Then Ned and Tara swap. Then Ned and Bo swap. Then Cleo and Eli swap. Then Ned and Cleo trade. Then Cleo swaps with Eli. Then Bo and Ned trade. \emph{Who has the hat at the end?} & Eli \\
\bottomrule
\end{tabular}

\caption{One test item of each depth-controlled task at depths $k = 1, 2, 4, 8$ (the item of median length at that depth). A liar chain contains a second chain of the same length that does not matter for the question; an object-swap item contains $k$ swaps of the queried object and $k$ further swaps.}
\label{tab:depth-examples}
\end{table*}

\paragraph{Training.} The models for the liar chains are trained for 2,000 steps on 17,336 program-generated items of depths $k = 1, \dots, 8$, 4,480 of them liar chains. The models for the object swaps are trained with the same recipe, also for 2,000 steps, on 4,480 items of depths $k = 1, \dots, 8$. \sansi{} is trained with eight loops and run for up to 16 loops at test time; Qwen3.5-4B makes a single pass. Each test set has 120 items at every depth $k = 1, \dots, 16$ (1,920 items per task), and all results are means of three seeds.

\paragraph{Checks against shortcuts.} Surface heuristics and a bag-of-words classifier stay at chance on both tasks (48--51\% on liar chains, 19.5--21.4\% on object swaps), so an item cannot be answered without following its chain.

\subsection{Results in summary}
\label{app:depth-summary}

Table~\ref{tab:depth-summary} gives, for both tasks, the depth that a model holds, its accuracy on the trained depths ($k \le 8$) and on the unseen depths ($k > 8$), and the difference between \sansi{} and Qwen3.5-4B. A model holds depth $k$ if its accuracy is at least 75\% at every depth up to $k$. The accuracy on a range of depths is the mean over its eight depths. For a difference, the correctness of every item is first averaged over the three seeds of each model, and the items are then resampled 2,000 times for the 95\% interval.

\begin{table*}[t]
\centering
\small
\setlength{\tabcolsep}{7pt}
\begin{tabular}{@{}l c cc cc@{}}
\toprule
 & Holds & \multicolumn{2}{c}{Accuracy (\%)} & \multicolumn{2}{c}{Difference to Qwen3.5-4B (points)} \\
\cmidrule(lr){3-4}\cmidrule(l){5-6}
Model & depth $k$ & $k \le 8$ & $k > 8$ & $k \le 8$ & $k > 8$ \\
\midrule
\rowcolor{groupbg}\multicolumn{6}{@{}l}{\emph{Liar chains (chance 50\%)}} \\
\sansi{}, loop 1 & 3 & 69.0 & 49.9 & \dworse{-5.3}{-7.3}{-3.4} & \dns{-0.2}{-1.6}{+1.2} \\
\sansi{}, loop 2 & 6 & 88.0 & 51.6 & \dbetter{+13.7}{+11.7}{+15.6} & \dns{+1.6}{-1.1}{+4.3} \\
\sansi{}, loop 4 & 11 & 96.6 & 70.9 & \dbetter{+22.3}{+20.5}{+24.1} & \dbetter{+20.8}{+18.9}{+22.9} \\
\sansi{}, loop 8 & 11 & 97.0 & 72.6 & \dbetter{+22.7}{+20.8}{+24.7} & \dbetter{+22.6}{+20.5}{+24.7} \\
\sansi{}, loop 16 & 11 & 96.9 & 72.7 & -- & -- \\
Qwen3.5-4B & 3 & 74.3 & 50.0 & -- & -- \\
\midrule
\rowcolor{groupbg}\multicolumn{6}{@{}l}{\emph{Object swaps (chance 20\%)}} \\
\sansi{}, loop 1 & 2 & 57.8 & 32.3 & \dworse{-10.1}{-12.3}{-7.9} & \dns{-1.9}{-4.3}{+0.5} \\
\sansi{}, loop 2 & 3 & 74.2 & 40.1 & \dbetter{+6.2}{+4.1}{+8.4} & \dbetter{+5.9}{+3.3}{+8.3} \\
\sansi{}, loop 4 & 7 & 86.8 & 57.3 & \dbetter{+18.9}{+16.5}{+21.2} & \dbetter{+23.1}{+20.3}{+25.9} \\
\sansi{}, loop 8 & 9 & 88.2 & 63.7 & \dbetter{+20.2}{+17.9}{+22.6} & \dbetter{+29.5}{+26.5}{+32.7} \\
\sansi{}, loop 16 & 7 & 87.6 & 64.0 & -- & -- \\
Qwen3.5-4B & 3 & 68.0 & 34.2 & -- & -- \\
\bottomrule
\end{tabular}

\caption{The two depth-controlled tasks in summary (means of three seeds; 120 test items per depth). Holds depth $k$: the accuracy is at least 75\% at every depth up to $k$. $k \le 8$: depths seen in training; $k > 8$: unseen depths. Differences with 95\% bootstrap intervals; loop 16 is beyond the trained loops. Differences: \textcolor{better}{\textbf{teal}}, \sansi{} is significantly better (the interval excludes 0); \textcolor{worse}{\textbf{red}}, significantly worse; \textcolor{black!55}{grey}, the interval includes 0.}
\label{tab:depth-summary}
\end{table*}

Read after one loop, \sansi{} is below Qwen3.5-4B on the trained depths of both tasks ($-$5.3 points on liar chains and $-$10.1 points on object swaps). From the second loop on it is above. Running 16 loops, twice the number of trained loops, gives the accuracy of eight loops on both tasks (72.7\% and 64.0\% on the unseen depths), so loops beyond the trained ones do not extend the depth. Figure~\ref{fig:nloops-grid} in Appendix~\ref{app:nloops} shows every loop up to the sixteenth.

\subsection{Accuracy at every depth}
\label{app:depth-byk}

Table~\ref{tab:depth-byk} gives the numbers behind Figure~\ref{fig:depth}: the accuracy at every depth for \sansi{} read after 1, 2, 4 and 8 loops and for Qwen3.5-4B.

\begin{table*}[t]
\centering
\footnotesize
\setlength{\tabcolsep}{3.2pt}
\begin{tabular}{@{}l rrrrrrrr @{\hspace{10pt}} rrrrrrrr@{}}
\toprule
 & \multicolumn{8}{c}{Depths seen in training} & \multicolumn{8}{c}{Unseen depths} \\
\cmidrule(lr){2-9}\cmidrule(l){10-17}
$k$ & 1 & 2 & 3 & 4 & 5 & 6 & 7 & 8 & 9 & 10 & 11 & 12 & 13 & 14 & 15 & 16 \\
\midrule
\rowcolor{groupbg}\multicolumn{17}{@{}l}{\emph{Liar chains (chance 50\%)}} \\
\sansi{}, loop 1 & \cellcolor{sansi!50}100.0 & \cellcolor{sansi!49}98.6 & \cellcolor{sansi!29}79.2 & \cellcolor{sansi!10}60.0 & \cellcolor{sansi!9}59.4 & \cellcolor{sansi!3}52.8 & \cellcolor{sansi!1}50.6 & \cellcolor{sansi!1}51.4 & 48.6 & \cellcolor{sansi!1}50.6 & \cellcolor{sansi!2}51.7 & \cellcolor{sansi!1}50.6 & \cellcolor{sansi!1}50.8 & 48.3 & 49.2 & 49.2 \\
 & \sd{0.0} & \sd{1.3} & \sd{9.8} & \sd{11.0} & \sd{3.4} & \sd{8.0} & \sd{7.1} & \sd{1.7} & \sd{3.8} & \sd{1.7} & \sd{4.4} & \sd{1.3} & \sd{2.5} & \sd{1.4} & \sd{2.2} & \sd{1.4} \\[1.5pt]
\sansi{}, loop 2 & \cellcolor{sansi!50}100.0 & \cellcolor{sansi!50}100.0 & \cellcolor{sansi!50}99.7 & \cellcolor{sansi!47}96.7 & \cellcolor{sansi!41}91.4 & \cellcolor{sansi!31}80.8 & \cellcolor{sansi!20}69.7 & \cellcolor{sansi!16}65.6 & \cellcolor{sansi!6}55.8 & \cellcolor{sansi!2}51.7 & \cellcolor{sansi!1}51.1 & \cellcolor{sansi!3}52.5 & \cellcolor{sansi!3}52.5 & \cellcolor{sansi!1}50.6 & \cellcolor{sansi!1}51.1 & 47.5 \\
 & \sd{0.0} & \sd{0.0} & \sd{0.5} & \sd{2.2} & \sd{2.7} & \sd{6.0} & \sd{7.7} & \sd{5.1} & \sd{3.6} & \sd{3.6} & \sd{4.3} & \sd{0.8} & \sd{3.8} & \sd{5.7} & \sd{1.9} & \sd{3.0} \\[1.5pt]
\sansi{}, loop 4 & \cellcolor{sansi!50}99.7 & \cellcolor{sansi!50}100.0 & \cellcolor{sansi!50}99.7 & \cellcolor{sansi!50}100.0 & \cellcolor{sansi!49}98.6 & \cellcolor{sansi!46}95.8 & \cellcolor{sansi!41}91.1 & \cellcolor{sansi!38}88.1 & \cellcolor{sansi!37}86.9 & \cellcolor{sansi!34}83.9 & \cellcolor{sansi!29}79.4 & \cellcolor{sansi!21}71.4 & \cellcolor{sansi!17}66.7 & \cellcolor{sansi!14}63.6 & \cellcolor{sansi!7}56.9 & \cellcolor{sansi!8}58.1 \\
 & \sd{0.5} & \sd{0.0} & \sd{0.5} & \sd{0.0} & \sd{1.7} & \sd{3.6} & \sd{1.7} & \sd{5.1} & \sd{4.6} & \sd{7.3} & \sd{6.4} & \sd{6.1} & \sd{8.5} & \sd{2.7} & \sd{8.9} & \sd{1.9} \\[1.5pt]
\sansi{}, loop 8 & \cellcolor{sansi!50}99.7 & \cellcolor{sansi!50}100.0 & \cellcolor{sansi!50}99.7 & \cellcolor{sansi!50}100.0 & \cellcolor{sansi!49}98.9 & \cellcolor{sansi!45}95.3 & \cellcolor{sansi!42}92.2 & \cellcolor{sansi!41}90.6 & \cellcolor{sansi!37}87.2 & \cellcolor{sansi!36}86.1 & \cellcolor{sansi!30}80.3 & \cellcolor{sansi!24}73.9 & \cellcolor{sansi!16}66.1 & \cellcolor{sansi!16}65.8 & \cellcolor{sansi!11}61.4 & \cellcolor{sansi!10}60.3 \\
 & \sd{0.5} & \sd{0.0} & \sd{0.5} & \sd{0.0} & \sd{1.9} & \sd{5.5} & \sd{1.7} & \sd{1.3} & \sd{4.8} & \sd{7.5} & \sd{5.7} & \sd{9.7} & \sd{10.1} & \sd{11.0} & \sd{10.1} & \sd{3.9} \\[1.5pt]
Qwen3.5-4B & \cellcolor{qwen!50}100.0 & \cellcolor{qwen!49}98.6 & \cellcolor{qwen!39}89.2 & \cellcolor{qwen!25}74.7 & \cellcolor{qwen!19}68.9 & \cellcolor{qwen!9}59.4 & \cellcolor{qwen!2}51.7 & \cellcolor{qwen!2}51.9 & 48.9 & 50.3 & 50.0 & \cellcolor{qwen!1}51.4 & \cellcolor{qwen!1}50.6 & 50.3 & 49.4 & 49.4 \\
{\scriptsize\quad 2,000 steps} & \sd{0.0} & \sd{1.7} & \sd{7.3} & \sd{11.3} & \sd{9.6} & \sd{10.1} & \sd{9.5} & \sd{5.4} & \sd{3.8} & \sd{2.1} & \sd{0.8} & \sd{1.0} & \sd{1.0} & \sd{2.1} & \sd{1.0} & \sd{1.0} \\[1.5pt]
Qwen3.5-4B & \cellcolor{qwen!50}100.0 & \cellcolor{qwen!49}98.6 & \cellcolor{qwen!44}93.9 & \cellcolor{qwen!39}89.2 & \cellcolor{qwen!31}80.8 & \cellcolor{qwen!16}65.8 & \cellcolor{qwen!11}60.8 & \cellcolor{qwen!5}54.7 & \cellcolor{qwen!4}54.2 & 48.9 & 48.6 & \cellcolor{qwen!4}54.2 & \cellcolor{qwen!2}51.9 & 48.6 & \cellcolor{qwen!1}51.4 & \cellcolor{qwen!4}53.9 \\
{\scriptsize\quad 8,000 steps} & \sd{0.0} & \sd{1.3} & \sd{7.7} & \sd{12.3} & \sd{9.6} & \sd{16.1} & \sd{6.3} & \sd{2.1} & \sd{3.0} & \sd{6.9} & \sd{5.5} & \sd{5.2} & \sd{5.4} & \sd{2.4} & \sd{1.3} & \sd{5.1} \\[1.5pt]
\midrule
\rowcolor{groupbg}\multicolumn{17}{@{}l}{\emph{Object swaps (chance 20\%)}} \\
\sansi{}, loop 1 & \cellcolor{sansi!49}98.3 & \cellcolor{sansi!48}96.7 & \cellcolor{sansi!31}68.9 & \cellcolor{sansi!16}46.4 & \cellcolor{sansi!18}49.2 & \cellcolor{sansi!10}36.1 & \cellcolor{sansi!10}36.4 & \cellcolor{sansi!7}30.8 & \cellcolor{sansi!9}34.2 & \cellcolor{sansi!7}30.8 & \cellcolor{sansi!7}31.1 & \cellcolor{sansi!10}36.7 & \cellcolor{sansi!6}29.2 & \cellcolor{sansi!9}35.0 & \cellcolor{sansi!5}28.6 & \cellcolor{sansi!8}32.5 \\
 & \sd{0.8} & \sd{0.8} & \sd{9.3} & \sd{4.9} & \sd{2.9} & \sd{5.0} & \sd{3.5} & \sd{4.6} & \sd{2.9} & \sd{1.4} & \sd{3.4} & \sd{2.9} & \sd{2.2} & \sd{3.3} & \sd{2.9} & \sd{2.2} \\[1.5pt]
\sansi{}, loop 2 & \cellcolor{sansi!50}99.7 & \cellcolor{sansi!49}98.9 & \cellcolor{sansi!48}96.1 & \cellcolor{sansi!33}72.8 & \cellcolor{sansi!27}62.5 & \cellcolor{sansi!28}65.3 & \cellcolor{sansi!22}54.4 & \cellcolor{sansi!15}43.6 & \cellcolor{sansi!16}45.3 & \cellcolor{sansi!13}41.1 & \cellcolor{sansi!12}39.7 & \cellcolor{sansi!13}40.3 & \cellcolor{sansi!11}37.5 & \cellcolor{sansi!15}44.4 & \cellcolor{sansi!8}33.1 & \cellcolor{sansi!12}39.4 \\
 & \sd{0.5} & \sd{1.0} & \sd{3.9} & \sd{4.6} & \sd{1.7} & \sd{3.4} & \sd{2.4} & \sd{1.0} & \sd{3.9} & \sd{1.0} & \sd{1.7} & \sd{2.1} & \sd{0.0} & \sd{1.7} & \sd{2.5} & \sd{2.4} \\[1.5pt]
\sansi{}, loop 4 & \cellcolor{sansi!50}99.7 & \cellcolor{sansi!50}99.4 & \cellcolor{sansi!48}97.2 & \cellcolor{sansi!43}88.6 & \cellcolor{sansi!37}79.4 & \cellcolor{sansi!40}84.2 & \cellcolor{sansi!35}75.8 & \cellcolor{sansi!31}70.0 & \cellcolor{sansi!34}74.4 & \cellcolor{sansi!28}65.0 & \cellcolor{sansi!27}63.6 & \cellcolor{sansi!23}57.5 & \cellcolor{sansi!22}55.0 & \cellcolor{sansi!19}49.7 & \cellcolor{sansi!17}47.8 & \cellcolor{sansi!16}45.3 \\
 & \sd{0.5} & \sd{1.0} & \sd{3.5} & \sd{8.9} & \sd{4.7} & \sd{3.8} & \sd{0.8} & \sd{2.2} & \sd{4.1} & \sd{2.5} & \sd{2.5} & \sd{2.5} & \sd{1.4} & \sd{2.1} & \sd{2.4} & \sd{4.8} \\[1.5pt]
\sansi{}, loop 8 & \cellcolor{sansi!50}99.4 & \cellcolor{sansi!50}99.4 & \cellcolor{sansi!48}97.2 & \cellcolor{sansi!43}88.6 & \cellcolor{sansi!37}79.7 & \cellcolor{sansi!40}83.6 & \cellcolor{sansi!39}81.9 & \cellcolor{sansi!35}75.3 & \cellcolor{sansi!36}76.9 & \cellcolor{sansi!34}74.7 & \cellcolor{sansi!31}70.3 & \cellcolor{sansi!31}69.7 & \cellcolor{sansi!26}60.8 & \cellcolor{sansi!22}55.6 & \cellcolor{sansi!19}50.8 & \cellcolor{sansi!19}50.8 \\
 & \sd{0.5} & \sd{1.0} & \sd{3.5} & \sd{8.9} & \sd{5.9} & \sd{5.4} & \sd{0.5} & \sd{4.2} & \sd{3.4} & \sd{5.4} & \sd{6.3} & \sd{6.7} & \sd{6.3} & \sd{1.7} & \sd{2.2} & \sd{3.8} \\[1.5pt]
Qwen3.5-4B & \cellcolor{qwen!50}100.0 & \cellcolor{qwen!50}99.7 & \cellcolor{qwen!46}93.3 & \cellcolor{qwen!28}65.0 & \cellcolor{qwen!19}50.3 & \cellcolor{qwen!19}49.7 & \cellcolor{qwen!17}46.9 & \cellcolor{qwen!12}38.6 & \cellcolor{qwen!9}34.7 & \cellcolor{qwen!7}31.1 & \cellcolor{qwen!8}32.5 & \cellcolor{qwen!12}39.2 & \cellcolor{qwen!9}33.6 & \cellcolor{qwen!12}39.2 & \cellcolor{qwen!7}31.7 & \cellcolor{qwen!7}31.7 \\
 & \sd{0.0} & \sd{0.5} & \sd{6.6} & \sd{18.4} & \sd{8.4} & \sd{3.4} & \sd{3.2} & \sd{2.5} & \sd{4.1} & \sd{3.2} & \sd{2.2} & \sd{2.2} & \sd{1.7} & \sd{1.7} & \sd{3.3} & \sd{2.2} \\[1.5pt]
\bottomrule
\end{tabular}

\caption{Accuracy (\%) at every depth $k$ on the two depth-controlled tasks (120 test items per depth; means of three seeds, with the standard deviation over the seeds on the line below). Qwen3.5-4B makes a single pass; on liar chains it is also shown after 8,000 training steps, four times as long as the other models (Appendix~\ref{app:depth-long}). Cells are shaded by the accuracy above chance (blue: \sansi{}; green: Qwen3.5-4B).}
\label{tab:depth-byk}
\end{table*}

\paragraph{Liar chains.} Read after one loop, \sansi{} answers the shortest chains (100.0\% at $k = 1$, 98.6\% at $k = 2$) and is within three points of chance from $k = 6$. Qwen3.5-4B also answers the shortest chains (100.0\% and 98.6\%) and is within two points of chance from $k = 7$. Loops five to eight add little: the accuracy on the unseen depths rises from 70.9\% at loop 4 to 72.6\% at loop 8. The three seeds of \sansi{} agree on the trained depths (95.7--98.1\% at loop 8) and differ on the unseen ones (64.7--79.4\%); those of Qwen3.5-4B reach 69.7--81.9\% on the trained depths. Confidence does not follow accuracy down. At $k = 16$, \sansi{} is right on 60.3\% of the items with a mean confidence of 0.73. Qwen3.5-4B, in contrast, has a low confidence where it is at chance (0.54 on average for $k \ge 9$).

\paragraph{Object swaps.} After one loop, \sansi{} follows the object through two swaps (96.7\% at $k = 2$, 68.9\% at $k = 3$). Unlike on liar chains, loops five to eight still help on the unseen depths (57.3\% after four loops, 63.7\% after eight). The lowest values among the seeds of \sansi{} at loop 8 (84.3\% on the trained and 60.2\% on the unseen depths) are above the highest among the seeds of Qwen3.5-4B (72.1\% and 34.9\%). Two limits remain. The trained depths are not learned completely (75--84\% for $k = 5, \dots, 8$ at loop 8). And the wrong answers at the unseen depths come with a mean confidence of 0.87--0.91 across the seeds, the same as for Qwen3.5-4B (0.86--0.91).

\begin{figure}[t]
\centering
\includegraphics[width=\columnwidth]{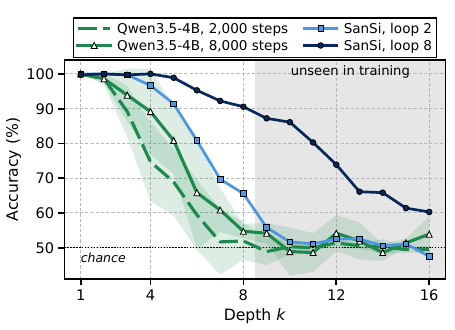}
\caption{Liar chains: accuracy at every depth $k$ (means of three seeds; 120 test items per depth) of Qwen3.5-4B after 2,000 and after 8,000 training steps, and of \sansi{} (2,000 steps) read after two and after eight loops. Bands: one standard deviation over the seeds of the two Qwen3.5-4B runs. Grey: depths not seen in training; dotted line: chance. Values in Table~\ref{tab:depth-byk}.}
\label{fig:depth-long}
\end{figure}

\begin{figure}[t]
\centering
\includegraphics[width=\columnwidth]{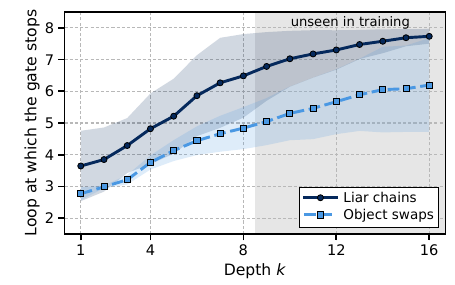}
\caption{Where Ouro's exit gate ($q = 0.5$; Appendix~\ref{app:stop}) stops on the two depth-controlled tasks, by depth $k$ (means of three seeds; bands: $\pm$1 standard deviation over the seeds). Grey: depths not seen in training. Values in Table~\ref{tab:depth-settle}.}
\label{fig:gate-depth}
\end{figure}

\subsection{When answers settle}
\label{app:depth-settle}

As on the main suite (\S\ref{sec:rq2}), the answers to deeper items settle at later loops. Table~\ref{tab:depth-settle} gives, at every depth, the mean loop at which the answer of \sansi{} settles, that is, the first loop from which it no longer changes, and the share of items whose answer at loop 8 differs from that at loop 1. On liar chains the settling loop grows from 1.0 at $k = 1$ to 2.3 at $k = 8$ and 4.2 at $k = 16$; on object swaps it grows from 1.0 to 3.0 and 4.5. From $k = 6$ on liar chains and from $k = 4$ on object swaps, the answer after eight loops differs from the answer after one loop on about half of the items or more (47--56\% and 51--70\%). Ouro's exit gate ($q = 0.5$; Appendix~\ref{app:stop}) also stops later on deeper items: on liar chains after 3.6 loops at $k = 1$, 6.5 at $k = 8$ and 7.7 at $k = 16$, on object swaps after 2.8, 4.8 and 6.2, with 6.2 and 4.8 loops on average over all depths (Figure~\ref{fig:gate-depth}).

\begin{table*}[t]
\centering
\small
\setlength{\tabcolsep}{5pt}
\begin{tabular}{@{}l rrrrrrrrrrrrrrrr@{}}
\toprule
$k$ & 1 & 2 & 3 & 4 & 5 & 6 & 7 & 8 & 9 & 10 & 11 & 12 & 13 & 14 & 15 & 16 \\
\midrule
\rowcolor{groupbg}\multicolumn{17}{@{}l}{\emph{Liar chains}} \\
Settles at loop & 1.0 & 1.0 & \cellcolor{sansi!3}1.2 & \cellcolor{sansi!6}1.5 & \cellcolor{sansi!8}1.6 & \cellcolor{sansi!12}1.9 & \cellcolor{sansi!14}2.1 & \cellcolor{sansi!17}2.3 & \cellcolor{sansi!22}2.7 & \cellcolor{sansi!24}2.9 & \cellcolor{sansi!27}3.1 & \cellcolor{sansi!31}3.4 & \cellcolor{sansi!37}3.9 & \cellcolor{sansi!39}4.0 & \cellcolor{sansi!41}4.2 & \cellcolor{sansi!41}4.2 \\
Changed, loop 1 to 8 (\%) & 0 & \cellcolor{sansi!1}1 & \cellcolor{sansi!13}21 & \cellcolor{sansi!24}40 & \cellcolor{sansi!25}41 & \cellcolor{sansi!28}47 & \cellcolor{sansi!30}50 & \cellcolor{sansi!31}52 & \cellcolor{sansi!32}54 & \cellcolor{sansi!31}52 & \cellcolor{sansi!31}52 & \cellcolor{sansi!32}54 & \cellcolor{sansi!33}55 & \cellcolor{sansi!34}56 & \cellcolor{sansi!32}53 & \cellcolor{sansi!31}52 \\
Gate stops at loop & \cellcolor{sansi!17}3.6 & \cellcolor{sansi!18}3.8 & \cellcolor{sansi!21}4.3 & \cellcolor{sansi!25}4.8 & \cellcolor{sansi!27}5.2 & \cellcolor{sansi!31}5.9 & \cellcolor{sansi!34}6.3 & \cellcolor{sansi!35}6.5 & \cellcolor{sansi!37}6.8 & \cellcolor{sansi!39}7.0 & \cellcolor{sansi!40}7.2 & \cellcolor{sansi!41}7.3 & \cellcolor{sansi!42}7.5 & \cellcolor{sansi!42}7.6 & \cellcolor{sansi!43}7.7 & \cellcolor{sansi!43}7.7 \\
\midrule
\rowcolor{groupbg}\multicolumn{17}{@{}l}{\emph{Object swaps}} \\
Settles at loop & 1.0 & \cellcolor{sansi!1}1.1 & \cellcolor{sansi!5}1.4 & \cellcolor{sansi!12}1.9 & \cellcolor{sansi!13}2.0 & \cellcolor{sansi!17}2.3 & \cellcolor{sansi!22}2.7 & \cellcolor{sansi!26}3.0 & \cellcolor{sansi!26}3.0 & \cellcolor{sansi!30}3.3 & \cellcolor{sansi!32}3.5 & \cellcolor{sansi!33}3.6 & \cellcolor{sansi!39}4.0 & \cellcolor{sansi!40}4.1 & \cellcolor{sansi!39}4.0 & \cellcolor{sansi!45}4.5 \\
Changed, loop 1 to 8 (\%) & \cellcolor{sansi!1}1 & \cellcolor{sansi!2}3 & \cellcolor{sansi!19}32 & \cellcolor{sansi!32}53 & \cellcolor{sansi!31}51 & \cellcolor{sansi!39}65 & \cellcolor{sansi!35}59 & \cellcolor{sansi!42}70 & \cellcolor{sansi!38}64 & \cellcolor{sansi!41}68 & \cellcolor{sansi!40}66 & \cellcolor{sansi!40}66 & \cellcolor{sansi!41}68 & \cellcolor{sansi!38}64 & \cellcolor{sansi!38}64 & \cellcolor{sansi!38}64 \\
Gate stops at loop & \cellcolor{sansi!11}2.8 & \cellcolor{sansi!13}3.0 & \cellcolor{sansi!14}3.2 & \cellcolor{sansi!18}3.8 & \cellcolor{sansi!20}4.1 & \cellcolor{sansi!22}4.4 & \cellcolor{sansi!24}4.7 & \cellcolor{sansi!25}4.8 & \cellcolor{sansi!26}5.0 & \cellcolor{sansi!28}5.3 & \cellcolor{sansi!29}5.5 & \cellcolor{sansi!30}5.7 & \cellcolor{sansi!31}5.9 & \cellcolor{sansi!32}6.1 & \cellcolor{sansi!33}6.1 & \cellcolor{sansi!33}6.2 \\
\bottomrule
\end{tabular}

\caption{\sansi{} on the depth-controlled tasks, by depth $k$: mean loop at which the answer settles, share of items whose answer at loop 8 differs from that at loop 1, and mean loop at which Ouro's exit gate ($q = 0.5$; Appendix~\ref{app:stop}) stops (means of three seeds). Darker cells: later settling, more changed answers and later stopping.}
\label{tab:depth-settle}
\end{table*}

\subsection{Longer training of Qwen3.5-4B}
\label{app:depth-long}

A single-pass model might fail on deeper chains only because 2,000 steps are too few for it. We therefore train Qwen3.5-4B on the liar-chain training set for 8,000 steps, four times as long, with the same recipe and three seeds; the learning-rate schedule is stretched over the 8,000 steps. On one RTX A6000, a run takes 195--198 GPU minutes, about the training time of \sansi{} with its 2,000 steps (193--238 GPU minutes on two cards), against 50--51 for Qwen3.5-4B with 2,000 steps. Table~\ref{tab:depth-long} gives the results after every 2,000 steps, and Figure~\ref{fig:depth-long} the accuracy at every depth.

\begin{table*}[t]
\centering
\small
\setlength{\tabcolsep}{3.5pt}
\begin{tabular}{@{}l c cc c@{}}
\toprule
Model & \begin{tabular}[b]{@{}c@{}}Training\\(GPU min)\end{tabular} & $k \le 8$ & $k > 8$ & Holds depth \\
\midrule
Qwen3.5-4B, 2,000 steps (\S\ref{sec:depth}) & 50--51 & 74.3\sd{6.6} & 50.0\sd{0.4} & 3 \\
\rowcolor{groupbg}\multicolumn{5}{@{}l}{\emph{Qwen3.5-4B trained for 8,000 steps, after step}} \\
2,000 & & 58.0\sd{7.8} & 49.9\sd{0.1} & 1 \\
4,000 & & 64.7\sd{12.8} & 50.1\sd{0.7} & 2 \\
6,000 & & 70.3\sd{17.4} & 50.4\sd{0.6} & 4 \\
8,000 & 195--198 & 80.5\sd{6.6} & 51.5\sd{1.5} & 5 \\
\midrule
\rowcolor{sansibg}\sansi{}, 2,000 steps, read at loop 2 & 193--238 & 88.0\sd{2.5} & 51.6\sd{0.9} & 6 \\
\rowcolor{sansibg}\sansi{}, 2,000 steps, read at loop 8 & 193--238 & 97.0\sd{1.2} & 72.6\sd{7.4} & 11 \\
\rowcolor{groupbg}\multicolumn{5}{@{}l}{\emph{Differences}} \\
Qwen3.5-4B, 8,000 $-$ 2,000 steps & & \dbetter{+6.2}{+4.4}{+7.9} & \dns{+1.4}{-0.4}{+3.2} & \\
\sansi{} at loop 2 $-$ Qwen3.5-4B, 8,000 steps & & \dbetter{+7.5}{+5.9}{+9.2} & \dns{+0.1}{-2.9}{+2.9} & \\
\sansi{} at loop 8 $-$ Qwen3.5-4B, 8,000 steps & & \dbetter{+16.6}{+14.8}{+18.3} & \dbetter{+21.2}{+18.6}{+23.8} & \\
\bottomrule
\end{tabular}

\caption{Qwen3.5-4B trained four times as long on the liar chains (mean $\pm$ standard deviation over three seeds; 120 test items per depth). Training: GPU minutes of one run on RTX A6000 cards; seed 2 of the 8,000-step runs ran on four RTX A5500 with gradient checkpointing and is not timed. $k \le 8$: depths seen in training; $k > 8$: unseen depths. Holds depth $k$: the mean accuracy over the seeds is at least 75\% at every depth up to $k$. Differences: the correctness of every item is averaged over the seeds and the items are resampled 2,000 times for the 95\% interval; \textcolor{better}{\textbf{teal}}, the first term is significantly higher; \textcolor{black!55}{grey}, the interval includes 0. The large standard deviations up to step 6,000 come from seed 0, which stays at chance on the liar chains until then.}
\label{tab:depth-long}
\end{table*}

Longer training helps on the trained depths but not beyond them. After 8,000 steps, Qwen3.5-4B holds depth 5 instead of 3 and reaches 80.5\% on the trained depths, 6.2 points [4.4, 7.9] more than after 2,000 steps; on the unseen depths it stays at chance (51.5\%; +1.4 [$-$0.4, 3.2]). \sansi{}, trained for 2,000 steps, still leads by 16.6 points [14.8, 18.3] on the trained depths and by 21.2 [18.6, 23.8] on the unseen ones. On the main suite, two loops cost less GPU time than one pass of Qwen3.5-4B (Table~\ref{tab:params}). On the liar chains, \sansi{} read after two loops is 7.5 points [5.9, 9.2] ahead of the 8,000-step Qwen3.5-4B on the trained depths and level on the unseen ones (+0.1 [$-$2.9, 2.9]).

One seed of Qwen3.5-4B is slow to learn the task. Until step 4,000 it gives option A to every liar-chain test item, with a mean confidence of 0.53--0.61, although it has learned other items of its training set; it learns the liar chains only between steps 6,000 and 8,000 (72.9\% on the trained depths, against 84.9\% and 83.6\% for the other two seeds). All three seeds of \sansi{} learn the trained depths within 2,000 steps (95.7--98.1\%).

\subsection{The Jev API on the two tasks}
\label{app:wall}

We queried the Jev API (jev-1.13.0) on the test items of the two depth-controlled tasks (1,920 items per task), three times per item. The API was not trained on these tasks, so its numbers are not comparable with those of the fine-tuned models in \S\ref{sec:depth}. They show that the Jev API, used as it is offered, follows only a few dependent steps. Table~\ref{tab:wall} gives its accuracy and its mean confidence at every depth.

On liar chains (two options) the API answers every chain of depth 1 and 86.4\% of the chains of depth 2. Its accuracy falls to 62.5\% at depth 3 and stays between 42.5\% and 53.3\% from depth 6 on, around the chance level of 50\%. On object swaps (five options) it follows one swap in 77.5\% of the items and two swaps in 42.5\%; from depth 3 on its accuracy is between 18.6\% and 31.1\%, close to the chance level of 20\%. On both tasks the confidence of the API falls with the depth (from 0.99 to 0.60 on liar chains and from 0.92 to 0.29 on object swaps) and, from depth 3 on, is above its accuracy at every depth but one (depth 14 of object swaps): over the depths 6 to 16 it is 0.61 on liar chains, where 47.6\% of the answers are right, and 0.33 on object swaps, where 23.6\% are right.

\begin{table*}[t]
\centering
\footnotesize
\setlength{\tabcolsep}{4.2pt}
\begin{tabular}{@{}l rrrrrrrrrrrrrrrr@{}}
\toprule
$k$ & 1 & 2 & 3 & 4 & 5 & 6 & 7 & 8 & 9 & 10 & 11 & 12 & 13 & 14 & 15 & 16 \\
\midrule
\rowcolor{groupbg}\multicolumn{17}{@{}l}{\emph{Liar chains (chance 50\%)}} \\
Accuracy (\%) & \cellcolor{jev!45}100.0 & \cellcolor{jev!33}86.4 & \cellcolor{jev!11}62.5 & \cellcolor{jev!6}56.9 & \cellcolor{jev!8}58.6 & 44.7 & 48.6 & 47.8 & 44.7 & 48.3 & \cellcolor{jev!1}51.1 & \cellcolor{jev!3}53.3 & 47.5 & 43.9 & 42.5 & \cellcolor{jev!1}51.1 \\
Mean confidence & .99 & .82 & .69 & .66 & .63 & .63 & .61 & .62 & .62 & .60 & .61 & .61 & .60 & .59 & .60 & .60 \\
\midrule
\rowcolor{groupbg}\multicolumn{17}{@{}l}{\emph{Object swaps (chance 20\%)}} \\
Accuracy (\%) & \cellcolor{jev!32}77.5 & \cellcolor{jev!13}42.5 & \cellcolor{jev!2}23.6 & 20.6 & \cellcolor{jev!2}24.2 & \cellcolor{jev!3}25.0 & 19.4 & \cellcolor{jev!1}21.4 & 18.6 & 20.3 & \cellcolor{jev!4}26.9 & \cellcolor{jev!2}24.2 & \cellcolor{jev!1}21.4 & \cellcolor{jev!6}31.1 & \cellcolor{jev!2}23.3 & \cellcolor{jev!4}27.8 \\
Mean confidence & .92 & .69 & .53 & .47 & .42 & .39 & .37 & .36 & .32 & .33 & .31 & .31 & .31 & .29 & .29 & .29 \\
\bottomrule
\end{tabular}

\caption{The Jev API on the two depth-controlled tasks: accuracy and mean confidence at every depth $k$ (1,920 items per task, three calls per item). Cells are shaded by the accuracy above chance.}
\label{tab:wall}
\end{table*}

\section{Ablations: Details}
\label{app:ablations}

This appendix supports \S\ref{sec:ablations} and follows its order: how the loops are trained (Appendices~\ref{app:train-loops} and~\ref{app:rl}), the number of loops (Appendix~\ref{app:nloops}), the backbone (Appendix~\ref{app:backbone}) and the averaging of the loops (Appendix~\ref{app:averaging}). Most intervals of accuracy differences in this appendix are listed in Tables~\ref{tab:ci} (Appendix~\ref{app:main-ci}) and~\ref{tab:nloops-ci}; the others are given only in the text.

\subsection{Which loops carry the loss, and the Brier term}
\label{app:train-loops}

\sansi{} puts the loss on every loop. With the loss on the last loop only, the model reaches 70.9\% at loop 8 (Table~\ref{tab:ablation} in \S\ref{sec:ablations}), 1.1 points below \sansi{} [0.7, 1.5], and its earlier loops are much weaker: 35.6\% at loop 1, 51.9\% at loop 2 and 67.9\% at loop 4 (3.7 points below \sansi{} [3.2, 4.2]). Training every loop therefore buys mainly a model that can be read at any budget, and about one point at the last loop. With the loss on loops 1, 2, 4 and 8 only, the model reaches 71.2\% at loop 8, 0.7 points below \sansi{} [0.4, 1.1], and the loops without a loss do not collapse: loop 3, read with the frozen head, is within 0.2 points of \sansi{} [$-$0.6, 0.2]. Sparser supervision thus costs little, but it buys nothing either.

\paragraph{Seeds and calibration.} The model trained with the loss on the last loop only is 0.8, 1.7 and 0.7 points below \sansi{} at loop 8 in the three seeds; it changes its answer on 58.4\% of the items between the first two loops, and its ECE at loop 8 is 0.073 against 0.093 (0.100 against 0.149 on near transfer). The model trained with the loss on loops 1, 2, 4 and 8 differs from \sansi{} by +0.1, $-$1.5 and $-$0.9 points in the three seeds; its deficit lies on near transfer (2.3 points [1.2, 3.4]), and its ECE is 0.098 against 0.093.

\paragraph{Cross-entropy without the Brier term.} This variant is trained with the cross-entropy term of Equation~\ref{eq:loss} alone; everything else is as in \sansi{}, including the three seeds. Its accuracy is close to that of \sansi{} at every loop: 58.8\% at loop 1 and 71.6\% at loop 8, against 58.4\% and 72.0\% ($-$0.4 points [$-$0.8, $-$0.0] at loop 8; +0.5, $-$0.9 and $-$0.8 in the three seeds). Its probabilities are worse. The ECE is higher at loop 3 (0.089 against 0.082; +0.007 [0.003, 0.012]) and at loop 8 (0.104 against 0.093; +0.011 [0.007, 0.016]; 0.112, 0.101 and 0.099 in the three seeds, against 0.099, 0.100 and 0.079 for \sansi{}), and on the 382 unanswerable items the share of hard answers is 20.0\% against 17.5\% (+2.5 points [0.7, 4.3]). The evidence AUROC and the AUROC of right against wrong answers do not differ ($-$0.003 [$-$0.009, 0.003] and $-$0.003 [$-$0.009, 0.002]). The intervals of the probability measures come from the same bootstrap over groups of items as those of Table~\ref{tab:probci}.

\subsection{Reinforcement learning instead of the supervised loss}
\label{app:rl}

Jev is reported to be trained with reinforcement learning \citep{typesafe2026jev}, so we also train \sansi{} with a reward: after the 200 warm-up steps, the model samples answers from its own distribution at every loop, and every sampled answer is rewarded with its correctness minus the probability that the model stated for it (the procedure is described below). Accuracy is unchanged: 71.7\% at loop 8 against 72.0\% ($-$0.3 points [$-$0.7, 0.1]), and 58.1\%, 66.5\% and 71.3\% at loops 1, 2 and 4, against 58.4\%, 66.9\% and 71.6\%. The probabilities differ. They are better calibrated (ECE 0.076 against 0.093, lower in all three seeds), but they separate items with and without their evidence slightly less well (evidence AUROC 0.919 against 0.935; hard answers to 21.6\% of the unanswerable items against 17.5\%; Table~\ref{tab:rl}). After a short supervised warm-up, a typed decision model can thus be trained from the outcomes of its own decisions alone, without a measurable loss of accuracy.

\paragraph{Procedure.} The loss of Equation~\ref{eq:loss} shows the model the target distribution of every item; the training with a reward does not. The first 200 steps, the warm-up of the learning rate, use Equation~\ref{eq:loss}. From step 201 on, the following is done for every item of a batch and every loop $t$.
\begin{enumerate}
\setlength\itemsep{0.1em}
\item \emph{Act.} The model samples $G = 32$ answers $a_{1}, \dots, a_{G}$ independently from its own distribution $p_{t}$.
\item \emph{Reward.} Every sampled answer receives
\begin{equation}
r_{g} = d(a_{g}) - p_{t}(a_{g}),
\label{eq:reward}
\end{equation}
its correctness minus the probability that the model stated for it: $d(a_{g})$ is 1 if $a_{g}$ is the gold option and 0 otherwise (for an unanswerable item it is $1/K$, and for a crowd-labelled item the share of annotators who chose $a_{g}$).
\item \emph{Advantage.} The baseline is the mean reward of the group: $A_{g} = r_{g} - \frac{1}{G} \sum_{h=1}^{G} r_{h}$.
\item \emph{Update.} The loss of the item at loop $t$ is
\begin{equation}
\mathcal{L}^{\mathrm{RL}}_{t} = -\frac{1}{G} \sum_{g=1}^{G} A_{g} \log p_{t}(a_{g}),
\label{eq:rl}
\end{equation}
and the losses of the loops are averaged as in Equation~\ref{eq:loss}.
\end{enumerate}
The reward is a number through which no gradient passes: the model learns only from the outcomes of the answers it sampled.

\paragraph{Setting.} The reinforcement-learning runs use the data, loops, readout, optimiser and seeds of the main model. With the same seed, steps 1--200 are identical to those of the main model (the same items, order and dropout); from step 201 on, only the training signal differs. The sampled answers come from a random stream of their own.

\paragraph{Relation to GRPO.} This training is the REINFORCE estimator with the mean reward of the group of samples as its baseline, as in GRPO \citep{shao2024grpo}; RLOO \citep{ahmadian2024rloo} leaves the sample itself out of the mean, which would remove the factor $(G-1)/G$ below. Three parts of GRPO are not needed. Every batch is sampled from the current model and used for one update, so there is no importance ratio and no clipping. There is no KL term. And the advantage is not divided by the standard deviation of the group: with two options, the divided advantages depend only on which of the two answers has the larger reward and on how often each was sampled, no longer on how far the stated probability is from the outcome.

\paragraph{Why the reward contains the stated probability.} With a reward of 1 for a correct and 0 for a wrong answer, the expected reward is the probability of the gold option, and it is largest when all probability is put on one option: such a reward trains the answer, not the probability. Subtracting the stated probability makes confidence costly. A wrong answer costs more the more probability the model gave it, and a correct answer earns more the less probability the model gave it. In expectation over the sampled answers, the gradient of Equation~\ref{eq:rl} is $(G-1)/G$ times the gradient of $\frac{1}{2} \sum_{k} (p_{t,k} - d_{k})^{2}$, half the Brier score. The policy gradient of this reward therefore follows a proper scoring rule, which is minimised by the target distribution.

\paragraph{Results.} Table~\ref{tab:rl} compares the two trainings on all metrics; on near transfer the ECE of reinforcement learning is 0.129 against 0.149.

\begin{table*}[t]
\centering
\small
\setlength{\tabcolsep}{4.5pt}
\begin{tabular}{@{}l cccc ccc cc ccc@{}}
\toprule
 & \multicolumn{4}{c}{Accuracy (\%) at loop} & \multicolumn{3}{c}{Accuracy at loop 8} & \multicolumn{2}{c}{ECE} & Hard & Evid. & AUROC \\
\cmidrule(lr){2-5}\cmidrule(lr){6-8}\cmidrule(lr){9-10}
Training & 1 & 2 & 4 & 8 & In & Near & Far & all & near & ans. (\%) & AUROC & r/w \\
\midrule
Loss (cross-entropy + Brier) & 58.4 & 66.9 & 71.6 & 72.0 & 86.6 & 67.7 & 68.0 & .093 & .149 & 17.5 & .935 & .795 \\
 & \sd{0.7} & \sd{0.6} & \sd{0.5} & \sd{0.7} & \sd{0.6} & \sd{2.7} & \sd{0.1} & \sd{.012} & \sd{.031} & \sd{1.3} & \sd{.004} & \sd{.005} \\[2pt]
Reinforcement learning & 58.1 & 66.5 & 71.3 & 71.7 & 86.1 & 66.8 & 67.9 & .076 & .129 & 21.6 & .919 & .793 \\
 & \sd{0.2} & \sd{0.4} & \sd{0.4} & \sd{0.1} & \sd{0.1} & \sd{2.7} & \sd{0.7} & \sd{.015} & \sd{.047} & \sd{4.5} & \sd{.008} & \sd{.002} \\[2pt]
\bottomrule
\end{tabular}

\caption{\sansi{} trained with the supervised loss of Equation~\ref{eq:loss} and with reinforcement learning (Equation~\ref{eq:rl}): 10,027 test items, means of three seeds with the standard deviation over the seeds on the line below. ECE, hard-answer rate and AUROC are those of loop 8.}
\label{tab:rl}
\end{table*}

\paragraph{Seeds.} In the three seeds, reinforcement learning reaches 71.6\%, 71.7\% and 71.7\% at loop 8, against 71.3\%, 72.6\% and 72.0\% for the supervised loss (differences of +0.3, $-$0.9 and $-$0.3 points). Its ECE is lower in all three seeds: 0.093, 0.071 and 0.064 against 0.099, 0.100 and 0.079.

\subsection{The number of loops}
\label{app:nloops}

\paragraph{Four trained loops or eight.} A model trained with four loops, the number of Ouro's pre-training, reaches 70.8\% at its fourth loop. This is 0.8 points below \sansi{} read at the same loop [0.4, 1.2] and 1.1 points below \sansi{} at loop 8 [0.7, 1.6] (Table~\ref{tab:nloops-ci}). Four trained loops are thus enough for most of the gain; training eight adds about one point.

\begin{table*}[t]
\centering
\small
\setlength{\tabcolsep}{8pt}
\begin{tabular}{@{}l cc c@{}}
\toprule
Difference (first $-$ second) & First & Second & Difference [95\% interval] \\
\midrule
\rowcolor{groupbg}\multicolumn{4}{@{}l}{\emph{Accuracy (\%)}} \\
\sansi{}: loop 8 $-$ loop 4 & 72.0 & 71.6 & \dns{+0.4}{-0.0}{+0.7} \\
\sansi{}: loop 12 $-$ loop 8 & 71.2 & 72.0 & \dworse{-0.8}{-1.1}{-0.5} \\
\sansi{}: loop 16 $-$ loop 8 & 70.1 & 72.0 & \dworse{-1.8}{-2.2}{-1.5} \\
Four-loop model: loop 8 $-$ loop 4 & 69.6 & 70.8 & \dworse{-1.3}{-1.7}{-0.9} \\
Four-loop model $-$ \sansi{}, both at loop 4 & 70.8 & 71.6 & \dworse{-0.8}{-1.2}{-0.4} \\
Four-loop model at loop 4 $-$ \sansi{} at loop 8 & 70.8 & 72.0 & \dworse{-1.1}{-1.6}{-0.7} \\
\midrule
\rowcolor{groupbg}\multicolumn{4}{@{}l}{\emph{ECE}} \\
\sansi{}: loop 16 $-$ loop 8 & 0.105 & 0.093 & \dworse{+0.013}{+0.009}{+0.017} \\
Four-loop model: loop 8 $-$ loop 4 & 0.106 & 0.094 & \dworse{+0.012}{+0.007}{+0.016} \\
\midrule
\rowcolor{groupbg}\multicolumn{4}{@{}l}{\emph{Hard-answer rate (\%)}} \\
\sansi{}: loop 16 $-$ loop 8 & 19.7 & 17.5 & \dworse{+2.3}{+0.7}{+3.9} \\
\midrule
\rowcolor{groupbg}\multicolumn{4}{@{}l}{\emph{Evidence AUROC}} \\
\sansi{}: loop 16 $-$ loop 8 & 0.926 & 0.935 & \dworse{-0.008}{-0.015}{-0.003} \\
\midrule
\rowcolor{groupbg}\multicolumn{4}{@{}l}{\emph{Answers between two loops (\% of the answerable items): changed / fixed / broken}} \\
\sansi{}: loop 4 to loop 8 & \multicolumn{3}{c}{9.8 / 4.0 / 3.6} \\
\sansi{}: loop 8 to loop 16 & \multicolumn{3}{c}{10.8 / 3.3 / 5.1} \\
Four-loop model: loop 4 to loop 8 & \multicolumn{3}{c}{11.8 / 3.8 / 5.2} \\
\bottomrule
\end{tabular}

\caption{The number of loops: differences with 95\% bootstrap intervals (means of three seeds), and the share of answers that change, are fixed and are broken between two loops. Colours: \textcolor{better}{\textbf{teal}}, the first term is significantly better (taking the direction of the metric into account; the interval excludes 0); \textcolor{worse}{\textbf{red}}, significantly worse; \textcolor{black!55}{grey}, the interval includes 0.}
\label{tab:nloops-ci}
\end{table*}

The difference between the model trained with four loops and \sansi{}, both read at loop 4, is concentrated on near transfer (2.7 points [1.7, 3.7]; 3.3, 3.6 and 1.2 in the three seeds).

\paragraph{More loops than trained.} Neither model gains from loops beyond the trained ones: both decline soon after the last trained loop (Figure~\ref{fig:nloops}; Table~\ref{tab:nloops-ci}). Table~\ref{tab:nloops} gives the accuracy and the ECE at every loop. The four-loop model falls from 70.8\% at loop 4 to 69.6\% at loop 8 ($-$1.3 points [$-$1.7, $-$0.9]), while \sansi{} is flat over these loops (+0.4 [$-$0.0, 0.7]). Run for 16 loops, \sansi{} falls from 72.0\% at loop 8 to 71.2\% at loop 12 and 70.1\% at loop 16 ($-$1.8 points [$-$2.2, $-$1.5]). Between loop 8 and loop 16 it changes 10.8\% of its answers and breaks more of them (5.1\%) than it fixes (3.3\%), and its ECE rises from 0.093 to 0.105 (+0.013 [0.009, 0.017]). The loops beyond the trained ones have no readout of their own: loops 5--8 of the four-loop model are read with the frozen head, and loops 9--16 of \sansi{} with the readout of loop 8.

\begin{table}[t]
\centering
\small
\setlength{\tabcolsep}{5pt}
\begin{tabular}{@{}c cc cc@{}}
\toprule
 & \multicolumn{2}{c}{\sansi{} (8 trained loops)} & \multicolumn{2}{c}{Trained with 4 loops} \\
\cmidrule(lr){2-3}\cmidrule(l){4-5}
Loop & Acc. (\%) & ECE & Acc. (\%) & ECE \\
\midrule
1 & 58.4\sd{0.7} & .105\sd{.020} & 59.7\sd{0.5} & .123\sd{.010} \\
2 & 66.9\sd{0.6} & .088\sd{.009} & 67.6\sd{0.7} & .096\sd{.005} \\
3 & 70.4\sd{0.6} & .082\sd{.010} & 70.4\sd{0.5} & .093\sd{.007} \\
4 & 71.6\sd{0.5} & .085\sd{.009} & 70.8\sd{0.4} & .094\sd{.005} \\
5 & 71.9\sd{0.7} & .087\sd{.008} & \untrained{70.9}\sd{0.4}$^{\dagger}$ & .103\sd{.009} \\
6 & 72.1\sd{0.7} & .089\sd{.010} & \untrained{70.5}\sd{0.4}$^{\dagger}$ & .103\sd{.007} \\
7 & 72.1\sd{0.7} & .090\sd{.012} & \untrained{70.0}\sd{0.3}$^{\dagger}$ & .104\sd{.005} \\
8 & 72.0\sd{0.7} & .093\sd{.012} & \untrained{69.6}\sd{0.1}$^{\dagger}$ & .106\sd{.005} \\
\midrule
9 & \untrained{71.9}\sd{0.6}$^{\dagger}$ & .093\sd{.011} & -- & -- \\
10 & \untrained{71.7}\sd{0.5}$^{\dagger}$ & .095\sd{.010} & -- & -- \\
11 & \untrained{71.4}\sd{0.6}$^{\dagger}$ & .099\sd{.008} & -- & -- \\
12 & \untrained{71.2}\sd{0.7}$^{\dagger}$ & .100\sd{.008} & -- & -- \\
13 & \untrained{71.1}\sd{0.7}$^{\dagger}$ & .101\sd{.007} & -- & -- \\
14 & \untrained{70.8}\sd{0.6}$^{\dagger}$ & .102\sd{.008} & -- & -- \\
15 & \untrained{70.4}\sd{0.6}$^{\dagger}$ & .105\sd{.006} & -- & -- \\
16 & \untrained{70.1}\sd{0.5}$^{\dagger}$ & .105\sd{.008} & -- & -- \\
\bottomrule
\end{tabular}

\caption{Accuracy and ECE at every loop of \sansi{}, run for 16 loops, and of the model trained with four loops, run for eight (10,027 test items; mean $\pm$ standard deviation over three seeds). $^{\dagger}$: loop beyond the trained ones. Grey values: loops beyond the trained ones.}
\label{tab:nloops}
\end{table}

\begin{figure*}[t]
\centering
\begin{minipage}[b]{0.62\textwidth}\centering
\includegraphics[width=\linewidth]{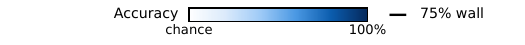}\\[1pt]
\begin{subfigure}[t]{0.5157\linewidth}\includegraphics[width=\linewidth]{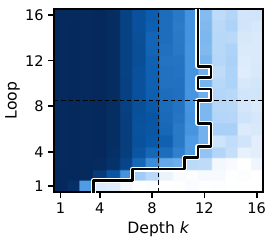}\caption{Liar chains}\end{subfigure}%
\begin{subfigure}[t]{0.4843\linewidth}\includegraphics[width=\linewidth]{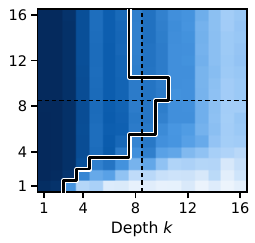}\caption{Object swaps}\end{subfigure}
\end{minipage}
\caption{Accuracy of \sansi{} at every depth $k$ after every loop on the two depth-controlled tasks, up to 16 loops (white: chance; means of three seeds). Dashed: the depths and the loops seen in training. Black line: the depth that the model holds. Table~\ref{tab:depth-summary} gives the summary.}
\label{fig:nloops-grid}
\end{figure*}

On the depth-controlled tasks, further loops do not extend the depth that the model holds (Figure~\ref{fig:nloops-grid}): on the unseen depths, sixteen loops give the accuracy of eight (72.7\% against 72.6\% on liar chains, 64.0\% against 63.7\% on object swaps). A looped decision model can therefore be read after fewer loops than it was trained with (\S\ref{sec:rq2}), but not after more.

\subsection{The backbone}
\label{app:backbone}

\paragraph{Looped pre-training.} The untuned Ouro already improves from loop 1 to loop 4 (33.0\% to 47.9\%), so part of what \sansi{} shows may come from Ouro's looped pre-training. To test whether our recipe alone can create useful loops, we add a loop to SmolLM2-1.7B, which was pre-trained without one: its 24 layers are applied eight times, each pass reading the final hidden state of the previous pass in place of the token embeddings, and the model is trained with the recipe of \sansi{}. The added loop does not train (Table~\ref{tab:retrofit}): the model reaches 33.5\% at loop 8, 24.9 points below SmolLM2 fine-tuned without a loop [23.6, 26.2] and below SmolLM2 without any fine-tuning (38.8\%). A gentler variant, in which the previous state is added to the token embeddings through a linear map that is zero at initialisation, trains stably but does not use its loops: it reaches 58.3\% at loop 8, the accuracy of single-pass SmolLM2 ($-$0.1 points [$-$0.5, 0.3]). With the same data, recipe and number of steps, training every loop thus yields a 13.5-point gain on a backbone that was pre-trained to loop, and no gain on a backbone of the same shape that was not. What \sansi{} gains from its loops was prepared by Ouro's pre-training; our recipe turns it into a decision model but does not create it. This does not show that loops cannot be added after pre-training: \citet{mcleish2025retrofitted} do so with continued training at a far larger budget than our fine-tuning.

\begin{table*}[t]
\centering
\small
\setlength{\tabcolsep}{4pt}
\begin{tabular}{@{}lccc@{}}
\toprule
 & \multicolumn{3}{c}{Accuracy (\%) at loop} \\
\cmidrule(l){2-4}
Model & 1 & 4 & 8 \\
\midrule
SmolLM2-1.7B, not fine-tuned & 38.8 & -- & -- \\
\quad + loop as in Ouro & 38.8 & 32.0 & 31.5 \\
SmolLM2-1.7B, fine-tuned & 58.4\sd{0.7} & -- & -- \\
\quad + loop as in Ouro & 35.4\sd{4.2} & 32.5\sd{0.6} & 33.5\sd{0.4} \\
\quad + loop through a linear map & 58.1\sd{1.1} & 58.3\sd{1.0} & 58.3\sd{1.0} \\
\rowcolor{sansibg}\sansi{} (Ouro-1.4B) & 58.4\sd{0.7} & 71.6\sd{0.5} & 72.0\sd{0.7} \\
\bottomrule
\end{tabular}

\caption{A loop added to SmolLM2-1.7B after pre-training (10,027 test items; mean $\pm$ standard deviation over the seeds: three for SmolLM2-1.7B and \sansi{}, two for the fine-tuned rows with an added loop; the untuned rows are single runs). ``As in Ouro'': the previous state replaces the token embeddings. ``Through a linear map'': the previous state is added to the token embeddings through a rank-64 map that is zero at initialisation.}
\label{tab:retrofit}
\end{table*}

\paragraph{A loop added after pre-training: details.} The first loop of the looped SmolLM2 is the model as released. Without fine-tuning, the added loops already lose seven points (38.8\% at loop 1, 31.5\% at loop 8), whereas the untuned Ouro gains 15 points from loop 1 to loop 4. After fine-tuning, the later loops are not better than the first ($-$1.9 points from loop 1 to loop 8 [$-$2.9, $-$0.8]), and the first loop itself stays 23.0 points below the single-pass model [22.0, 24.2]: the loss on seven loops that cannot yet use their input also prevents the first loop from learning. The probabilities carry no information about missing evidence (evidence AUROC 0.500). During training the gradient norm before clipping is between $10^{3}$ and $10^{6}$, against 3.5--4.6 for single-pass SmolLM2 and 6--20 for \sansi{}, and the training loss does not decrease.

This failure could be an artefact of the abrupt change: from the first step, loops 2--8 read an input that the layers have never seen. We therefore also tried gentler ways of passing the state on, in which training starts from the single-pass model: from loop 2 on, the input is the token embeddings plus a learned function of the previous state that is zero at initialisation. With one scalar gate per loop, the gates stayed within $\pm 0.01$ of zero and all eight loops gave the accuracy of the single-pass model (one seed, stopped after 500 steps: 54.1--55.0\% on the development set, against 53.6\% and 54.7\% for single-pass SmolLM2 at the same step). With a rank-64 linear map of the previous state, shared by all loops, training is stable and the map is used: its norm grows from zero throughout training. The loops nevertheless add nothing. The model reaches 58.3\% at loop 8, the accuracy of single-pass SmolLM2 ($-$0.1 points [$-$0.5, 0.3]) and 13.7 points below \sansi{} [12.6, 14.7]; the answer at loop 8 differs from the answer at loop 1 on only 3.7\% and 5.4\% of the items in the two seeds, and the evidence AUROC stays at the single-pass level (0.774 against 0.765; \sansi{}: 0.935). At the scale of our fine-tuning (1,000 steps, about 5.6 million tokens), a loop added after pre-training thus either does not train or is not put to use. \citet{mcleish2025retrofitted} convert pre-trained models into depth-recurrent ones with a curriculum of recurrences during continued training, at a far larger training budget, and \citet{shapiro2026retrofitting} study the same question.

\paragraph{A larger backbone.} We also train \sansi{} on Ouro-2.6B, the larger backbone of the same family (48 shared layers instead of 24; 2.67B parameters), with the same recipe and eight loops (Table~\ref{tab:main-full} in Appendix~\ref{app:main} and Table~\ref{tab:perloop26}). \sansi{}-2.6B reaches 75.8\% at loop 8, 3.8 points above \sansi{} [3.3, 4.4] and 2.0 points above Qwen3.5-4B [1.4, 2.6], with 63\% of the parameters of the latter; it passes Qwen3.5-4B at its third loop (+0.9 [0.3, 1.5]). The loops add as much as on the smaller backbone: 13.4 points from loop 1 to loop 8 [12.6, 14.3], against 13.6 for \sansi{} (we trained no one-loop control for this backbone, so both numbers compare two readings of one model). Read after one loop, \sansi{}-2.6B is at 62.4\%, 4.3 points below Qwen3.5-2B [3.6, 5.1]: its lead comes from looping, not from a stronger backbone. The price is again computation: a decision of \sansi{}-2.6B takes 14.8 times the GPU time of one loop of Ouro-1.4B and about 6.3 times that of Qwen3.5-4B (Tables~\ref{tab:main-full} and~\ref{tab:params}).

\begin{table*}[t]
\centering
\resizebox{\textwidth}{!}{
\begin{tabular}{@{}c ccccc ccc cc ccc c@{}}
\toprule
 & \multicolumn{5}{c}{Accuracy (\%)} & \multicolumn{3}{c}{ECE} & \multicolumn{2}{c}{Confidence} & AUROC & Evid. & Hard & Answers \\
\cmidrule(lr){2-6}\cmidrule(lr){7-9}\cmidrule(lr){10-11}
Loop & All & In-dist. & Near & Far & JevB. & In-dist. & Near & Far & right & wrong & r/w & AUROC & (\%) & changed (\%) \\
\midrule
1 & 62.4 & 78.8 & 61.9 & 56.6 & 60.6 & .032 & .081 & .118 & .789 & .596 & .756 & .839 & 31.5 & -- \\
 & \sd{0.4} & \sd{0.5} & \sd{1.7} & \sd{0.8} & \sd{0.7} & \sd{.009} & \sd{.022} & \sd{.014} & \sd{.011} & \sd{.014} & \sd{.004} & \sd{.006} & \sd{3.5} &  \\[2pt]
2 & 71.8 & 85.2 & 70.0 & 67.6 & 70.7 & .049 & .103 & .078 & .852 & .650 & .780 & .914 & 22.6 & 27.5 \\
 & \sd{0.7} & \sd{0.2} & \sd{3.0} & \sd{0.2} & \sd{1.7} & \sd{.002} & \sd{.023} & \sd{.010} & \sd{.008} & \sd{.014} & \sd{.003} & \sd{.007} & \sd{3.0} & \sd{0.9} \\[2pt]
3 & 74.7 & 87.8 & 72.5 & 70.6 & 74.6 & .049 & .117 & .073 & .873 & .660 & .795 & .935 & 18.1 & 12.2 \\
 & \sd{0.4} & \sd{0.2} & \sd{2.6} & \sd{0.2} & \sd{2.0} & \sd{.004} & \sd{.020} & \sd{.015} & \sd{.008} & \sd{.015} & \sd{.003} & \sd{.007} & \sd{1.1} & \sd{0.4} \\[2pt]
4 & 75.6 & 88.8 & 73.6 & 71.5 & 76.5 & .044 & .117 & .074 & .878 & .660 & .800 & .938 & 17.0 & 6.4 \\
 & \sd{0.5} & \sd{0.1} & \sd{2.6} & \sd{0.3} & \sd{2.9} & \sd{.002} & \sd{.015} & \sd{.012} & \sd{.008} & \sd{.017} & \sd{.005} & \sd{.003} & \sd{2.2} & \sd{0.2} \\[2pt]
5 & 76.1 & 88.8 & 74.5 & 71.9 & 77.9 & .046 & .111 & .076 & .882 & .666 & .801 & .942 & 17.5 & 3.8 \\
 & \sd{0.6} & \sd{0.5} & \sd{2.5} & \sd{0.3} & \sd{1.9} & \sd{.005} & \sd{.020} & \sd{.012} & \sd{.007} & \sd{.016} & \sd{.007} & \sd{.001} & \sd{2.9} & \sd{0.2} \\[2pt]
6 & 76.1 & 88.6 & 74.7 & 71.9 & 78.8 & .045 & .107 & .076 & .884 & .667 & .804 & .943 & 17.5 & 2.8 \\
 & \sd{0.5} & \sd{0.3} & \sd{2.6} & \sd{0.1} & \sd{0.7} & \sd{.003} & \sd{.019} & \sd{.012} & \sd{.006} & \sd{.014} & \sd{.004} & \sd{.001} & \sd{2.7} & \sd{0.1} \\[2pt]
7 & 76.0 & 88.6 & 74.7 & 71.7 & 78.8 & .047 & .106 & .079 & .885 & .668 & .804 & .943 & 17.7 & 2.4 \\
 & \sd{0.5} & \sd{0.2} & \sd{2.4} & \sd{0.1} & \sd{1.3} & \sd{.008} & \sd{.022} & \sd{.011} & \sd{.004} & \sd{.014} & \sd{.003} & \sd{.001} & \sd{2.6} & \sd{0.3} \\[2pt]
8 & 75.8 & 88.4 & 74.4 & 71.6 & 79.4 & .046 & .106 & .082 & .885 & .669 & .804 & .943 & 18.2 & 2.1 \\
 & \sd{0.6} & \sd{0.8} & \sd{2.5} & \sd{0.3} & \sd{0.7} & \sd{.009} & \sd{.027} & \sd{.011} & \sd{.004} & \sd{.014} & \sd{.004} & \sd{.000} & \sd{2.4} & \sd{0.1} \\[2pt]
\bottomrule
\end{tabular}
}
\caption{\sansi{}-2.6B after every loop (10,027 test items; means of three seeds, with the standard deviation over the seeds on the line below). Columns as in Table~\ref{tab:perloop}.}
\label{tab:perloop26}
\end{table*}

\paragraph{\sansi{}-2.6B: details.} The three seeds of \sansi{}-2.6B reach 75.2\%, 76.4\% and 75.8\% at loop 8 and are 0.9, 3.3 and 1.8 points above Qwen3.5-4B (Table~\ref{tab:seeds}). The lead over Qwen3.5-4B lies in near transfer (5.5 points [3.9, 7.3]) and far transfer (1.5 points [0.6, 2.4]); in distribution the two models are level ($-$0.1 [$-$1.1, 0.8]). \sansi{}-2.6B gains nothing after its fourth loop (+0.2 [$-$0.1, 0.5] from loop 4 to loop 8). Its ECE is 0.078, against 0.113 for Qwen3.5-4B.

\subsection{Averaging the loops}
\label{app:averaging}

The ECE of \sansi{} rises again after loop 3, because confidence keeps rising after the answers have settled (\S\ref{sec:analysis}). A model that is read after every loop offers a remedy that a single-pass model does not have: the option probabilities of its eight loops can be averaged, without labelled data or further training. The average is as accurate as loop 8 (71.8\% against 72.0\%; $-$0.2 points [$-$0.5, 0.1]) and less confident, and its ECE is half as large: 0.044 against 0.093 ($-$0.049 [$-$0.052, $-$0.045]), in each of the three seeds (Table~\ref{tab:loopavg}; the intervals are in Table~\ref{tab:probci}). Temperature scaling \citep{guo2017calibration}, which needs labelled items, does better only when these items cover all test groups; when they come from the training sources alone, the averaged loops have the lower ECE on the other test items (0.049 against 0.060; Table~\ref{tab:loopavg}).

\begin{table*}[t]
\centering
\small
\setlength{\tabcolsep}{4pt}
\begin{tabular}{@{}l c ccc@{}}
\toprule
 & Acc. & \multicolumn{3}{c}{ECE $\downarrow$} \\
\cmidrule(l){3-5}
 & (\%) & all & near & far \\
\midrule
\rowcolor{groupbg}\multicolumn{5}{@{}l}{\emph{All test items; no calibration data}} \\
Loop 8 & 72.0\sd{0.7} & .093\sd{.012} & .149\sd{.031} & .092\sd{.019} \\
\rowcolor{sansibg}Mean of loops 1--8 & 71.8\sd{0.7} & .044\sd{.009} & .084\sd{.028} & .045\sd{.010} \\
\midrule
\rowcolor{groupbg}\multicolumn{5}{@{}l}{\emph{Temperature fitted on in-distribution items}} \\
Loop 8 & 70.2\sd{0.7} & .098\sd{.013} & .151\sd{.032} & .092\sd{.019} \\
\rowcolor{sansibg}Mean of loops 1--8 & 70.0\sd{0.8} & .049\sd{.009} & .086\sd{.029} & .045\sd{.010} \\
Loop 8 + temp. & 70.3\sd{0.7} & .060\sd{.009} & .102\sd{.037} & .054\sd{.013} \\
\rowcolor{sansibg}Mean of loops + temp. & 69.9\sd{0.8} & .066\sd{.008} & .108\sd{.034} & .062\sd{.004} \\
\midrule
\rowcolor{groupbg}\multicolumn{5}{@{}l}{\emph{Temperature fitted on items of all groups}} \\
Loop 8 & 71.7\sd{0.8} & .094\sd{.013} & .157\sd{.032} & .093\sd{.019} \\
\rowcolor{sansibg}Mean of loops 1--8 & 71.5\sd{0.8} & .046\sd{.008} & .090\sd{.033} & .048\sd{.009} \\
Loop 8 + temp. & 71.9\sd{0.8} & .025\sd{.003} & .068\sd{.028} & .031\sd{.003} \\
\rowcolor{sansibg}Mean of loops + temp. & 71.5\sd{0.8} & .019\sd{.005} & .061\sd{.023} & .026\sd{.002} \\
\bottomrule
\end{tabular}

\caption{Averaging the option probabilities of the eight loops, and temperature scaling. In the two lower blocks a random half of the groups of related items in the in-distribution test set, or in the whole test set, serves as calibration data, and the rows are evaluated on the remaining test items (means of ten splits). All numbers are mean $\pm$ standard deviation over three seeds; temp.: temperature scaling. Shaded rows: the average of the eight loops.}
\label{tab:loopavg}
\end{table*}

\paragraph{Comparison with temperature scaling.} Table~\ref{tab:loopavg} also compares the average of the eight loops with temperature scaling \citep{guo2017calibration}, which fits one temperature on labelled items that the model was not trained on. The average is less confident than loop 8 (mean confidence 0.752 against 0.805); on near transfer its ECE is 0.084 against 0.149 ($-$0.065 [$-$0.075, $-$0.056]). When the temperature is fitted on held-out items of the training sources, the averaged loops have the lower ECE on the other test items (about 8,940): 0.049 against 0.060; scaling the average as well does not help (0.066). When the temperature is fitted on items of all test groups, temperature scaling is better (0.025 against 0.046), and scaling the average gives the lowest ECE (0.019). We did not compute intervals for the comparisons with temperature scaling.

\section{Verifier Case Study: Details}
\label{app:verifier}

This appendix supports \S\ref{sec:verifier}. It gives the settings of the case study (Appendix~\ref{app:verifier-settings}), how the answers are graded (Appendix~\ref{app:verifier-grading}), its results (Appendix~\ref{app:verifier-results}), and the reason why exact match does not rise (Appendix~\ref{app:verifier-em}).

\subsection{Settings}
\label{app:verifier-settings}

The generator has 18.1M trained parameters (LoRA adapters). We use 8,000 training, 500 development and 3,000 test questions of 2WikiMultiHopQA, balanced over the four question types; each question comes with its supporting paragraphs and distractors (five paragraphs in total). The verifier is seed 0 of the main model, called in the prompt format it was trained with. Answers that are equal after normalisation or overlap with a token F1 of at least 0.8 count as one option. If fewer than four distinct answers are sampled, short spans of the paragraphs are added as further options. Training runs for 400 steps; at each step eight questions are drawn and eight answers are sampled for each. Advantages are normalised within the eight answers of a question, and each batch is used for one update. Besides the grade of the greedy answer (Appendix~\ref{app:verifier-grading}), we report its token F1 against the gold answer, exact match (EM) and the share of answers that contain the gold answer.

\subsection{Grading the answers}
\label{app:verifier-grading}

Token F1 and exact match compare the words of an answer with those of the gold answer, so they also measure its form. We therefore grade every answer to the 3,000 test questions with a language model, Claude Sonnet 5.5 (\texttt{claude-sonnet-5-5}, October 2026, with the default settings of the API and without extended thinking), as is common for open-ended answers \citep{zheng2023judging}. The grader is used only for evaluation, never as a reward. It sees the question, the gold answers and the answer, but not the paragraphs, and returns one of four labels (Figure~\ref{fig:grader-prompt}): \emph{correct}, the gold answer or an equivalent of it, possibly with further words that only add detail; \emph{partial}, only part of the gold answer, or a broader or narrower answer; \emph{hedged}, several candidate answers of which one is right; and \emph{wrong}. An answer counts as right only when it is graded correct. Answers to the same question that are equal after lowercasing and removing extra spaces are graded once, so the 39,000 answers of the 13 runs (the untrained generator and four rewards with three seeds each) give 13,294 distinct pairs of question and answer. A reply that is not one of the labels was requested again, up to three times; six answers remain without a label and count as wrong.

Three checks support the grades. All 9,144 answers that match the gold answer exactly are graded correct. All 499 answers that were moved to a different question are graded wrong. Of 500 pairs graded a second time with an instruction that is worded differently and gives the answer before the question (Figure~\ref{fig:grader-prompt}), 97.4\% receive the same label (Cohen's $\kappa = 0.95$) and 98.0\% the same decision between correct and not correct.

\begin{figure*}[t]
\centering
\begin{tcolorbox}[colback=white, colframe=black!45, boxrule=0.4pt, arc=0pt, left=4pt, right=4pt, top=3pt, bottom=3pt, width=\textwidth, fontupper=\footnotesize]
\begin{minipage}[t]{0.485\textwidth}
\textbf{(a) Instruction used for all grades}\par\smallskip
\raggedright\ttfamily Question: $\langle$question$\rangle$\\
Reference answer: $\langle$gold answer$\rangle$\\
Answer to check: $\langle$answer$\rangle$\par\smallskip
Label the answer to check with exactly one word:\\
CORRECT - it gives the reference answer or an equivalent of it (other wording, an abbreviation, or extra words that only add detail, such as a date or a title), and nothing that contradicts it.\\
PARTIAL - it gives only part of the reference answer, or a broader or narrower answer that does not identify it (for example only a surname, or a country where a city is asked for).\\
HEDGED - it gives several different candidate answers, one of which matches the reference answer.\\
WRONG - anything else.\\
Reply with the label only.
\end{minipage}\hfill
\begin{minipage}[t]{0.485\textwidth}
\textbf{(b) Second wording, for the agreement check}\par\smallskip
\raggedright\ttfamily Here is a question with its reference answer, and an answer to grade.\par\smallskip
Answer to grade: $\langle$answer$\rangle$\\
Question: $\langle$question$\rangle$\\
Reference answer: $\langle$gold answer$\rangle$\par\smallskip
Reply with one word: CORRECT if the answer to grade gives the reference answer or an equivalent of it (other wording, an abbreviation, or extra words that only add detail, such as a date or a title) and nothing that contradicts it; PARTIAL if it gives only part of the reference answer or a broader or narrower answer that does not identify it; HEDGED if it gives several different candidate answers, one of which matches the reference answer; WRONG otherwise.
\end{minipage}
\end{tcolorbox}
\caption{Instructions to the grader. The system message is ``You check short answers to questions against reference answers. You reply with a single word.'' When a question has several gold answers, ``reference answer'' reads ``reference answers (any one is right)'', and the answers are separated by ``|''.}
\label{fig:grader-prompt}
\end{figure*}

\subsection{Results}
\label{app:verifier-results}

\begin{figure*}[t]
\centering
\includegraphics[width=3.3in]{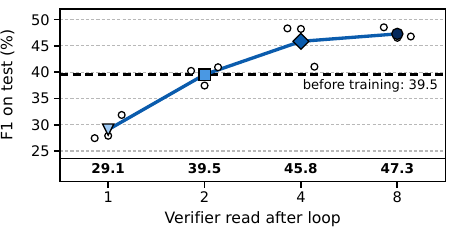}
\caption{Token F1 of the generator on the 3,000 test questions, by the loop at which \sansi{} is read (filled: mean of three seeds; hollow: seeds; dashed: before training). Figure~\ref{fig:grpo} gives the share of answers graded correct in the same layout.}
\label{fig:grpo-f1}
\end{figure*}

Table~\ref{tab:grpo-ci} gives the share of answers graded correct and the F1 of the generator after training, their change against the generator before training with a 95\% interval, and the quality of the reward early in training. Figure~\ref{fig:grpo-f1} shows the F1 in the layout of Figure~\ref{fig:grpo}: it rises with the loop of the verifier as the share of correct answers does, but falls below its value before training with the one-loop reward, whose answers are twice as long (Appendix~\ref{app:verifier-em}). The quality of the reward is the AUROC with which it separates the sampled answers that match the gold answer from those that do not, in the first 20 training steps.

\begin{table*}[t]
\centering
\small
\setlength{\tabcolsep}{4pt}
\begin{tabular}{@{}l cc cc c@{}}
\toprule
 & \multicolumn{2}{c}{Graded correct (\%)} & \multicolumn{2}{c}{F1} & Reward AUROC, \\
\cmidrule(lr){2-3}\cmidrule(lr){4-5}
Reward & After training & Change [95\% interval] & After training & Change [95\% interval] & first 20 steps \\
\midrule
Not trained & 37.0 & -- & 39.5 & -- & -- \\
\midrule
\sansi{} read at loop 1 & 37.5\sd{0.6} & \dns{+0.4}{-1.2}{+2.2} & 29.1\sd{2.4} & \dworse{-10.5}{-11.9}{-9.0} & .779\sd{.050} \\
\sansi{} read at loop 2 & 46.0\sd{1.2} & \dbetter{+9.0}{+7.3}{+10.8} & 39.5\sd{1.8} & \dns{-0.0}{-1.6}{+1.6} & .857\sd{.030} \\
\sansi{} read at loop 4 & 51.1\sd{0.5} & \dbetter{+14.0}{+12.1}{+15.8} & 45.8\sd{4.2} & \dbetter{+6.3}{+4.6}{+8.0} & .914\sd{.032} \\
\sansi{} read at loop 8 & 52.1\sd{1.9} & \dbetter{+15.1}{+13.3}{+16.9} & 47.3\sd{1.1} & \dbetter{+7.7}{+6.1}{+9.4} & .905\sd{.025} \\
\midrule
Loop 8 $-$ loop 4 & -- & \dbetter{+1.0}{+0.3}{+1.8} & -- & \dbetter{+1.4}{+0.8}{+2.0} & -- \\
\bottomrule
\end{tabular}

\caption{The verifier case study in summary (3,000 test questions; mean $\pm$ standard deviation over three training seeds). Graded correct: share of answers graded correct (Appendix~\ref{app:verifier-grading}). Change: against the generator before training, with a 95\% bootstrap interval over the test questions. Reward AUROC: how well the reward separates the sampled answers that match the gold answer from those that do not, in the first 20 training steps. Colours: \textcolor{better}{\textbf{teal}}, the first term is significantly better (the interval excludes 0); \textcolor{worse}{\textbf{red}}, significantly worse; \textcolor{black!55}{grey}, the interval includes 0.}
\label{tab:grpo-ci}
\end{table*}

Table~\ref{tab:grpo} gives all measures of the generator: the share of answers graded correct, F1, exact match, the share of answers that contain the gold answer, and the length of the answers. Table~\ref{tab:grpo-grades} gives the shares of all four grades and the share graded correct by question type.

\begin{table*}[t]
\centering
\small
\setlength{\tabcolsep}{2.5pt}
\begin{tabular}{@{}l cccc cccc c@{}}
\toprule
 & \multicolumn{4}{c}{All questions} & \multicolumn{4}{c}{F1 by question type} & Answer \\
\cmidrule(lr){2-5}\cmidrule(lr){6-9}
Reward & Graded & F1 & EM & Contains & Comparison & \begin{tabular}[b]{@{}c@{}}Bridge\\comparison\end{tabular} & Compositional & Inference & words \\
\midrule
Not trained & 37.0 & 39.5 & 32.5 & 34.5 & 50.6 & 48.2 & 30.7 & 28.6 & 2.6 \\
\midrule
\sansi{} read at loop 1 & 37.5\sd{0.6} & 29.1\sd{2.4} & 9.6\sd{3.0} & 38.5\sd{1.0} & 34.0\sd{3.2} & 32.7\sd{4.9} & 28.6\sd{2.8} & 21.0\sd{2.7} & 5.5\sd{0.5} \\
\sansi{} read at loop 2 & 46.0\sd{1.2} & 39.5\sd{1.8} & 19.6\sd{5.4} & 45.0\sd{1.9} & 38.9\sd{2.9} & 42.4\sd{7.5} & 36.3\sd{1.5} & 40.5\sd{3.4} & 4.3\sd{0.5} \\
\sansi{} read at loop 4 & 51.1\sd{0.5} & 45.8\sd{4.2} & 29.4\sd{10.7} & 49.7\sd{0.9} & 49.2\sd{8.2} & 48.5\sd{6.2} & 37.3\sd{1.9} & 48.4\sd{1.1} & 3.9\sd{0.8} \\
\sansi{} read at loop 8 & 52.1\sd{1.9} & 47.3\sd{1.1} & 32.2\sd{3.8} & 49.7\sd{2.0} & 49.4\sd{5.1} & 51.6\sd{0.7} & 38.3\sd{1.5} & 49.7\sd{0.9} & 3.6\sd{0.3} \\
\bottomrule
\end{tabular}

\caption{The generator on the 3,000 test questions of 2WikiMultiHopQA (750 per question type) after 400 GRPO steps, by the loop at which the rewarding \sansi{} is read. Mean $\pm$ standard deviation over three training seeds (Table~\ref{tab:grpo-seeds} lists them); the untrained generator is a single run. Graded: share of answers graded correct (Appendix~\ref{app:verifier-grading}). Contains: share of answers in which the gold answer occurs as a sequence of whole words, after normalisation. Answer words: mean length of the generated answer.}
\label{tab:grpo}
\end{table*}

\begin{table*}[t]
\centering
\small
\setlength{\tabcolsep}{3pt}
\begin{tabular}{@{}l ccccc cccc@{}}
\toprule
 & \multicolumn{5}{c}{Grades of all answers (\%)} & \multicolumn{4}{c}{Graded correct by question type (\%)} \\
\cmidrule(lr){2-6}\cmidrule(lr){7-10}
Reward & Correct & Partial & Hedged & Wrong & \begin{tabular}[b]{@{}c@{}}Correct,\\not EM\end{tabular} & Comparison & \begin{tabular}[b]{@{}c@{}}Bridge\\comparison\end{tabular} & Compositional & Inference \\
\midrule
Not trained & 37.0 & 2.2 & 0.0 & 60.8 & 4.6 & 50.1 & 48.0 & 34.7 & 15.3 \\
\midrule
\sansi{} read at loop 1 & 37.5\sd{0.6} & 1.3\sd{0.0} & 0.5\sd{0.1} & 60.8\sd{0.5} & 27.9\sd{2.7} & 49.2\sd{4.4} & 41.2\sd{0.3} & 46.3\sd{2.6} & 13.2\sd{3.2} \\
\sansi{} read at loop 2 & 46.0\sd{1.2} & 1.2\sd{0.3} & 0.4\sd{0.2} & 52.4\sd{1.2} & 26.4\sd{5.9} & 52.8\sd{1.5} & 47.2\sd{2.5} & 51.5\sd{2.6} & 32.7\sd{4.6} \\
\sansi{} read at loop 4 & 51.1\sd{0.5} & 1.0\sd{0.2} & 0.7\sd{0.3} & 47.3\sd{0.1} & 21.6\sd{10.2} & 54.8\sd{0.6} & 50.0\sd{3.2} & 55.7\sd{1.1} & 43.7\sd{1.6} \\
\sansi{} read at loop 8 & 52.1\sd{1.9} & 1.0\sd{0.1} & 0.3\sd{0.2} & 46.7\sd{2.0} & 19.9\sd{5.0} & 55.0\sd{1.0} & 51.8\sd{0.5} & 56.0\sd{3.1} & 45.6\sd{3.5} \\
\bottomrule
\end{tabular}

\caption{Grades of the generator's answers to the 3,000 test questions (Appendix~\ref{app:verifier-grading}; mean $\pm$ standard deviation over three training seeds; the untrained generator is a single run). Wrong includes the six answers without a grade. Correct, not EM: answers graded correct that do not match the gold answer exactly. By question type: 750 questions each.}
\label{tab:grpo-grades}
\end{table*}

\begin{figure*}[t]
\centering
\includegraphics[width=3.3in]{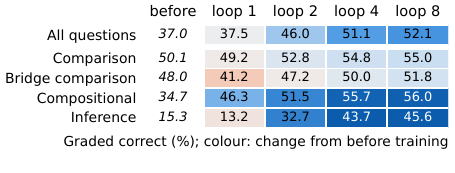}
\caption{Share of the generator's answers graded correct (\%), before training and by the loop at which \sansi{} is read, by question type (3,000 test questions, 750 per type; means of three seeds; values in Table~\ref{tab:grpo-grades}). Colour: change from before training (blue: higher; red: lower).}
\label{fig:grpo-types}
\end{figure*}

\paragraph{Question types.} Figure~\ref{fig:grpo-types} gives the change in the share of answers graded correct by question type. With eight loops the gain is largest on inference questions, which require combining two facts (15.3\% to 45.6\%), and on compositional questions (34.7\% to 56.0\%); on the two kinds of comparison, whose answer is yes, no or one of two names in the question, the gain is small (50.1\% to 55.0\% and 48.0\% to 51.8\%). The one-loop reward gains on compositional questions (+11.6 points) but loses on bridge comparisons ($-$6.8), so that its overall share is unchanged.

\begin{figure*}[t]
\centering
\includegraphics[width=3.3in]{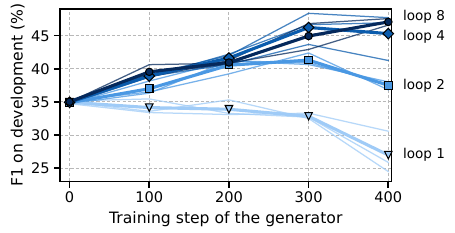}
\caption{F1 on the 500 development questions during the training of the generator, by the loop at which the verifier is read (thick lines: means of three training seeds; thin lines: the seeds).}
\label{fig:grpo-curves}
\end{figure*}

\paragraph{Training seeds.} Table~\ref{tab:grpo-seeds} lists every training seed, and Figure~\ref{fig:grpo-curves} shows the F1 on the development questions during training. Graded, the three seeds agree closely: 50.6--54.3\% of the answers are correct with the eight-loop reward, 50.5--51.4\% with four loops, 44.7--47.2\% with two and 36.7--37.9\% with one. F1 varies more, because it also depends on the length of the answers (46.5--48.5, 41.0--48.3, 37.4--40.9 and 27.4--31.9). The run with the four-loop reward that lost F1 during its last 100 steps (41.0, against 48.2 and 48.3) wrote longer answers (4.7 words on the test questions, against 3.4 and 3.5) but has as many correct answers as the other two (50.5\%, against 51.4\% and 51.3\%). In every seed, the one-loop reward gives the fewest correct answers and the two-loop reward the second fewest. The four- and eight-loop rewards cannot be separated: the eight-loop run has 0.0, $-$0.7 and +3.8 points more correct answers in the three seeds. The interval of their difference in Table~\ref{tab:grpo-ci} (+1.0 points [0.3, 1.8]) resamples the test questions and does not cover this variation between training runs.

\begin{table*}[t]
\centering
\small
\setlength{\tabcolsep}{5pt}
\begin{tabular}{@{}l ccc ccc ccc ccc ccc@{}}
\toprule
 & \multicolumn{3}{c}{Graded correct} & \multicolumn{3}{c}{F1} & \multicolumn{3}{c}{EM} & \multicolumn{3}{c}{Contains} & \multicolumn{3}{c}{Answer words} \\
\cmidrule(lr){2-4}\cmidrule(lr){5-7}\cmidrule(lr){8-10}\cmidrule(lr){11-13}\cmidrule(lr){14-16}
Reward, seed & 0 & 1 & 2 & 0 & 1 & 2 & 0 & 1 & 2 & 0 & 1 & 2 & 0 & 1 & 2 \\
\midrule
\sansi{} read at loop 1 & 36.7 & 37.7 & 37.9 & 27.4 & 27.9 & 31.9 & 8.2 & 7.5 & 13.0 & 37.5 & 38.4 & 39.5 & 5.9 & 5.6 & 4.9 \\
\sansi{} read at loop 2 & 44.7 & 46.2 & 47.2 & 40.2 & 37.4 & 40.9 & 24.0 & 13.6 & 21.3 & 43.0 & 45.4 & 46.7 & 3.9 & 4.8 & 4.3 \\
\sansi{} read at loop 4 & 51.4 & 51.3 & 50.5 & 48.3 & 48.2 & 41.0 & 36.3 & 34.8 & 17.1 & 48.9 & 49.5 & 50.7 & 3.4 & 3.5 & 4.7 \\
\sansi{} read at loop 8 & 51.4 & 50.6 & 54.3 & 48.5 & 46.5 & 46.8 & 36.5 & 30.7 & 29.3 & 48.7 & 48.3 & 52.0 & 3.4 & 3.5 & 4.0 \\
\bottomrule
\end{tabular}

\caption{The verifier case study seed by seed: the generator on the 3,000 test questions after 400 GRPO steps. Before training, 37.0\% of its answers are graded correct, and it has F1 39.5, EM 32.5, 34.5\% of answers that contain the gold answer, and 2.6 words per answer.}
\label{tab:grpo-seeds}
\end{table*}

\subsection{Exact match and the form of the answers}
\label{app:verifier-em}

Exact match does not improve. It is 32.2 with the eight-loop reward, against 32.5 before training, and lower with fewer loops (29.4, 19.6 and 9.6). The cause is the form of the answers, not their correctness. The gold answers are short, and the generator learns to write longer ones: 3.6 words on average with the eight-loop reward and 5.5 with the one-loop reward, against 2.6 before training. With the eight-loop reward, 17.5\% of the answers contain the gold answer together with further words, for instance ``Dr. Socrates (1935)'' where the gold answer is ``Dr. Socrates''; before training, 2.0\% do. The grader counts 95.6\% of these answers as correct, and 1.2\% as hedged. Over all answers, at most 0.7\% are graded hedged with any reward (Table~\ref{tab:grpo-grades}), so the longer answers are not a way of guessing several times. Part of the cause lies in the reward: a variant with one added word counts as the same option as the shorter answer whenever that answer has at least two words, so the reward cannot prefer the shorter form. The one-loop reward changes the form of the answers but not their correctness: they are twice as long as before training, and exact match falls from 32.5 to 9.6 and F1 from 39.5 to 29.1, but as many answers are graded correct (37.5\%, against 37.0\%). Exact match also varies more across the seeds than F1, because it depends on whether a run has learned to write its answers with further words.

\clearpage
\section{Error Analysis: Details}
\label{app:errors}

This appendix supports the error analysis of \S\ref{sec:examples}. All numbers are computed on the 8,879 test items with one gold option, that is, without the unanswerable and the crowd-labelled items, and over the three seeds of every model (26,637 pairs of an item and a seed). An item is an error of a model when its most probable option is not the gold option. \sansi{} is read at loop 8. Table~\ref{tab:errors} gives all numbers of this analysis. Most errors are not specific to looping: Qwen3.5-4B also fails on 68.0\% of them, and 60.9\% are wrong at every loop of \sansi{}.

\begin{table}[t]
\centering
\small
\setlength{\tabcolsep}{14pt}
\begin{tabular}{@{}l r@{}}
\toprule
\rowcolor{groupbg}\multicolumn{2}{@{}l}{\emph{Share of the items (\%)}} \\
\sansi{} wrong at loop 8 & 27.2 \\
Qwen3.5-4B wrong & 25.7 \\
SmolLM2-1.7B wrong & 41.1 \\
\sansi{} and Qwen3.5-4B both wrong & 18.5 \\
Only \sansi{} wrong & 8.7 \\
Only Qwen3.5-4B wrong & 7.2 \\
Wrong at loop 1, right at loop 8 (\textcolor{better}{fixed}) & 20.9 \\
Right at loop 1, wrong at loop 8 (\textcolor{worse}{broken}) & 6.4 \\
\midrule
\rowcolor{groupbg}\multicolumn{2}{@{}l}{\emph{Share of the errors of \sansi{} at loop 8 (\%)}} \\
Qwen3.5-4B is also wrong & 68.0 \\
Qwen3.5-4B chooses the same wrong option & 51.7 \\
SmolLM2-1.7B is also wrong & 69.6 \\
Wrong at every loop & 60.9 \\
Right at some earlier loop & 39.1 \\
Right at loop 1 & 23.6 \\
Confidence at least 0.9 & 19.7 \\
\midrule
\rowcolor{groupbg}\multicolumn{2}{@{}l}{\emph{Share of the errors of Qwen3.5-4B (\%)}} \\
\sansi{} is also wrong at loop 8 & 72.0 \\
Confidence at least 0.9 & 24.0 \\
\midrule
\rowcolor{groupbg}\multicolumn{2}{@{}l}{\emph{Error rate of \sansi{} at loop 8 by test group (\%)}} \\
In distribution & 13.2 \\
Near transfer & 30.2 \\
Far transfer & 31.2 \\
JevBench & 27.7 \\
\bottomrule
\end{tabular}

\caption{Error analysis on the 8,879 test items with one gold option (three seeds of every model). Confidence: the probability of the chosen option.}
\label{tab:errors}
\end{table}

\paragraph{Errors shared with the single-pass models.} \sansi{} is wrong on 27.2\% of the items, Qwen3.5-4B on 25.7\% and SmolLM2-1.7B on 41.1\%. \sansi{} and Qwen3.5-4B are both wrong on 18.5\% of the items, only \sansi{} on 8.7\% and only Qwen3.5-4B on 7.2\%. Qwen3.5-4B is therefore wrong on 68.0\% of the errors of \sansi{}, and on 51.7\% of them it chooses the same wrong option. In the other direction, \sansi{} is wrong on 72.0\% of the errors of Qwen3.5-4B. SmolLM2-1.7B is wrong on 69.6\% of the errors of \sansi{}.

\paragraph{Errors across the loops.} Of the errors of \sansi{} at loop 8, 60.9\% are wrong at every loop. The other 39.1\% were right at some earlier loop, and 23.6\% were right at loop 1. Between loop 1 and loop 8 the loops fix 20.9\% of the items and break 6.4\%. On the 9,645 answerable items, which also include the 766 crowd-labelled items, the same quantities are 21.0\% and 7.3\% (\S\ref{sec:rq2}).

\paragraph{Errors with high confidence.} Of the errors of \sansi{}, 19.7\% carry a confidence of at least 0.9. For Qwen3.5-4B this share is 24.0\%. These errors would pass any confidence threshold below 0.9. This is expected from \S\ref{sec:analysis}: the loops raise the confidence of wrong answers as well as of right ones.

\paragraph{Errors by test group.} The error rate of \sansi{} is 13.2\% in distribution, 30.2\% on near transfer, 31.2\% on far transfer and 27.7\% on the JevBench items.

\paragraph{Examples of errors.} Table~\ref{tab:examples} in Appendix~\ref{app:examples} contains three errors of \sansi{} at loop 8: an item from WANLI on which Qwen3.5-4B and SmolLM2-1.7B choose the same wrong option; an item from QNLI that is right at loop 1 and broken by the later loops; and a science question from ARC that Qwen3.5-4B answers correctly and \sansi{} gets wrong at every loop.

\providecommand{\ans}[2]{#1 (#2)}
\providecommand{\ansok}[2]{\textcolor{better}{#1} (#2)}
\providecommand{\answrong}[2]{\textcolor{worse}{#1} (#2)}
\section{Examples}
\label{app:examples}
Table~\ref{tab:examples} summarises five test items with the answers of \sansi{} after its first and its last loop and the answers of two single-pass models; in each case the other two seeds of \sansi{} give the same answer at loop 8. The examples that follow show, for seven items, the prompt as the model reads it and the probability of every option after each of the eight loops of \sansi{} and for the three single-pass models (seed 0). The gold option is marked with a star (in \textcolor{better}{teal}), the largest probability of every column is in bold, and every cell is shaded by its probability (blue: \sansi{}; green: Qwen3.5; orange: SmolLM2-1.7B). Examples 1 to 5 are the item of Figure~\ref{fig:example} and the first four items of Table~\ref{tab:examples}; Examples 6 and 7 are one question with and without its key evidence.\par

\begin{table*}[t]
\centering
\footnotesize
\setlength{\tabcolsep}{3.5pt}
\begin{tabular}{@{}>{\raggedright\arraybackslash}p{1.6cm}>{\raggedright\arraybackslash}p{4.2cm}>{\raggedright\arraybackslash}p{1.65cm}>{\raggedright\arraybackslash}p{1.7cm}>{\raggedright\arraybackslash}p{1.7cm}>{\raggedright\arraybackslash}p{1.7cm}>{\raggedright\arraybackslash}p{1.7cm}@{}}
\toprule
 & & & \multicolumn{2}{c}{\sansi{}} & & \\
\cmidrule(lr){4-5}
 & Item (source) & Gold & loop 1 & loop 8 & Qwen3.5-4B & SmolLM2 \\
\midrule
Fixed by the loops & ``No road is dustless. Some streets are roads.'' Is ``Some streets are dustless.'' true, false, or unknown? (FOLIO) & unknown & \answrong{true}{.83} & \ansok{unknown}{.97} & \ansok{unknown}{.89} & \answrong{true}{.57} \\
\addlinespace
Fixed; larger model wrong & If 30,000 is divided by 10 and then divided by 10 again, what will be the resulting number? 3 / 30 / 300 / 3,000 (MMLU) & 300 & \answrong{3}{.60} & \ansok{300}{.91} & \answrong{3,000}{.87} & \ansok{300}{.38} \\
\addlinespace
Missing knowledge & When cold temperatures are produced in a chemical reaction, the reaction is known as \dots{} (ARC) & endothermic & \answrong{exothermic}{.96} & \answrong{exothermic}{.99} & \ansok{endothermic}{.97} & \answrong{exothermic}{.48} \\
\addlinespace
Broken by the loops & ``By 9000 BP, Europe was fully forested.'' Does the sentence contain the answer to ``When was Europe fully forested and recovered from the last Ice Age?'' (QNLI) & yes & \ansok{yes}{.98} & \answrong{no}{.89} & \answrong{no}{.77} & \ansok{yes}{.96} \\
\addlinespace
Shared error & ``He can't be here.'' Hypothesis: ``He is here.'' (WANLI) & neutral & \answrong{contradiction}{.99} & \answrong{contradiction}{.96} & \answrong{contradiction}{.99} & \answrong{contradiction}{.71} \\
\bottomrule
\end{tabular}
\caption{Five test items, all from sources not seen in training, with the answers of \sansi{} after its first and its last loop and of two single-pass models. Each cell gives the answer (\textcolor{better}{right}, \textcolor{worse}{wrong}) and, in parentheses, its probability (seed 0).}
\label{tab:examples}
\end{table*}

\par\vspace{10pt}   

\noindent\begin{minipage}{\columnwidth}
\textbf{Example 1 (PAWS, far transfer).} The item of Figure~\ref{fig:example}: the first loop follows the word overlap; the answer is right from the second loop.\par
\vspace{4pt}
{\footnotesize\raggedright\ttfamily\noindent The six people killed were four Burmese citizens and two Russians .\\
\mbox{}\\
Question: Does this sentence mean the same thing: "The six people killed were four Russian and two Burmese citizens ."\\
Options: (A) no: Different meaning, even if most words match (B) yes: Same meaning, possibly reworded\\
Answer:\par}
\vspace{4pt}
\noindent{\scriptsize\setlength{\tabcolsep}{2.6pt}
\begin{tabular}{@{}l cccccccc @{\hspace{6pt}} ccc@{}}
\toprule
 & \multicolumn{8}{c}{\sansi{}, after loop} & Qwen & Qwen & Smol \\
\cmidrule(lr){2-9}
 & 1 & 2 & 3 & 4 & 5 & 6 & 7 & 8 & 4B & 2B & LM2 \\
\midrule
\textcolor{better}{\textbf{(A)}$^{\star}$} & \cellcolor{sansi!9}.19 & \cellcolor{sansi!31}\textbf{.69} & \cellcolor{sansi!44}\textbf{.98} & \cellcolor{sansi!45}\textbf{.99} & \cellcolor{sansi!45}\textbf{.99} & \cellcolor{sansi!45}\textbf{.99} & \cellcolor{sansi!45}\textbf{.99} & \cellcolor{sansi!45}\textbf{.99} & \cellcolor{qwen!43}\textbf{.95} & \cellcolor{qwen!3}.06 & \cellcolor{smol!28}\textbf{.62} \\
(B) & \cellcolor{sansi!36}\textbf{.81} & \cellcolor{sansi!14}.31 & \cellcolor{sansi!1}.02 & .01 & .01 & .01 & .01 & .01 & \cellcolor{qwen!2}.05 & \cellcolor{qwen!42}\textbf{.94} & \cellcolor{smol!17}.38 \\
\bottomrule
\end{tabular}\par}
\end{minipage}\par\vspace{10pt}

\noindent\begin{minipage}{\columnwidth}
\textbf{Example 2 (FOLIO, far transfer).} Fixed by the loops: the first loop answers ``true'', the second ``unknown''.\par
\vspace{4pt}
{\footnotesize\raggedright\ttfamily\noindent No road is dustless. Some streets are roads.\\
\mbox{}\\
Question: Using only the facts and rules above, is the statement "Some streets are dustless." true, false, or unknown?\\
Options: (A) true (B) false (C) unknown\\
Answer:\par}
\vspace{4pt}
\noindent{\scriptsize\setlength{\tabcolsep}{2.6pt}
\begin{tabular}{@{}l cccccccc @{\hspace{6pt}} ccc@{}}
\toprule
 & \multicolumn{8}{c}{\sansi{}, after loop} & Qwen & Qwen & Smol \\
\cmidrule(lr){2-9}
 & 1 & 2 & 3 & 4 & 5 & 6 & 7 & 8 & 4B & 2B & LM2 \\
\midrule
(A) & \cellcolor{sansi!37}\textbf{.83} & \cellcolor{sansi!3}.07 & .01 & .01 & .01 & .01 & .01 & \cellcolor{sansi!1}.02 & .00 & .00 & \cellcolor{smol!26}\textbf{.57} \\
(B) & .01 & .01 & .00 & .00 & .00 & .00 & .01 & .01 & \cellcolor{qwen!5}.11 & \cellcolor{qwen!17}.38 & \cellcolor{smol!3}.07 \\
\textcolor{better}{\textbf{(C)}$^{\star}$} & \cellcolor{sansi!7}.16 & \cellcolor{sansi!41}\textbf{.92} & \cellcolor{sansi!45}\textbf{.99} & \cellcolor{sansi!45}\textbf{.99} & \cellcolor{sansi!45}\textbf{.99} & \cellcolor{sansi!45}\textbf{.99} & \cellcolor{sansi!44}\textbf{.98} & \cellcolor{sansi!44}\textbf{.97} & \cellcolor{qwen!40}\textbf{.89} & \cellcolor{qwen!27}\textbf{.61} & \cellcolor{smol!16}.36 \\
\bottomrule
\end{tabular}\par}
\end{minipage}\par\vspace{10pt}

\noindent\begin{minipage}{\columnwidth}
\textbf{Example 3 (MMLU, far transfer).} Fixed by the loops where both Qwen models are wrong: the answer moves from 3 to 300 between loops 2 and 3.\par
\vspace{4pt}
{\footnotesize\raggedright\ttfamily\noindent \{\\
\hspace*{1.0em}"subject": "elementary mathematics",\\
\hspace*{1.0em}"question": "If 30,000 is divided by 10 and then divided by 10 again, what will be the resulting number?"\\
\}\\
\mbox{}\\
Question: Which option correctly answers the question?\\
Options: (A) a: 3 (B) b: 30 (C) c: 300 (D) d: 3,000\\
Answer:\par}
\vspace{4pt}
\noindent{\scriptsize\setlength{\tabcolsep}{2.6pt}
\begin{tabular}{@{}l cccccccc @{\hspace{6pt}} ccc@{}}
\toprule
 & \multicolumn{8}{c}{\sansi{}, after loop} & Qwen & Qwen & Smol \\
\cmidrule(lr){2-9}
 & 1 & 2 & 3 & 4 & 5 & 6 & 7 & 8 & 4B & 2B & LM2 \\
\midrule
(A) & \cellcolor{sansi!27}\textbf{.60} & \cellcolor{sansi!27}\textbf{.61} & \cellcolor{sansi!5}.12 & \cellcolor{sansi!1}.02 & .01 & .01 & .01 & \cellcolor{sansi!1}.02 & .01 & .01 & \cellcolor{smol!7}.16 \\
(B) & \cellcolor{sansi!3}.06 & \cellcolor{sansi!2}.05 & \cellcolor{sansi!6}.14 & \cellcolor{sansi!9}.19 & \cellcolor{sansi!6}.13 & \cellcolor{sansi!4}.09 & \cellcolor{sansi!4}.08 & \cellcolor{sansi!3}.06 & \cellcolor{qwen!1}.03 & \cellcolor{qwen!1}.02 & \cellcolor{smol!6}.14 \\
\textcolor{better}{\textbf{(C)}$^{\star}$} & \cellcolor{sansi!5}.12 & \cellcolor{sansi!14}.30 & \cellcolor{sansi!32}\textbf{.71} & \cellcolor{sansi!35}\textbf{.77} & \cellcolor{sansi!38}\textbf{.85} & \cellcolor{sansi!40}\textbf{.89} & \cellcolor{sansi!41}\textbf{.90} & \cellcolor{sansi!41}\textbf{.91} & \cellcolor{qwen!4}.09 & \cellcolor{qwen!1}.03 & \cellcolor{smol!17}\textbf{.38} \\
(D) & \cellcolor{sansi!10}.22 & \cellcolor{sansi!2}.04 & \cellcolor{sansi!1}.03 & .01 & .01 & .01 & .01 & .01 & \cellcolor{qwen!39}\textbf{.87} & \cellcolor{qwen!42}\textbf{.94} & \cellcolor{smol!14}.31 \\
\bottomrule
\end{tabular}\par}
\end{minipage}\par\vspace{10pt}

\noindent\begin{minipage}{\columnwidth}
\textbf{Example 4 (ARC, far transfer).} Missing knowledge: wrong and confident at every loop.\par
\vspace{4pt}
{\footnotesize\raggedright\ttfamily\noindent \{\\
\hspace*{1.0em}"question": "When cold temperatures are produced in a chemical reaction, the reaction is known as"\\
\}\\
\mbox{}\\
Question: Which option correctly answers the question?\\
Options: (A) a: exothermic. (B) b: endothermic. (C) c: suspension. (D) d: vaporization.\\
Answer:\par}
\vspace{4pt}
\noindent{\scriptsize\setlength{\tabcolsep}{2.6pt}
\begin{tabular}{@{}l cccccccc @{\hspace{6pt}} ccc@{}}
\toprule
 & \multicolumn{8}{c}{\sansi{}, after loop} & Qwen & Qwen & Smol \\
\cmidrule(lr){2-9}
 & 1 & 2 & 3 & 4 & 5 & 6 & 7 & 8 & 4B & 2B & LM2 \\
\midrule
(A) & \cellcolor{sansi!43}\textbf{.96} & \cellcolor{sansi!43}\textbf{.96} & \cellcolor{sansi!45}\textbf{.99} & \cellcolor{sansi!45}\textbf{.99} & \cellcolor{sansi!45}\textbf{1.0} & \cellcolor{sansi!45}\textbf{1.0} & \cellcolor{sansi!45}\textbf{1.0} & \cellcolor{sansi!45}\textbf{.99} & \cellcolor{qwen!1}.02 & \cellcolor{qwen!5}.10 & \cellcolor{smol!22}\textbf{.48} \\
\textcolor{better}{\textbf{(B)}$^{\star}$} & \cellcolor{sansi!1}.03 & \cellcolor{sansi!1}.03 & .01 & .00 & .00 & .00 & .00 & .00 & \cellcolor{qwen!44}\textbf{.97} & \cellcolor{qwen!36}\textbf{.79} & \cellcolor{smol!17}.38 \\
(C) & .00 & .01 & .00 & .00 & .00 & .00 & .00 & .00 & .00 & \cellcolor{qwen!3}.07 & \cellcolor{smol!3}.06 \\
(D) & .01 & .01 & .00 & .00 & .00 & .00 & .00 & .00 & .01 & \cellcolor{qwen!2}.04 & \cellcolor{smol!4}.09 \\
\bottomrule
\end{tabular}\par}
\end{minipage}\par\vspace{10pt}

\noindent\begin{minipage}{\columnwidth}
\textbf{Example 5 (QNLI, far transfer).} Broken by the loops: right after the first two loops, wrong from the third.\par
\vspace{4pt}
{\footnotesize\raggedright\ttfamily\noindent By 9000 BP, Europe was fully forested.\\
\mbox{}\\
Question: Does the sentence contain the answer to this question: "When was Europe fully forested and recovered from the last Ice Age?"\\
Options: (A) no (B) yes\\
Answer:\par}
\vspace{4pt}
\noindent{\scriptsize\setlength{\tabcolsep}{2.6pt}
\begin{tabular}{@{}l cccccccc @{\hspace{6pt}} ccc@{}}
\toprule
 & \multicolumn{8}{c}{\sansi{}, after loop} & Qwen & Qwen & Smol \\
\cmidrule(lr){2-9}
 & 1 & 2 & 3 & 4 & 5 & 6 & 7 & 8 & 4B & 2B & LM2 \\
\midrule
(A) & \cellcolor{sansi!1}.02 & \cellcolor{sansi!17}.37 & \cellcolor{sansi!27}\textbf{.60} & \cellcolor{sansi!34}\textbf{.76} & \cellcolor{sansi!38}\textbf{.85} & \cellcolor{sansi!39}\textbf{.87} & \cellcolor{sansi!40}\textbf{.88} & \cellcolor{sansi!40}\textbf{.89} & \cellcolor{qwen!35}\textbf{.77} & \cellcolor{qwen!16}.35 & \cellcolor{smol!2}.04 \\
\textcolor{better}{\textbf{(B)}$^{\star}$} & \cellcolor{sansi!44}\textbf{.98} & \cellcolor{sansi!28}\textbf{.63} & \cellcolor{sansi!18}.40 & \cellcolor{sansi!11}.24 & \cellcolor{sansi!7}.15 & \cellcolor{sansi!6}.13 & \cellcolor{sansi!5}.12 & \cellcolor{sansi!5}.11 & \cellcolor{qwen!10}.23 & \cellcolor{qwen!29}\textbf{.65} & \cellcolor{smol!43}\textbf{.96} \\
\bottomrule
\end{tabular}\par}
\end{minipage}\par\vspace{10pt}

\noindent\begin{minipage}{\columnwidth}
\textbf{Example 6 (Kev unknowable pairs, far transfer).} An answerable item (the applicant's age is given): every loop answers ``yes''.\par
\vspace{4pt}
{\footnotesize\raggedright\ttfamily\noindent \{\\
\hspace*{1.0em}"policy": "Applicants must be at least 16 years old to be eligible for the rental agreement.",\\
\hspace*{1.0em}"case": "Elin applied to join the rental agreement. The application form was complete and signed. Elin is 18 years old."\\
\}\\
\mbox{}\\
Question: Is the applicant eligible?\\
Options: (A) no (B) yes\\
Answer:\par}
\vspace{4pt}
\noindent{\scriptsize\setlength{\tabcolsep}{2.6pt}
\begin{tabular}{@{}l cccccccc @{\hspace{6pt}} ccc@{}}
\toprule
 & \multicolumn{8}{c}{\sansi{}, after loop} & Qwen & Qwen & Smol \\
\cmidrule(lr){2-9}
 & 1 & 2 & 3 & 4 & 5 & 6 & 7 & 8 & 4B & 2B & LM2 \\
\midrule
(A) & .00 & .00 & .00 & .00 & .00 & .00 & .00 & .00 & .00 & .00 & .01 \\
\textcolor{better}{\textbf{(B)}$^{\star}$} & \cellcolor{sansi!45}\textbf{1.0} & \cellcolor{sansi!45}\textbf{1.0} & \cellcolor{sansi!45}\textbf{1.0} & \cellcolor{sansi!45}\textbf{1.0} & \cellcolor{sansi!45}\textbf{1.0} & \cellcolor{sansi!45}\textbf{1.0} & \cellcolor{sansi!45}\textbf{1.0} & \cellcolor{sansi!45}\textbf{1.0} & \cellcolor{qwen!45}\textbf{1.0} & \cellcolor{qwen!45}\textbf{1.0} & \cellcolor{smol!45}\textbf{.99} \\
\bottomrule
\end{tabular}\par}
\end{minipage}\par\vspace{10pt}

\noindent\begin{minipage}{\columnwidth}
\textbf{Example 7 (Kev unknowable pairs, far transfer).} The same item with the age removed, which makes it unanswerable (the target is the uniform distribution). The first loop still answers ``yes'' with 0.97, a hard answer; from the second loop on the top probability is below the threshold of 0.75.\par
\vspace{4pt}
{\footnotesize\raggedright\ttfamily\noindent \{\\
\hspace*{1.0em}"policy": "Applicants must be at least 16 years old to be eligible for the rental agreement.",\\
\hspace*{1.0em}"case": "Elin applied to join the rental agreement. The application form was complete and signed."\\
\}\\
\mbox{}\\
Question: Is the applicant eligible?\\
Options: (A) no (B) yes\\
Answer:\par}
\vspace{4pt}
\noindent{\scriptsize\setlength{\tabcolsep}{2.6pt}
\begin{tabular}{@{}l cccccccc @{\hspace{6pt}} ccc@{}}
\toprule
 & \multicolumn{8}{c}{\sansi{}, after loop} & Qwen & Qwen & Smol \\
\cmidrule(lr){2-9}
 & 1 & 2 & 3 & 4 & 5 & 6 & 7 & 8 & 4B & 2B & LM2 \\
\midrule
(A) & \cellcolor{sansi!1}.03 & \cellcolor{sansi!13}.28 & \cellcolor{sansi!13}.28 & \cellcolor{sansi!12}.26 & \cellcolor{sansi!13}.28 & \cellcolor{sansi!16}.35 & \cellcolor{sansi!18}.39 & \cellcolor{sansi!19}.42 & \cellcolor{qwen!32}\textbf{.72} & \cellcolor{qwen!1}.02 & \cellcolor{smol!11}.24 \\
(B) & \cellcolor{sansi!44}\textbf{.97} & \cellcolor{sansi!32}\textbf{.72} & \cellcolor{sansi!32}\textbf{.72} & \cellcolor{sansi!33}\textbf{.74} & \cellcolor{sansi!32}\textbf{.72} & \cellcolor{sansi!29}\textbf{.65} & \cellcolor{sansi!27}\textbf{.61} & \cellcolor{sansi!26}\textbf{.58} & \cellcolor{qwen!13}.28 & \cellcolor{qwen!44}\textbf{.98} & \cellcolor{smol!34}\textbf{.76} \\
\bottomrule
\end{tabular}\par}
\end{minipage}\par\vspace{10pt}

\end{document}